\documentclass[sn-basic]{sn-jnl}

\usepackage{graphicx}%
\usepackage{multirow}%
\usepackage{amsmath,amssymb,amsfonts}%
\usepackage{amsthm}%
\usepackage{mathrsfs}%
\usepackage[title]{appendix}%
\usepackage{xcolor}%
\usepackage{textcomp}%
\usepackage{manyfoot}%
\usepackage{booktabs}%
\usepackage{algorithm}%
\usepackage{algorithmicx}%
\usepackage{algpseudocode}%
\usepackage{listings}%
\usepackage{subcaption}
\usepackage{array}
\usepackage{multirow}
\usepackage{booktabs} 
\usepackage{float}
\usepackage{titlesec}
\usepackage{longtable}
\usepackage{verbatim}
\usepackage{siunitx}

\titleclass{\subsubsubsection}{straight}[\subsubsection]
\newcounter{subsubsubsection}[subsubsection]
\renewcommand{\thesubsubsubsection}{\thesubsubsection.\arabic{subsubsubsection}}
\titleformat{\subsubsubsection}{\normalfont\normalsize\bfseries}{\thesubsubsubsection}{1em}{}
\titlespacing*{\subsubsubsection}{0pt}{1.5ex plus 1ex minus .2ex}{1ex}

\theoremstyle{thmstyleone}%
\theoremstyle{thmstyletwo}%

\theoremstyle{thmstylethree}%

\begin{document}

\title[Article Title]{Histogram Approaches for Imbalanced
Data Streams Regression}

\author*[1,3]{\fnm{Ehsan} \sur{Aminian}}\email{ehsan.aminian@inesctec.pt}
\author[1,3]{\fnm{Rita P.} \sur{Ribeiro}}\email{rpribeiro@fc.up.pt}
\author[2,3]{\fnm{Joao} \sur{Gama}}\email{jgama@fep.up.pt}

\affil*[1]{\orgdiv{Faculty of Sciences}, \orgname{University of Porto}, \orgaddress{\city{Porto}, \country{Portugal}}}

\affil*[2]{\orgdiv{Faculty of Economics}, \orgname{University of Porto}, \orgaddress{\city{Porto}, \country{Portugal}}}

\affil[3]{\orgdiv{LIAAD}, \orgname{INESC TEC}, \orgaddress{\city{Porto}, \country{Portugal}}}

\abstract{
Imbalanced domains pose a significant challenge in real-world predictive analytics, particularly in the context of regression. While existing research has primarily focused on batch learning from static datasets, limited attention has been given to imbalanced regression in online learning scenarios. Intending to address this gap, in prior work, we proposed sampling strategies based on Chebyshev's inequality as the first methodologies designed explicitly for data streams. However, these approaches operated under the restrictive assumption that rare instances exclusively reside at distribution extremes. This study introduces histogram-based sampling strategies to overcome this constraint, proposing flexible solutions for imbalanced regression in evolving data streams.  The proposed techniques - Histogram-based Undersampling (HistUS) and Histogram-based Oversampling (HistOS) - employ incremental online histograms to dynamically detect and prioritize rare instances across arbitrary regions of the target distribution to improve predictions in the rare cases. Comprehensive experiments on synthetic and real-world benchmarks demonstrate that HistUS and HistOS substantially improve rare-case prediction accuracy, outperforming baseline models while maintaining competitiveness with Chebyshev-based approaches.

}

\keywords{Imbalanced Regression, Data Streams, Online Histogram}

\maketitle
\section{Introduction}
\label{Introduction}
In various application domains, predicting rare events is often more critical than forecasting frequent occurrences. This significance is particularly evident in areas such as financial fraud detection~\citep{hernandez2024financial}, where identifying uncommon fraudulent activities can prevent substantial monetary losses. Similarly, the prediction of rare medical conditions~\citep{gupta2023machine} plays a crucial role in timely interventions and treatment advancements. Additionally, understanding extreme environmental phenomena, such as catastrophic flood~\citep{snieder2021resampling} is essential for effective disaster management and mitigation strategies. These challenges are encapsulated within the concept of Imbalanced Domain Learning, which has emerged as a prominent concern in the machine learning community due to the complexities involved in accurately modeling and predicting such rare events~\citep{krawczyk2016learning}. 
In the regression context, the problem arises when the distribution of the target variable 
is highly skewed, leading to a scenario in which specific values or ranges of the target variable are considerably underrepresented compared to others. Unlike standard regression tasks that treat all target values equally, imbalanced regression requires special consideration of rare cases that often carry critical domain significance. 
The fundamental problem lies in the conflict between standard regression objectives, which aim to optimize global error metrics such as mean squared error, and the need to prioritize accurate predictions for sparsely represented target values. This imbalance poses three key challenges: 
i) identifying the rare and relevant regions within the continuous domain of the target variable;
ii) overcoming the bias of standard regression algorithms towards predicting the most frequent cases; and, iii) the need for proper evaluation metrics given the inadequacy of standard error metrics in measuring the models' performance on these rare values. 

Tackling imbalanced regression is especially challenging in the context of data streams, where information arrives continuously and models must process data in real-time. The digital age has resulted in an influx of data from sources such as IoT devices, social media, and sensors, leading to continuous, high-speed, and potentially limitless data streams. 
In data stream regression, the goal is to predict a target variable from sequentially incoming data instances continuously. This paradigm diverges fundamentally from traditional batch regression by imposing three critical operational constraints: single-pass instance processing, strict memory limitations, and continuous adaptation to non-stationary data distributions~\citep{gama2014survey,bifet2023machine}. Incremental learning emerges as the core methodology addressing these operational requirements. It is categorized into \emph{instance-incremental} learning, where the model updates with each new instance, and \emph{batch-incremental} learning, which updates in small chunks to balance efficiency and robustness~\citep{gama2014survey,montiel2018scikit}. 
Unlike traditional batch evaluation, data stream regression employs \emph{prequential evaluation}~\citep{Gama13}, where predictions are made before true labels are revealed, enabling real-time model updates~\citep{kolajo2019big}. 

This paper tackles the combined challenges of data stream learning and imbalanced domain learning by focusing on \emph{imbalanced regression in data streams}, a domain where prior works have largely centered on classification~\citep{RibeiroMoniz20}, leaving regression understudied. In our previous work~\citep{aminian2021chebyshev}, we presented our initial attempt to address imbalanced regression in data streams by introducing two methods based on Chebyshev's inequality, titled ChebyUS and ChebyOS. The methods aimed to improve predictions in the rare cases by undersampling or oversampling frequent and rare cases, respectively. However, a notable limitation of those approaches is the assumption that rare cases are confined solely to the extreme values (tails) of the target variable's distribution. 

Identifying rare regions requires knowledge of the data distribution, which is generally unavailable in data streams. Histograms offer a computationally efficient and interpretable solution for estimating the density of a target variable in streaming environments. Unlike parametric approaches, which assume a predefined distribution, histograms enable non-parametric, data-driven modeling that adapts to arbitrary and evolving data distributions. This property is particularly important to overcome the above-mentioned limitation in data stream imbalanced regression as it enables the detection of rare instances irrespective of their locations. By incrementally constructing histograms using Partition Incremental Discretization algorithm ($PiD$ for short)~\citep{gama2006discretization}, we propose two novel data-level approaches that dynamically capture variations in data density, facilitating the identification of rare regions and enabling sampling strategies that pave the way for more accurate prediction of the rare cases in any regions of the target value.

The remainder of this paper is organized as follows. Section~\ref{sec:Related_Work} reviews the relevant literature on imbalanced regression and data stream regression, highlighting existing challenges and research gaps. In Section~\ref{sec:Online_Histograms_for_Imbalanced_Regression}, we introduce our proposed histogram-based sampling approaches, HistUS and HistOS, detailing their theoretical foundations and implementation. Section~\ref{sec4:Experimental_Evaluation} presents a comprehensive experimental evaluation of our methods, including synthetic and benchmark datasets, performance metrics, and comparative analysis with the Chebyshev's-based approaches. Finally, Section~\ref{sec4:Conclusion_and_Future_work} summarizes our findings, discusses the implications of our work, and outlines potential directions for future research.

\section{Related Work}
\label{sec:Related_Work}
Our proposal intersects two domains— imbalanced regression and regression in data streams. This section reviews relevant literature for each domain in separate subsections. Finally, we identify research gaps in data stream regression.


   \subsection{Imbalanced Regression Problems}
    \label{sec-2:Identifying_Rare_Instances_in_Imbalanced_Regression}
     The challenge of \emph{imbalanced domain learning} arises in predictive tasks where a model approximates an unknown function $f: \mathcal{X} \rightarrow \mathcal{Y}$ from a training set $D = \{ \langle \mathbf{x}_i, y_i \rangle \}_{i=1}^n$, where $\mathbf{x}_i \in \mathcal{X}$ is a feature vector and $y_i \in \mathcal{Y}$ is the target variable with an uneven distribution of values. 
Standard regression prioritizes minimizing errors for frequent cases, as metrics like mean absolute error and mean squared error are dominated by common target values. However, in \emph{imbalanced regression}, obtaining accurate predictions for rare values is essential~\citep{BTR16,RibeiroMoniz20}.

The first challenge 
resides in
the identification of rare instances. Contrary to classification problems where the minority class of interest is typically well defined, in regression that definition of what are the rare relevant values is more complex by the continuous nature of the target variable.

To tackle the second challenge, which is the tendency of models to bias their predictions toward frequent target values, the existing approaches can be broadly categorized into data-level, algorithm-level, and hybrid techniques. Extensive research has explored imbalanced learning, with \emph{data-level} approaches widely adopted in classification and regression tasks~\citep{hoens2013imbalanced}. These methods modify the dataset rather than the learning algorithm, making them model-agnostic and preserving interpretability~\citep{branco2019pre}.
Among them, sampling techniques are the most common, where overersampling replicates rare instances to enhance their representation~\citep{batista2004study}, and undersampling reduces frequent instances to mitigate bias~\citep{kubat1997addressing}. Some methods combine both strategies to achieve a balanced dataset without excessive data inflation~\citep{RN91}.
A well-known data-level approach is the Synthetic Minority overersampling Technique (SMOTE) \citep{RN91}, designed initially to address class imbalance in classification tasks. 
Several extensions of SMOTE have been developed for regression tasks. The first, SMOTER~\citep{torgo2013smote}, generates synthetic instances using relevance-based methods derived from boxplot functions~\citep{ritathesis2011}. Other adaptations include SMOGN ~\citep{branco2017smogn}, which refines the synthetic generation process by the introduction of Gaussian noise; 
G-SMOTER~\citep{camacho2022geometric}, which focuses on geometric sample generation; and WSMOTER~\citep{camacho2024wsmoter} , which assigns weights to data points based on their rarity. Furthermore, beyond the SMOTE framework, methods like ROGN~\citep{branco2019pre}, which merges random overersampling with Gaussian noise, and WERCS~\citep{branco2019pre}, which allocates instance weights for a combined over- and undersampling strategy.

The third challenge lies in effectively evaluating model performance in predicting rare target values. To address this, specialized evaluation criteria have been proposed, including the $\phi()$-weighted Root Mean Square Error ($RMSE_{\phi}$)~\citep{ritathesis2011} and the Squared Error Relevance Area ($SERA$)~\citep{RibeiroMoniz20}. $RMSE_{\phi}$ emphasizes underrepresented regions by incorporating a relevance function $\phi() \in [0,1]$, derived from the target values' boxplot, into the standard $RMSE$. It only considers instances with relevance values above a given threshold, reweighting their prediction errors by multiplying the prediction error for each example by the relevance value assigned to it. $SERA$, in contrast, evaluates errors across the entire domain while placing greater emphasis on relevant instances. Unlike $RMSE_{\phi}$, which applies a single threshold, $SERA$ progressively integrates errors at multiple relevance levels, capturing both overall predictive performance and the model’s ability to handle rare cases. Consequently, $SERA$ serves as an intermediate metric between standard $RMSE$ and $RMSE_{\phi}$, balancing global error assessment with a focus on rare target values.

    \subsection{Learning from Data Streams}
    \label{sec-2:Regression_from_Data_Streams_Related_Work}
    The complexity increases in \emph{online learning}, where data arrives sequentially - unlike batch learning where the entire dataset is available at once - as pairs $\langle \mathbf{x}_t, y_t \rangle$, drawn from an unknown probability distribution $p_t(\mathbf{x}, y)$. At each time step $t$, the learner maintains a hypothesis $H_t: \mathcal{X} \rightarrow \mathcal{Y}$, predicts $\hat{y}_t = H_t(\mathbf{x}_t)$, and updates $H_t$ upon receiving the true $y_t$. 
Depending on the nature of $y$, the learning task is classified as classification when $y$ is categorical and as regression when $y$ is continuous.

Resampling techniques are widely used in handling class imbalance in data stream by either oversampling the minority class or undersampling the majority class. Incremental Oversampling for Data Streams (IOSDS)~\citep{anupama2019novel} selectively replicates minority instances while filtering out noisy or overlapping examples to maintain model stability. Selection-Based Resampling (SRE)~\citep{ren2019selection} iteratively removes majority instances, preventing excessive bias towards the minority class. Streaming versions of the popular SMOTE algorithm, such as Adaptive Windowing SMOTE (AWSMOTE)~\citep{wang2021awsmote} and Continuous SMOTE (C-SMOTE)~\citep{bernardo2020c}, generate synthetic instances dynamically based on evolving class distributions. Another method, Very Fast Continuous SMOTE (VFC-SMOTE)~\citep{bernardo2021vfc}, enhances reactivity to sudden shifts in data distributions by utilizing data sketching techniques.

Several approaches have also been proposed for online regression, including tree-based methods such as FIMT-DD~\citep{ikonomovska2011learning}, ORF~\citep{ikonomovska2015online}, ORTO-A and ORTO-BT~\citep{ikonomovska2011speeding}, ARF-Reg~\citep{gomes2018adaptive}, iSOUP-PCT~\citep{osojnik2020incremental}, as well as neural network-based models like OS-ELM~\citep{huang2005line} and ROS-ELM~\citep{liu2015ros}. Aiming to cope with the challenges of online imbalanced regression, we have introduced~\citep{aminian2021chebyshev} two sampling strategies: ChebyUS and ChebyOS, 
designed to enhance the learning of rare and extreme values in data streams. These methods leverage Chebyshev's probability to guide undersampling and oversampling, thereby improving predictive accuracy in imbalanced regression settings. ChebyUS applies an undersampling strategy where each incoming instance is assigned a probability based on Chebyshev's probability. Instances with higher probabilities (frequent cases) will likely be discarded, ensuring that the training set focuses on rare samples. In contrast, ChebyOS employs an oversampling strategy, converting the assigned probability into an integer factor, $K$, which determines how often the model is trained on that instance.

Histograms are widely used in \textit{batch processing} to approximate data distributions by partitioning static datasets into discrete bins. However, adapting histograms to \textit{data streams} poses significant challenges, as traditional methods rely on multiple data passes and whole dataset storage—both infeasible in streaming environments. \citet{garofalakis2002querying} pioneered \textit{incremental histogram techniques} to address these challenges, enabling dynamic adjustments to bin boundaries and frequencies as new data arrives. While their methods advanced real-time summarization, they struggled with \textit{non-uniform or skewed distributions}, where abrupt variations in data density rendered static binning strategies ineffective~\citep{bahri2021data}. To overcome this limitation, \citet{gama2006discretization} introduced the \textit{Partitional Incremental Discretization} ($PiD$) method, a streaming-based discretization technique that incrementally partitions continuous attributes without requiring historical data storage. Unlike traditional histogram approaches that maintain fixed bin structures, $PiD$ dynamically updates partitions based on incoming data, making it adaptable to evolving distributions. $PiD$ initializes without predefined bin boundaries and continuously refines partitions as new data arrives. No updates are necessary when a data point falls within an existing partition. However, suppose a new data point lies outside the current partition structure. In that case, $PiD$ adjusts the boundaries by applying a \emph{greedy splitting strategy}, ensuring that partitions reflect the most recent distribution. This strategy enables $PiD$ to maintain a compact and relevant histogram representation in real-time environments while being efficient in memory and computation.

    \subsection{Research Gaps}
    \label{sec-2:Research_Gaps}
    Existing imbalanced regression methods largely assume static datasets with full data access, enabling precise rarity estimation. However, these assumptions break down in online regression, where data arrives incrementally, and target distributions evolve dynamically. Most of data-level strategies for imbalanced regression require complete dataset access, some relying 
on kernel density estimation, which becomes infeasible under streaming constraints. Algorithm-level approaches, such as cost-sensitive learning, struggle with scalability due to their reliance on batch updates. Additionally, relevance functions based on fixed statistical thresholds fail to adapt to shifting data distributions. This highlights a critical gap: the lack of mechanisms to dynamically identify rare instances of the continuous target variable and adjust sampling strategies in real time without storing historical data. While our prior work~\citep{aminian2021chebyshev} tackled this problem in scenarios where rare instances are located in the extreme tails of the target value distribution, it remains limited in capturing the full spectrum of rarity, underscoring the need for a more adaptive and comprehensive solution.

\section{Online Histograms for Imbalanced Regression}
\label{sec:Online_Histograms_for_Imbalanced_Regression}
Leveraging online histograms produced by the 
\emph{Partitional Incremental Discretization} (PiD) method described in~\citep{gama2006discretization}, we propose two novel data-level sampling strategies to tackle imbalanced regression in data streams. The first approach, termed Histogram-based Undersampling (HistUS), reduces the prevalence of frequent instances, while the second approach, called Histogram-based Oversampling (HistOS), increases the representation of rare cases in the training of the learner models. The following subsections provide a detailed explanation of each method.

\subsection{Probability Function for Informed Sampling}
    \label{sec4:ProbabilityFunction}
    We propose to provide informed sampling approaches for building regression models from imbalanced data streams, guided by the $PiD$ method~\citep{gama2006discretization}. By employing counts and bins extracted from a dynamically created online histogram, we compute a probability value that offers insight into the degree of rarity associated with each example. A high probability suggests that the example is frequent, while a low probability indicates that it is rare. The primary objective of this probability function is to regulate the selection of instances for training, thereby implementing sampling strategies based on instance rarity. It leverages count values derived from the histogram constructed by PiD and is designed to be adaptable, enabling fine-tuning based on domain-specific knowledge of the underlying data distribution.

Assuming that the target value of the $i$-th instance is associated with the $j$-th bin, i.e. $y_i \in \left[b_{j-i},b_j\right)$, the probability for that instance is computed by
\begin{equation}
    P(i) = \exp(-\beta \times \rho_j)
    \label{equa3}
    \end{equation}

\noindent where $\beta$ is a parameter that controls the degree of sampling, and $\rho_j$ represents 
the relative density of the $j$-th bin to which $y_i$ belongs that is defined by

    \begin{equation}
    \rho_j = \frac{{n_j}}{{
    \max\left\{n_l:l=1,\dots,k\right\}}}
    \label{equa4}
    \end{equation}

\noindent 
where $n_j$ and  $n_l$ refer to the number of instances in the $j$-th and the $l$-th bin, respectively and $k$ is the total number of bins. 
This scaling ensures that the count values are normalized between 0 and 1. The exponential function in Equation~\ref{equa3} assigns a higher probability to instances in bins with lower counts (rarer instances) and a lower probability to instances in bins with higher counts (more frequent instances). The parameter $\beta$ adjusts the steepness of this probability decay.

    \subsection{HistUS: Histogram-based Undersampling Approach} 
    \label{sec4:HistUS}
    Drawing on the probability function outlined in the previous section, we now introduce our undersampling method, HistUS. An initial histogram is generated based on samples from the data stream using the $PiD$ algorithm. This histogram serves as a heuristic for assessing the expected rarity of incoming examples.

Algorithm~\ref{alg:histUS_Combined} presents the main procedure, HistUS, which initializes the model and histogram and orchestrates the training and testing phases. Training the learner model without a prior estimation of the $PiD$ model is ineffective. So, the warming phase is introduced to construct an initial estimation of the histogram built by the $PiD$ model. During the training phase, the learner model is trained using an undersampling strategy based on probabilities derived from the histogram. Simultaneously, the $PiD$ model is updated throughout the training process and, as we also see in the testing phase, to ensure that the probabilities remain accurate and reflective of the evolving data distribution. Throughout the testing phase, the model continues to process the data stream, making predictions and updating itself. 
During the training and testing phases, the probability value (i.e. $p$) is calculated for each target value $y$, representing its frequency within the data stream. A random number $r$ is then generated, and if 
$r \leq p$, the instance is selected for training. This undersampling approach mitigates the over-representation of frequent cases, resulting in a more balanced training dataset.

\begin{algorithm}[tb]
    \scriptsize
  \caption{HistUS: Histogram-based Undersampling Algorithm for Data Streams}
  \label{alg:histUS_Combined}
  \begin{algorithmic}[1]
    \Procedure{HistUS}{$\mathcal{S}$: data stream, $n_{w}$: warming set size, $n_{t}$: training set size, $\beta$: sampling decay}
      \State $model \gets InitializeModel()$ \Comment{Create an empty predictive model}
      \State $hist \gets PiD()$ \Comment{Create an empty histogram model}
      \State $i \gets 0$
      \While{$i \leq n_{w}$} \Comment{Warming Phase}
          \State $\langle \mathrm{x}, y\rangle \gets GetExample(\mathcal{S})$ \Comment{Get next example from data stream}
          \State $hist \gets UpdatePiDModel(hist, y)$
          \State $i \gets i+1$
      \EndWhile
      \State $i \gets 0$
      \While{$i \leq n_{t}$} \Comment{Training Phase}
          \State $\langle \mathrm{x}, y\rangle \gets GetExample(\mathcal{S})$
          \State $hist \gets UpdatePiDModel(hist, y)$
          \State $p \gets ComputeProbability(y, hist, \beta)$
          \State $r \gets RandomNumber()$
          \If{$r \leq p$}
              \State $model \gets TrainModel(model,\langle \mathrm{x}, y\rangle)$
          \EndIf
          \State $i \gets i+1$
      \EndWhile
      \While{($GetExample(\mathcal{S}) \neq NULL) $} \Comment{Testing Phase}
          \State $\langle \mathrm{x}, y\rangle \gets GetExample(\mathcal{S})$
          \State $\hat{y} \gets Predict(model, y)$
          \State $hist \gets UpdatePiDModel(hist, y)$
          \State $p \gets ComputeProbability(y, hist, \beta)$
          \State $r \gets RandomNumber()$
          \If{$r \leq p$}
              \State $model \gets TrainModel(model,\langle \mathrm{x}, y\rangle)$
          \EndIf
      \EndWhile
    \EndProcedure
  \end{algorithmic}
\end{algorithm}

    \subsection{HistOS: Histogram-based Oversampling Approach} 
    \label{sec4:HistOS}
    The estimated histogram can also be employed to oversample rare cases. The learning model may undergo multiple training iterations for a single incoming example by utilizing probability values derived from the histogram's counts and bins. The implementation of this approach is detailed in Algorithm~\ref{alg:histOS_Combined}, which closely resembles the structure of the undersampling approach while introducing a reduction coefficient $\alpha$. This parameter regulates the oversampling rate by progressively reducing the computed probability for each incoming example until the termination condition in the training and test phases is satisfied.

During the training phase, the data stream is processed sequentially, updating both the histogram model ($hist$) based on the $PiD$ method and the learner model ($model$). Each example is used for training at least once. However, rare examples that have a higher probability value $p$ may undergo multiple training iterations, enhancing the model's ability to learn from infrequent cases. The training process continues until the predefined training set size ($n_t$) is reached. The probability $p$ is computed based on the histogram and adjusted using the reduction coefficient $\alpha$ and the sampling decay parameter $\beta$. If a randomly generated number $r$ satisfies $r \leq p$, the model undergoes additional training iterations, with $p$ being progressively reduced by a factor of $\alpha$ after each iteration. This mechanism ensures that rare instances contribute more significantly to the training process while preventing excessive oversampling.
A smaller $\alpha$ value results in more frequent training iterations for rare cases, improving the model's capacity to learn from infrequent events. The while loop continues until the condition $r > p$ is met, ensuring that additional training iterations are proportional to the rarity of the example. The testing phase continuously processes the data stream, making predictions, updating the histogram, and retraining the model. The oversampling mechanism remains consistent with the training phase, ensuring continued model adaptation.

\begin{algorithm}[!h]
    \scriptsize
  \caption{HistOS: Histogram-based oversampling Algorithm for Data Streams}
  \label{alg:histOS_Combined}
  \begin{algorithmic}[1]
    \Procedure{HistOS}{$\mathcal{S}$: data stream, $n_{w}$: warming set size, $n_{t}$: training set size, $\alpha$: reduction coefficient, $\beta$: sampling decay}
      \State $model \gets InitializeModel()$ \Comment{Create an empty predictive model}
      \State $hist \gets PiD()$ \Comment{Create an empty histogram model}
      \State $i \gets 0$
      \While{$i \leq n_{w}$} \Comment{Warming Phase}
          \State $\langle \mathrm{x}, y\rangle \gets GetExample(\mathcal{S})$ \Comment{Get next example from data stream}
          \State $hist \gets UpdatePiDModel(hist, y)$
          \State $i \gets i+1$
      \EndWhile
      \State $i \gets 0$
      \While{$i \leq n_{t}$} \Comment{Training Phase}
          \State $\langle \mathrm{x}, y\rangle \gets GetExample(\mathcal{S})$
          \State $hist \gets UpdatePiDModel(hist, y)$
          \State $model \gets TrainModel(model,\langle \mathrm{x}, y\rangle)$
          \State $p \gets ComputeProbability(y, hist, \alpha, \beta)$
          \State $r \gets RandomNumber()$
          \While{$r \leq p$}
              \State $model \gets TrainModel(model,\langle \mathrm{x}, y\rangle)$
              \State $p \gets p / \alpha$
          \EndWhile
          \State $i \gets i+1$
      \EndWhile
      \While{($GetExample(\mathcal{S}) \neq NULL) $} \Comment{Testing Phase}
          \State $\langle \mathrm{x}, y\rangle \gets GetExample(\mathcal{S})$
          \State $\hat{y} \gets Predict(model, y)$
          \State $hist \gets UpdatePiDModel(hist, y)$
          \State $p \gets ComputeProbability(y, hist, \alpha, \beta)$
          \State $r \gets RandomNumber()$
          \While{$r \leq p$}
              \State $model \gets TrainModel(model,\langle \mathrm{x}, y\rangle)$
              \State $p \gets p / \alpha$
          \EndWhile
      \EndWhile
    \EndProcedure
  \end{algorithmic}
\end{algorithm}

\section{Experimental Evaluation} 
\label{sec4:Experimental_Evaluation}

    This section presents the experimental evaluation of the proposed histogram-based methods for handling imbalanced regression in data streams. The primary objective of this evaluation is to assess the effectiveness of the histogram-based undersampling (HistUS) and oversampling (HistOS) strategies in improving the predictive performance of regression models, particularly for rare and extreme target values. To this end, we conduct extensive experiments on synthetic and real-world datasets, integrating the proposed methods into widely used regression algorithms. The evaluation focuses on key performance metrics, including relevance-weighted error functions, to evaluate the models based on their ability to capture rare instances. The following subsections provide a detailed description of the experimental setup, the used datasets, the evaluation metrics, the obtained results, and a comprehensive analysis of the findings.

    \subsection{Relevance Function}     
    \label{sec4:Relevance_function}
    In our pursuit of conveying the varying importance attributed to different values of a target variable, we propose a new method to define the so-called relevance function \mbox{$\phi\colon\mathcal{Y} \to [0,1]$} that maps the continuous target variable domain into a scale of importance~\citep{RibeiroMoniz20}.
Through our method, the relevance linked to each target value is contingent on the number of instances within the same range, using counts derived from the Equal-width Histogram method. 
We begin by constructing an equal-frequency histogram for the domain of the target variable, defined by a set of $k$ bins.
Comparing the count values associated with each bin interval can give us insights into each interval's density. So, depending on the specific bin interval in which a target value resides, we gauge its relative density to determine its relevance. The relevance value for a target value $y_i$ belonging to the $j$-th bin
is estimated by
\begin{equation}
\phi(y_i) = 1 - \rho_j 
\end{equation}
\noindent where $\rho_j$ represents the relative density of the $j$-th bin, 
as defined by Equation~\ref{equa4}.

The function $\phi()$ is a non-continuous function that conveys domain-specific bias, with values of $0$ and $1$ representing the minimum and maximum relevance, respectively. The relevance values for bins containing a high number of examples, typically considered frequent cases, are low, whereas the values are high for bins with a small number of examples.

    \subsection{Experiments on Synthetic and Real-World Data}
    \label{sec4:results_synthetic}
To evaluate the effectiveness of our approach in handling data streams where rare cases may emerge at any point—contrary to our previous assumption that rarity is confined to extreme values—we conduct experiments on both a synthetic and a real-world dataset titled \emph{Range Queries Aggregates}~\citep{Dua19} . In this section, we provide a detailed analysis of the experimental setup and the impact of our sampling strategies, focusing on the synthetic dataset. This choice is motivated by its lower dimensionality and the well-defined relationship between its independent and dependent variables, which facilitate visualization and interpretation. Nevertheless, the observed behavior is consistent across the real-world dataset.
Further details on the Range Queries Aggregates dataset are provided in Appendix~\ref{apdx:Range-Queries-Aggregates}.

We created a simple synthetic dataset and the details of the dataset are provided in Appendix~\ref{apdx:Synthetic_Data_Structure}. Figure~\ref{fig:sythetic_data_entire_dataset_histogram} illustrates the distribution of target values \( y \) in the synthetic dataset, emphasizing the distinction between frequent and rare instances through color-coded segments. A threshold (\(thr_\phi\)) is applied to the relevance function to identify rare values, with all instances where \(thr_\phi \geq 0.9\) classified as rare cases.

\begin{figure}[bt]
    \centering
      \begin{subfigure}{.30\textwidth}
        \centering    \includegraphics[width=\textwidth]{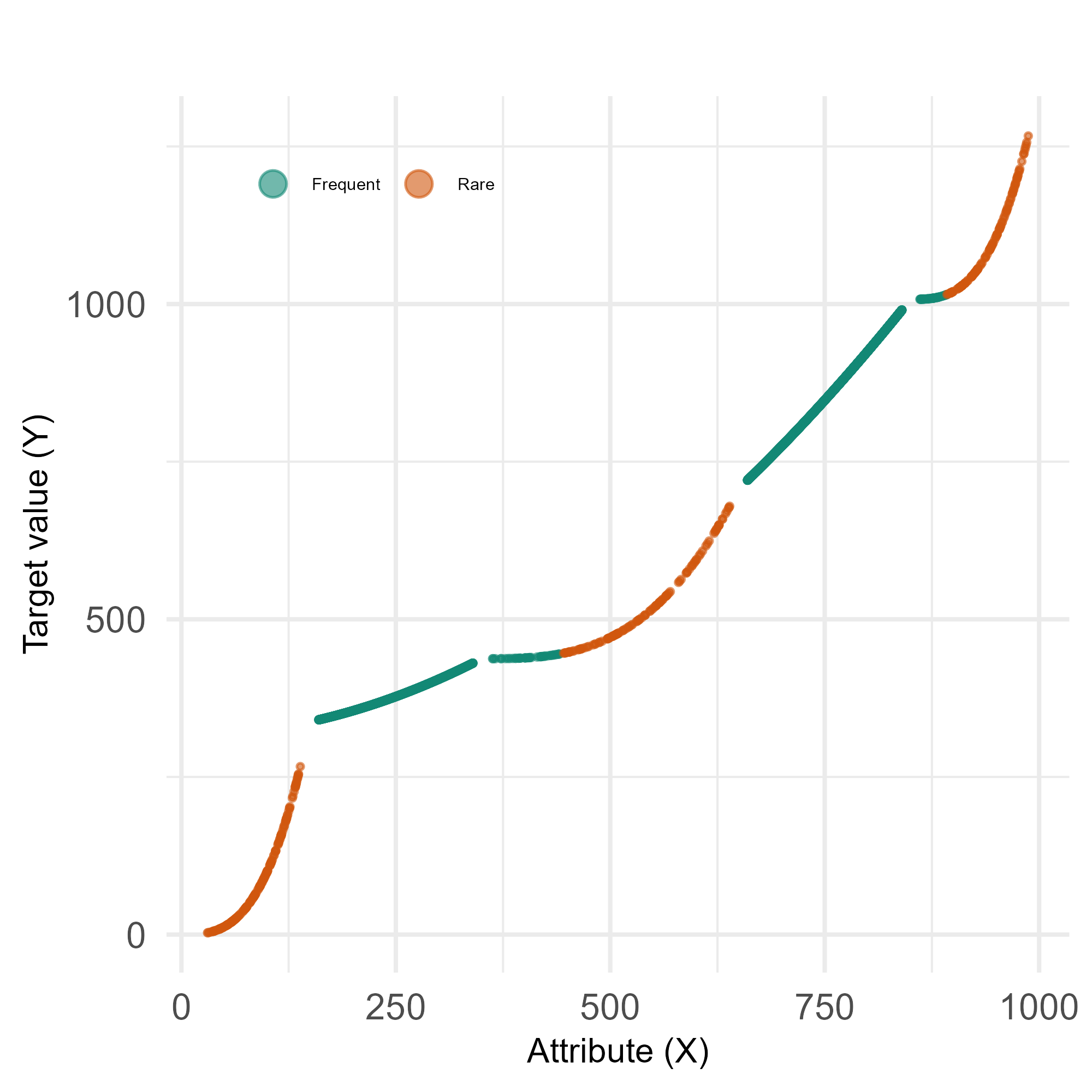}
        \caption{}
        \label{fig:Synthetic_X_Y}
    \end{subfigure}
    \begin{subfigure}{.30\textwidth}
        \centering    \includegraphics[width=\textwidth]{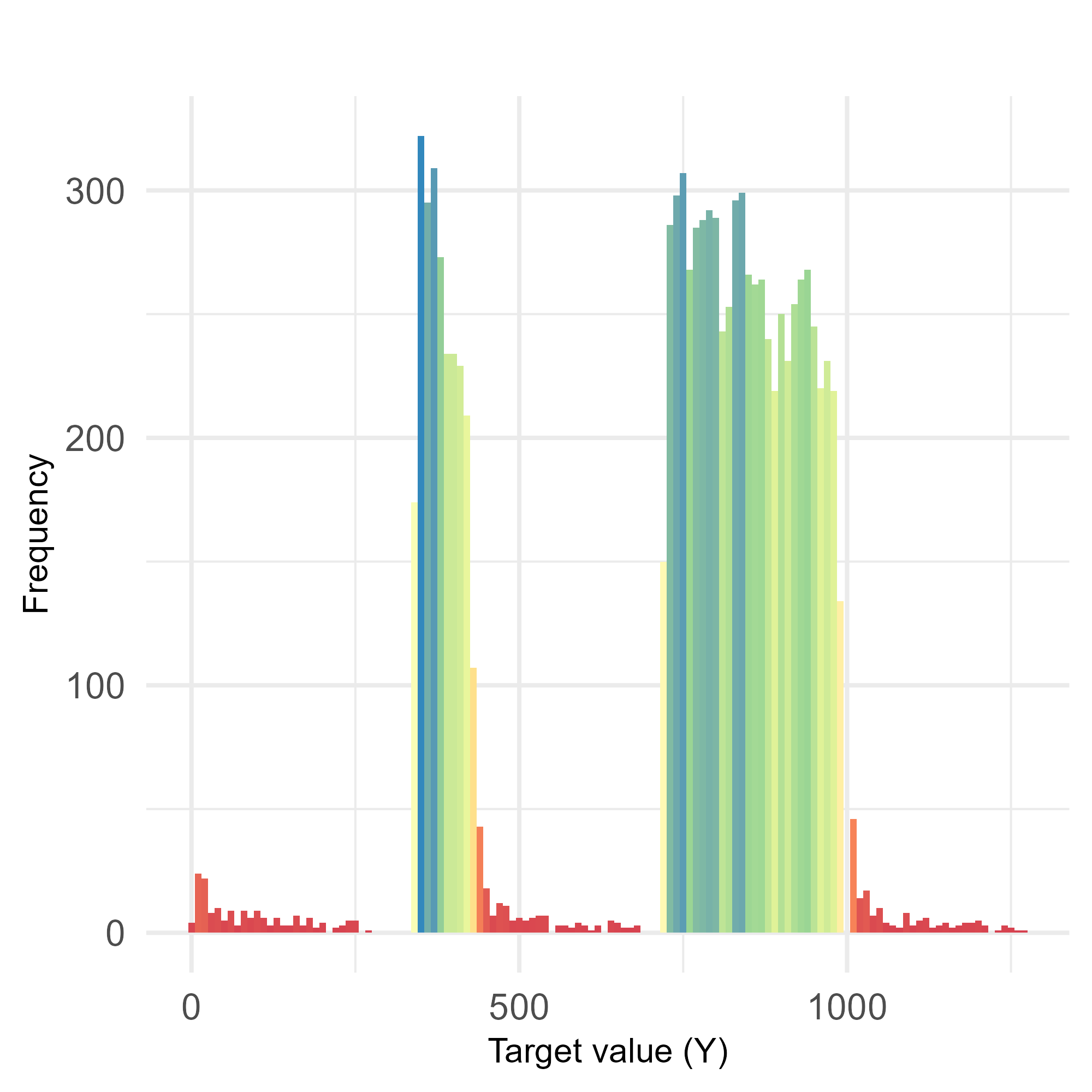}
        \caption{}
    \label{fig:Synthetic_histogram}
    \end{subfigure}%
    \begin{subfigure}{.30\textwidth}
        \centering    \includegraphics[width=\textwidth]{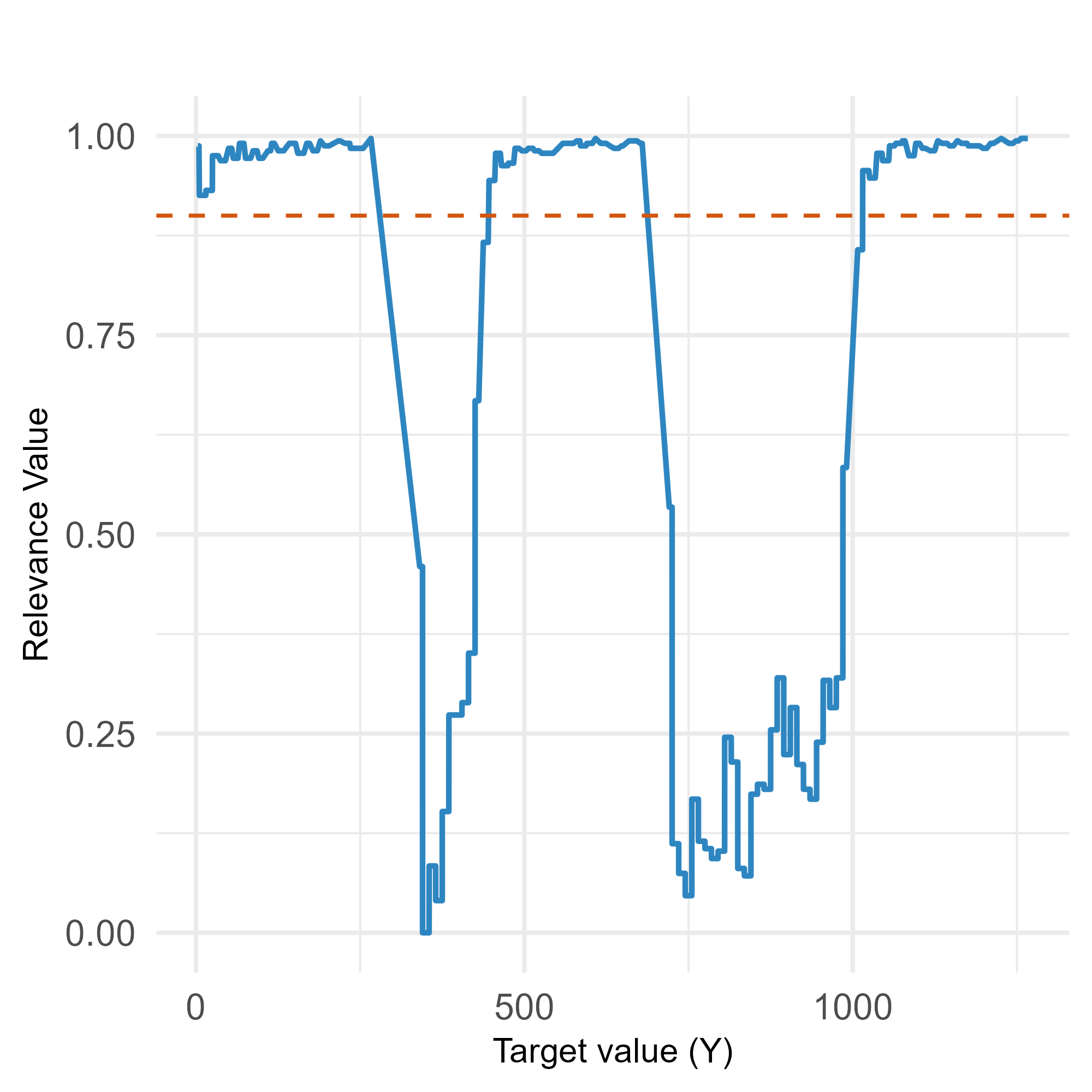}
        \caption{}
        \label{fig:Synthetic_relevanceValues}
    \end{subfigure}
    \hfill

    \caption{Synthetic dataset (a), histogram of target values (b) and respective relevance values (c) of common and rare cases considering $thr_\phi$ of 0.9.}\label{fig:sythetic_data_entire_dataset_histogram}
\end{figure}

Figure~\ref{fig:Synthetic_X_Y} illustrates the relationship between the independent variable \( x \) and the dependent variable \( y \), highlighting the distinction between rare and frequent cases. Figure~\ref{fig:Synthetic_histogram} depicts the histogram of \( y \), demonstrating the concentration of frequent instances within specific intervals and the scattered distribution of rare cases. Figure~\ref{fig:Synthetic_relevanceValues} represents the relevance values assigned to different target values, showing a sharp decline in relevance around the densely populated frequent regions, while maintaining high values in the sparse rare areas. The dashed orange line represents the threshold (\(thr_\phi\)) used to distinguish rare cases.

\begin{figure}[bt] 
    \centering
    \includegraphics[width=0.6\textwidth]{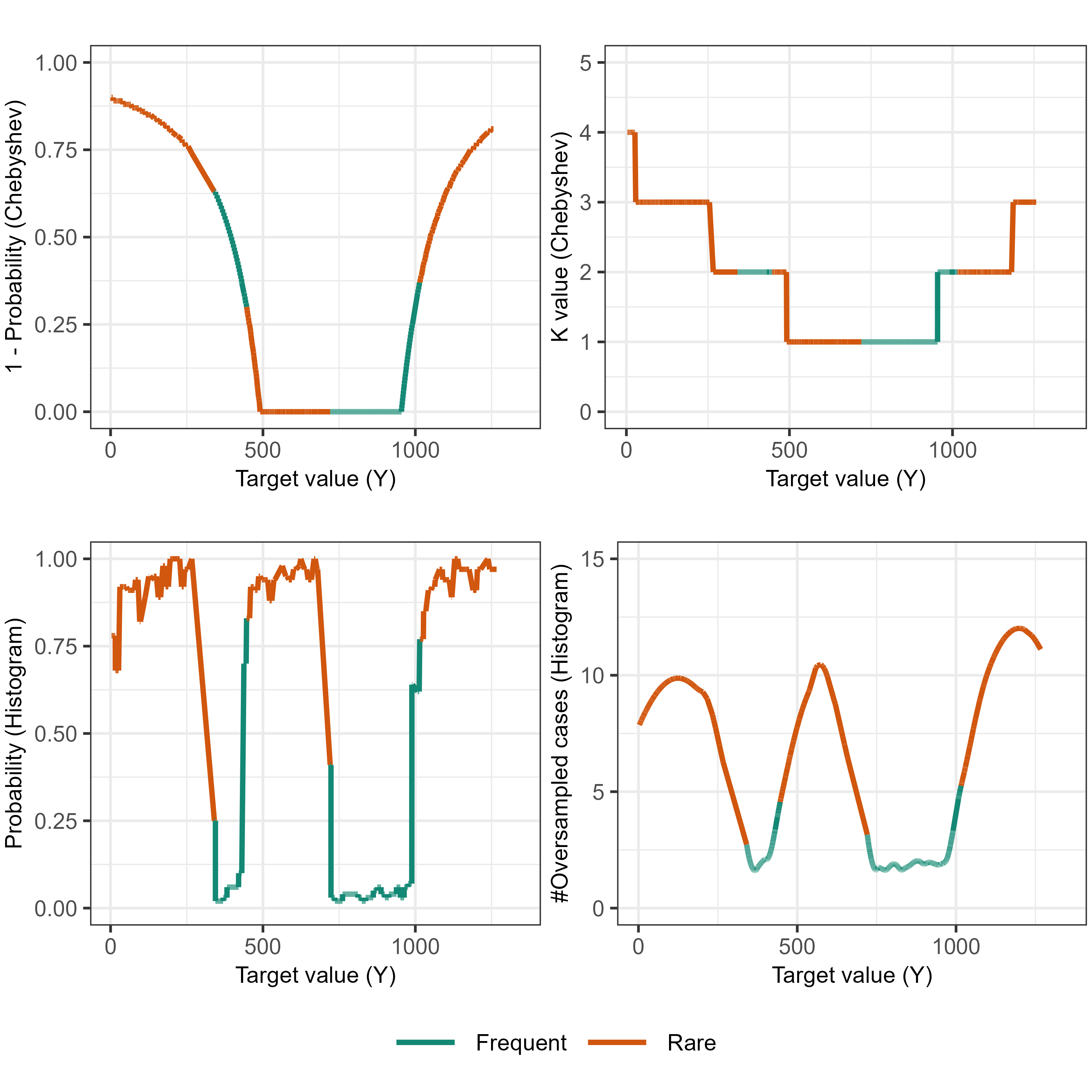}
    \caption{Comparison of Probability and \( K \) values between the Chebyshev-based and the histogram-based approaches for the synthetic dataset.}
        \label{fig:sythetic_data_probablilityRelevancy}
\end{figure}

To illustrate the limitations of Chebyshev-based approaches for this type of dataset, we depict its derived values in Figure \ref{fig:sythetic_data_probablilityRelevancy}. 
The first two plots at the top of the figure depict the inverse Chebyshev probability (\(1 - \text{Probability}\)) and the \(K\) values. The inverse Chebyshev probability governs the selection likelihood of instances in the ChebyUS method, prioritizing rare examples (with high inverse probability values). Conversely, \(K\) dictates the replication frequency of examples during training in the ChebyOS method, where larger \(K\) values correspond to repeated oversampling of rare instances. These plots highlight how the Chebyshev-based probability measure assigns low values in the central region of the distribution while increasing toward the tails. Despite the presence of rare cases in the central region — confirmed by high relevance values in Fig~\ref{fig:sythetic_data_probablilityRelevancy} —the Chebyshev probability and \( K \) values remain low, leading to an underestimation of rare instances in these regions. This results in inadequate model training, as the approach assumes that rarity is confined to the distribution's tails.

The bottom plots in the figure illustrate the probability values and the number of oversampled cases obtained through histogram-based approaches. Unlike Chebyshev, these methods effectively detect rare cases across the entire distribution, including the central region. The higher probability and oversampling values in this region demonstrate the superiority of the histogram-based approach in capturing rare instances, ensuring sufficient training data across all rare areas of the dataset. Consequently, we expect these methods to provide a more reliable framework for training machine learning models in scenarios where rare cases are not 
at the tails of the distribution.

\begin{figure}[!hbtp] 
    \centering
    \includegraphics[width=0.9\textwidth]{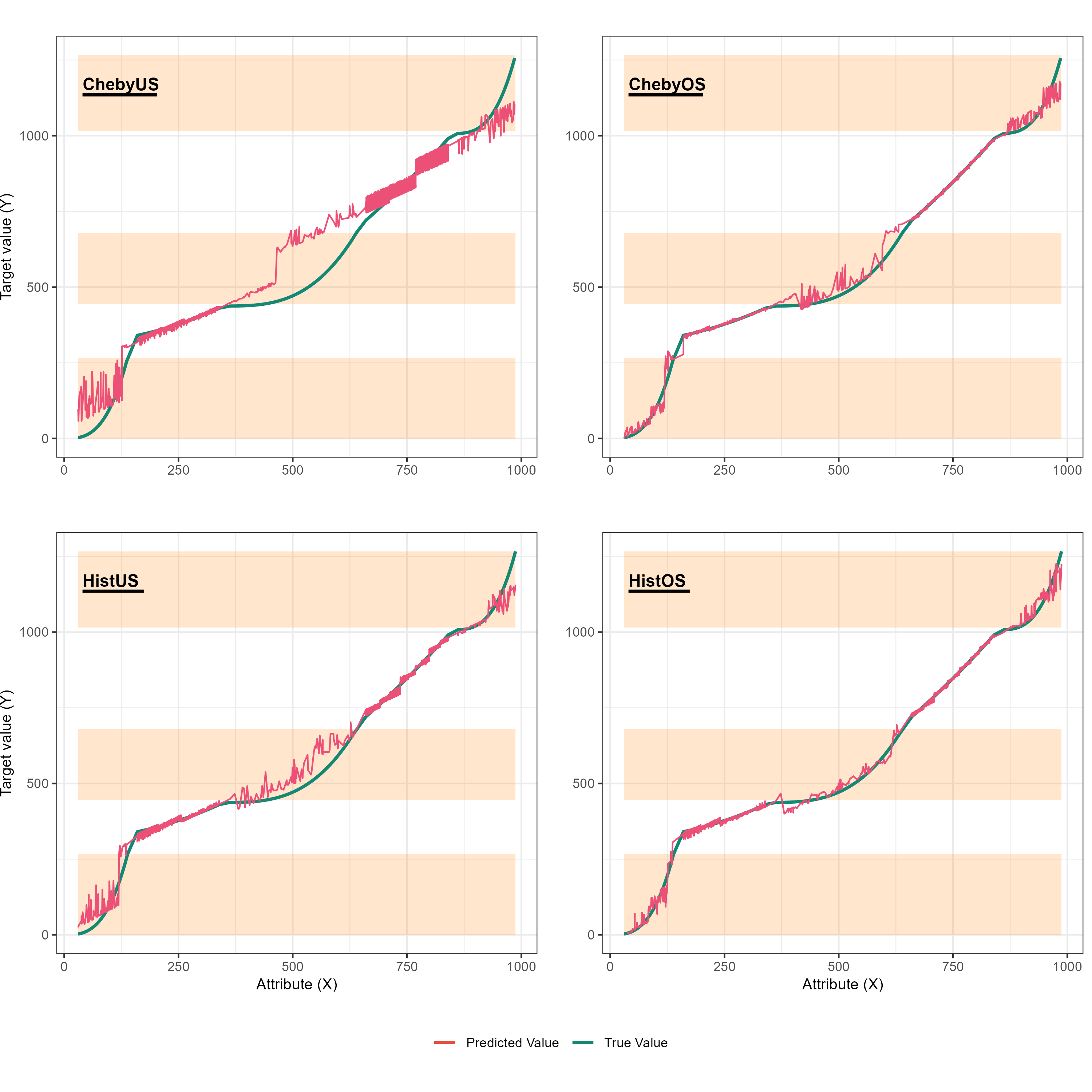}
    \caption{True target values and respective predictions using the Chebyshev-based sampling and histogram-based sampling for the synthetic dataset. The orange-highlighted areas in the background correspond to the rare regions in the target value domain.}
    \label{fig:sythetic_data_SynthticChebyHist}
\end{figure}

Figure~\ref{fig:sythetic_data_SynthticChebyHist} presents the predicted values for the target variable using FIMT-DD models with default parameters, trained with the proposed Chebyshev- and histogram-based strategies. Since the target variable domain consists of distinct regions governed by different nonlinear functions and FIMT-DD can identify these splits and fit local models for each segment, we selected this model. The top plots in the figure represent models trained with the Chebyshev-based approach, while the bottom plots illustrate the performance of models trained using histogram-based methods. 

A clear discrepancy is observed between the predictions: histogram-based models exhibit a superior ability to capture rare cases in the central region of the target variable. This aligns with our earlier findings in Figure~\ref{fig:sythetic_data_probablilityRelevancy}, where the Chebyshev-based probability measure failed to effectively identify rare cases in this region. As a result, models trained with the Chebyshev approach show higher prediction errors, as reflected in the deviation of predicted values from the true values. In contrast, histogram-based models successfully learn from rare cases throughout the distribution, resulting in improved predictive performance and better generalization.

The results for the both datasets (synthetic and Range Queries Aggregates) are summarized in Table~\ref{tab:result_phi_sera_synthetic}, which presents the average errors obtained from FIMT-DD models trained with different sampling strategies over ten runs, evaluated using $RMSE_{\phi}$, $SERA$, and $RMSE$ metrics. The table also includes the standard deviations and the ranking of each method based on predictive performance, where lower values indicate better performance. In our implementation, the $\beta$ and $\alpha$ parameters for HistUS and HistOS are set to $4$ and $1.02$, respectively. For the ChebyUS approach, we set the second chance parameter to $0.15$, following the value used in its original paper~\citep{aminian2021chebyshev}. 

The results indicate that histogram-based sampling methods (HistUS and HistOS) consistently outperform their Chebyshev-based counterparts (ChebyUS and ChebyOS) across all evaluation metrics. Among them, HistOS achieves the lowest $RMSE$ and $RMSE_{\phi}$ values, demonstrating its effectiveness in improving predictive accuracy and minimizing error propagation. It is important to note the difference in magnitude between $RMSE$ and $RMSE_{\phi}$. The $RMSE_{\phi}$ metric only considers the prediction errors for instances with a relevance value greater than a predefined threshold ($thr_\phi$), set to 0.90 in this paper, whereas $RMSE$ accounts for the prediction error across the entire dataset. Since most instances belong to frequent regions, and models tend to learn these regions well due to the abundance of training examples, the overall $RMSE$ remains lower. In contrast, rare regions, which are given higher importance in $RMSE_{\phi}$, are more challenging to learn due to fewer training examples, leading to a higher error magnitude in $RMSE_{\phi}$ compared to $RMSE$. 

For the $SERA$ metric, ChebyUS achieves the lowest score, suggesting its strength in capturing overall errors across the entire domain. However, when evaluating predictive performance specifically in rare regions, as measured by $RMSE_{\phi}$, ChebyUS performs significantly worse than histogram-based methods, ranking last in this metric. Meanwhile, ChebyOS, although more accurate than undersampling approaches (ChebyUS and HistUS) in rare regions, still falls short of outperforming its histogram-based counterpart, HistOS. Overall, the findings confirm the superiority of histogram-based methods, particularly HistOS, in achieving lower errors across different evaluation criteria. 

To finalize this analysis, it is essential to emphasize that the computed errors account for all rare cases across the entire distribution, including the extreme values in the tails of the target value distribution.  As a result, in certain models and datasets, Chebyshev-based methods may produce lower errors for extreme rare values compared to histogram-based methods, leading to overall lower metric values.    

\begin{table}[!bt]
    \centering
    \caption{
    Summary of results of no sampling (Baseline) and sampling strategies (HistUS, HistOS, ChebyUS, ChebyOS) on the synthetic and Range Queries Aggregates (RQA) datasets: mean, standard deviation and average rank of $RMSE_{\phi}$, $SERA$, and $RMSE$ for the FIMT-DD algorithm.}
    \label{tab:result_phi_sera_synthetic}
 \sisetup{table-format=3.2, detect-weight=true, detect-inline-weight=math}
    \begin{tabular*}{\textwidth}{@{\extracolsep{\fill}} ll 
        S[table-format=3.2] @{} l @{\hspace{-0.5em}} r 
        S[table-format=1.2] @{} l @{\hspace{-0.5em}} r 
        S[table-format=2.2] @{} l @{\hspace{-0.5em}} r}
        \toprule
        \textbf{Dataset} & \textbf{Approach} & \multicolumn{2}{c}{\textbf{$RMSE_{\phi}$}} & avg.rank & \multicolumn{2}{c}{\textbf{$SERA$}} & avg.rank & \multicolumn{2}{c}{\textbf{$RMSE$}} & avg.rank \\
        \midrule
        \multirow{5}{*}{\textbf{Synthetic}} 
        & Baseline  & 61.16 & $\pm$ 5.12  & (4.0)  & 0.87 & $\pm$ 0.00  & (4.2)  & 13.93 & $\pm$ 0.28  & (4.0)  \\
        & HistUS    & 51.38 & $\pm$ 10.22 & (3.0)  & 0.78 & $\pm$ 0.00  & (2.7)  & 12.62 & $\pm$ 0.44  & (3.0)  \\
        & HistOS    & \textbf{26.17} & $\pm$ \textbf{2.02}  & (1.0)  & 0.76 & $\pm$ 0.00  & (2.3)  & \textbf{6.82} & $\pm$ \textbf{0.28}  & (1.0)   \\
        & ChebyUS   & 90.31 & $\pm$ 83.02 & (5.0)  & \textbf{0.59} & $\pm$ \textbf{0.00}  & (1.0)  & 26.63 & $\pm$ 3.33  & (5.0)  \\
        & ChebyOS   & 37.87 & $\pm$ 8.09  & (2.0)  & 0.89 & $\pm$ 0.00  & (4.8)  & 8.60 & $\pm$ 0.26  & (2.0)   \\

        \midrule
        \multirow{5}{*}{\textbf{RQA}} 
        & Baseline  & 105.15 & $\pm$ 15.85  & (4.0)  & 0.90 & $\pm$ 0.00  & (5.0)  & 65.90 & $\pm$ 3.41  & (3.9)  \\
        & HistUS    & 98.23 & $\pm$ 7.60  & (2.9)  & 0.88 & $\pm$ 0.00  & (2.8)  & 63.20 & $\pm$ 1.28  & (3.0)  \\
        & HistOS    & \textbf{90.57} & $\pm$ \textbf{4.24}  & (1.2)  & 0.87 & $\pm$ 0.00  & (2.5)  & \textbf{59.55} & $\pm$ \textbf{1.73}  & (1.3)   \\
        & ChebyUS   & 119.31 & $\pm$ 12.29  & (5.0)  & \textbf{0.83} & $\pm$ \textbf{0.00}  & (1.0)  & 86.57 & $\pm$ 2.39  & (5.0)  \\
        & ChebyOS   & 92.93 & $\pm$ 6.44  & (1.9)  & 0.88 & $\pm$ 0.00  & (3.7)  & 60.27 & $\pm$ 2.38  & (1.8)   \\
        \bottomrule
    \end{tabular*}
\end{table}

In the following subsections, we conduct a detailed investigation into the influence of the parameters $\beta$ and $\alpha$ on this synthetic dataset. Specifically, we analyze their impact on the probability values and the oversampling rates calculated in the HistUS and HistOS algorithms.


    \subsubsection{Impact of parameter $\beta$ on the probability function}     \label{sec4:Impact_of_parameter_beta_on_the_probability_function}
\begin{figure}[!h]
\centering    
    \includegraphics[width=\textwidth]{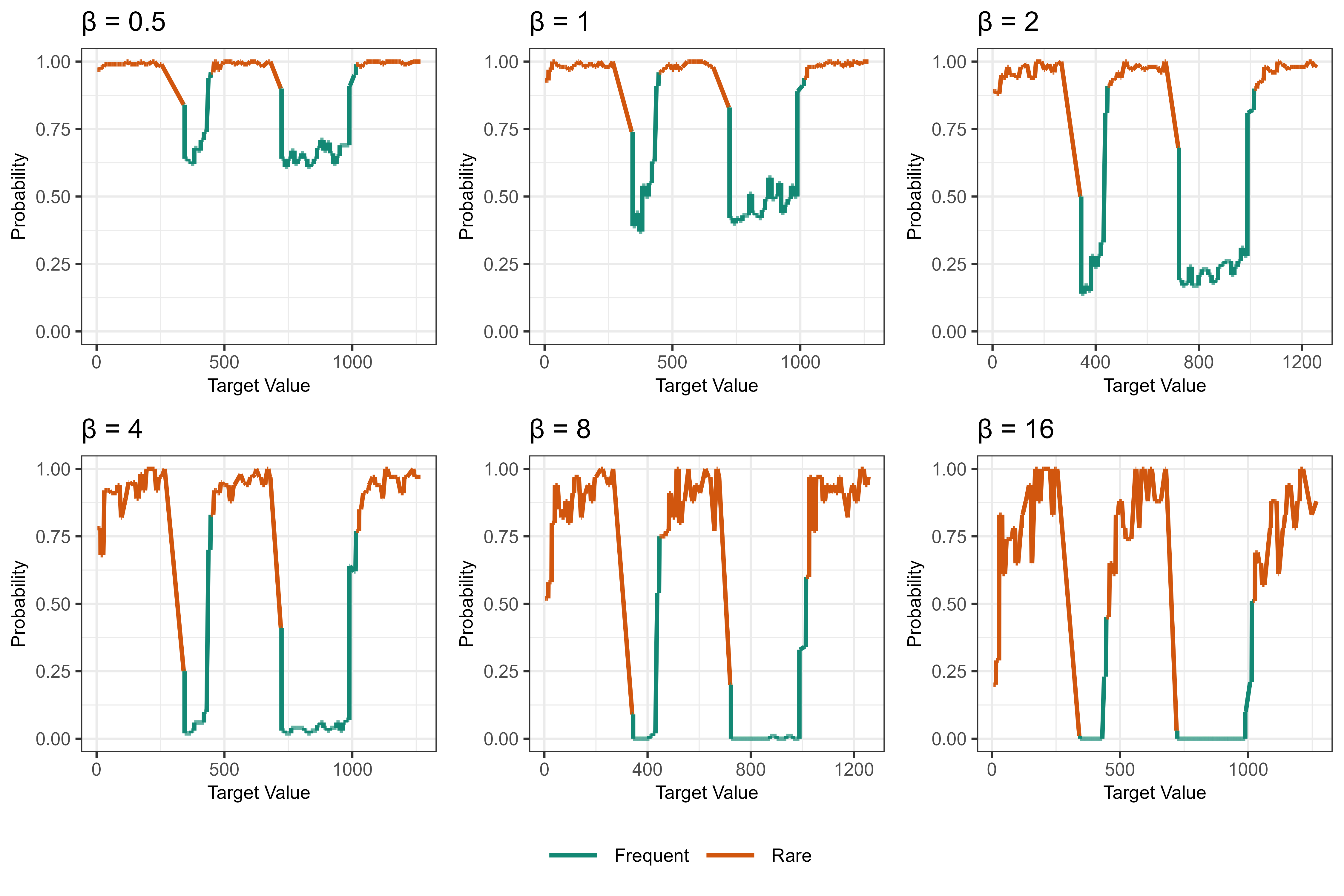}
    \caption{Influence of different values of the sampling decay parameter $\beta$ on the probability values obtained over the synthetic dataset.}
    \label{fig:betaImpact}
\end{figure}

    Figure~\ref{fig:betaImpact} illustrates the influence of different values of the parameter $\beta$ on the probability values obtained over our synthetic dataset. The parameter $\beta$ governs how frequent and rare cases are favored in the selection process. As shown in the figure, higher values of $\beta$ result in a steeper decline in probability, increasing the likelihood of selecting rare cases and reducing the selection of frequent cases. Conversely, lower values of $\beta$ flatten the probability curve, making the selection more uniform across different bins.

    \subsubsection{Impact of parameter $\alpha$ on the oversampling rate}
    \label{sec4:Impact_of_parameter_alpha_on_the_oversampling_rate}
\begin{figure}[!tb]
\centering    
    \includegraphics[width=.55\textwidth]{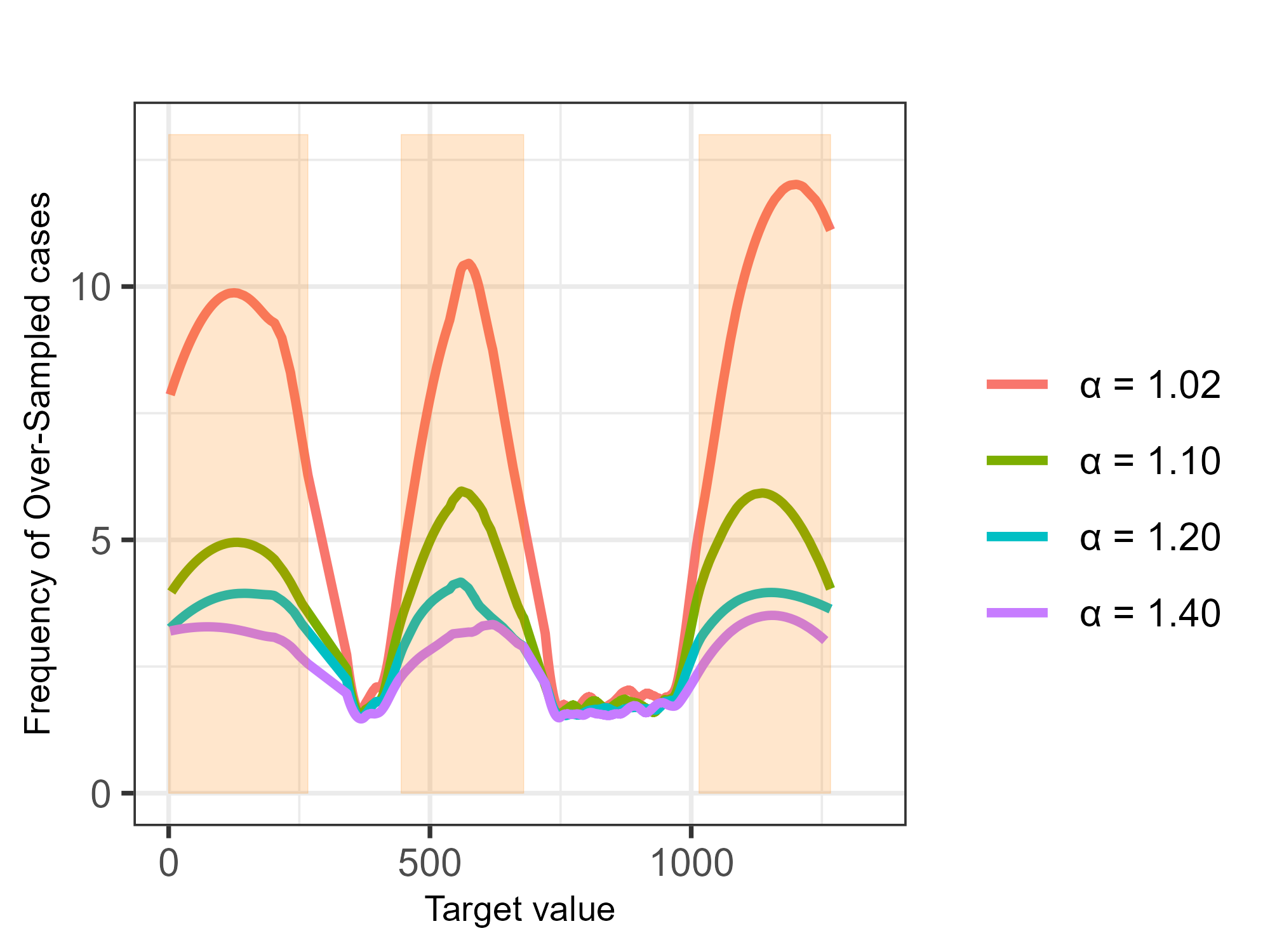}
    \caption{Effect of different values of the reduction coefficient parameter $\alpha$ on the oversampling rate. The parameter $\beta$ is fixed at $4$, while $\alpha$ is varied to analyze its impact on oversampling frequency across target values. The orange-highlighted areas in the background correspond to the rare regions in the target value domain.}
    \label{fig:alphaImpact}
\end{figure}

    Figure~\ref{fig:alphaImpact} illustrates the effect of different values of the parameter $\alpha$ on the frequency of oversampling across the target values of the generated synthetic dataset. The x-axis represents the target values, while the y-axis denotes the times each target value has been oversampled. The curves correspond to different $\alpha$ values, where lower $\alpha$ leads to a higher oversampling frequency. This trend highlights the role of $\alpha$ in controlling the balance between rare and frequent target values, with smaller $\alpha$ intensifying the oversampling of rare cases. It is worth mentioning that the numbers on the y-axis have been smoothed to enhance clarity, ensuring a more interpretable representation of the underlying trends.

    \subsection{Experiments on Benchmark Data}
    \label{sec4:experimental_settings}
    
\subsubsection{Experimental Setup}

Our experiments were conducted on twelve benchmark datasets sourced from the UCI repository~\citep{Dua19}. Our analysis designated examples with a relevance value exceeding $0.9$ as rare cases. It is important to note that no preprocessing step has been done on the data set, and all features of the data sets feed the learner models by their original form and value, ensuring the thoroughness and reliability of our experimental setup. Appendix~\ref{apdx:Additional_Dataset_Insights} provides key dataset statistics, including the number of attributes, examples, and the count and percentage of rare cases for each dataset as well as target value distributions and the relevance values assigned to each target value. The proposed methodologies were implemented in Kotlin 1.4.10 using MOA 
framework~\citep{MOA}, an open-source tool designed for mining and analyzing large datasets streamingly. We used R programming language - version 4.4.2 for error computation and result analysis.

In alignment with our previous chapter, we utilized four diverse regression models as base learners to alleviate potential algorithm-dependent biases that could introduce distortions to the results. These models included the perceptron~\citep{Block}, fast incremental model trees with drift detection (FIMT-DD)~\citep{ikonomovska2011learning}, TargetMean, and AMRules~\citep{DuarteGB16}. The TargetMean algorithm is a simple baseline for data stream regression. It incrementally updates the mean of the target variable using an online update rule, predicting the running mean for each incoming instance. The use of these diverse models is meant to assess the robustness of our study. In each experiment, we employed two instances of each learner with identical initial parameters. One instance was trained using our HistOS sampling method, while the other, referred to as the Baseline, underwent training with all examples in the dataset. 

To evaluate each model, we used the $RMSE$, $RMSE_\phi$~\citep{aminian2021chebyshev}, and $SERA$~\citep{RibeiroMoniz20} evaluation functions. For the $RMSE_\phi$, The threshold $thr_\phi$ is set to $0.9$ in our experiments. In the $SERA$ calculation, we normalize the computed errors relative to the total error when all data points are considered. This normalization is achieved by dividing each error value by the initial full-dataset error, ensuring that the largest error is approximately one and all others are expressed as fractions of this value. Thus, the $SERA$ scores become independent of absolute error scales, enabling meaningful comparisons between models and datasets.

Each experiment was conducted ten times, and the results were reported as averages and standard deviations. Throughout all experiments, we adhered to a prequential evaluation procedure~\citep{Gama13}, where models were continuously updated after processing each example. Specifically, each example was used for both testing and training. The evaluation involved initially assessing the model on the instance, followed by a single incremental learning step. 
Moreover, 15\% of examples in each dataset were allocated to the warming phase, 20\% for training purposes, and the remaining 65\% 
earmarked for model testing.
Regarding the parameter $\beta$ within the probability function, as previously highlighted, its value signifies a balance between the algorithm's capability to detect rare cases and its inclination to disregard frequent ones. We maintained a consistent value of $4$ for $\beta$ across all experiments after minor adjustments to our datasets. Nonetheless, optimizing this parameter could yield improved outcomes in specific scenarios, albeit these are not detailed here.
Similarly, for the parameter $\alpha$ in HistOS, we set its value to $1.02$ across all the experiments.    
        
    \subsubsection{Results}
    \label{sec4:results}
    In this section, we provide a concise overview of our results, while a comprehensive breakdown of all findings is shown in Tables~\ref{tab:detail_HistUS_RMSE_phi}-\ref{tab:detail_HistOS_RMSE} and Figures~\ref{fig:fig_CD_B_Phi}-\ref{fig:cd_cheby_hist} in Appendix~\ref{apdx:Additional_Results}. 

\begin{figure}[!h]
    \centering
    
    \begin{subfigure}{\textwidth}
        \centering
    \includegraphics[width=\textwidth]{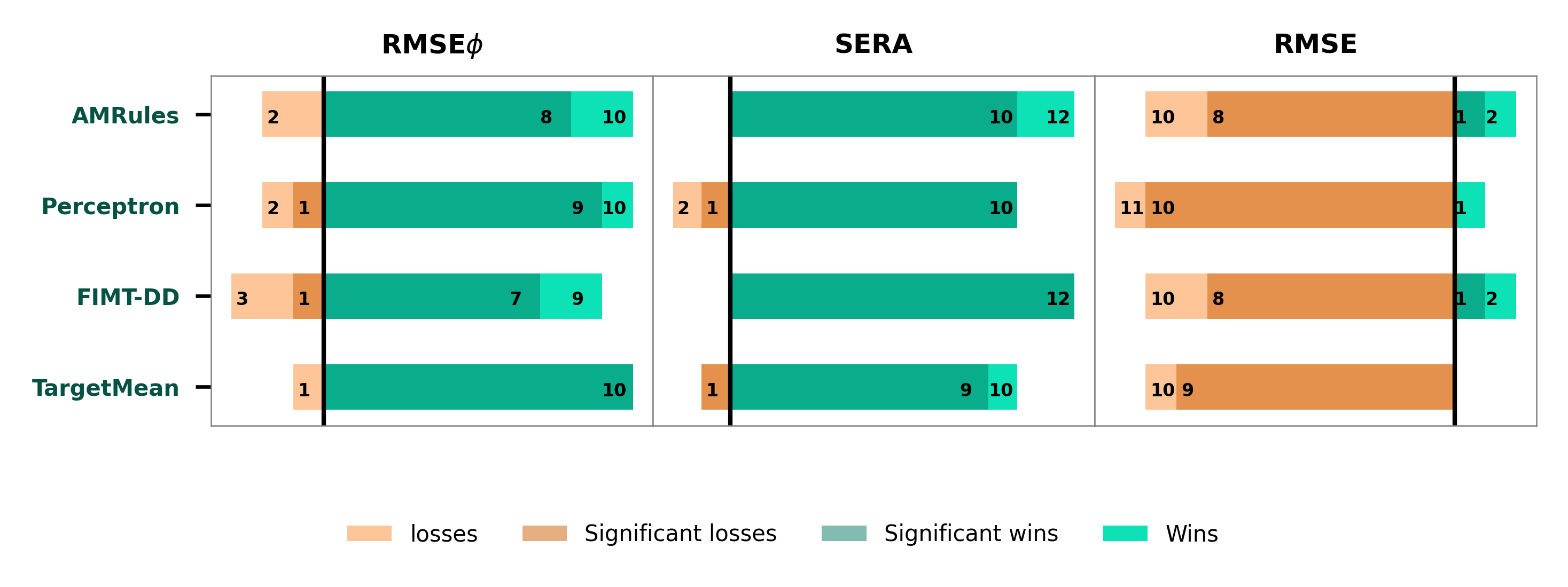}
    \caption{
    Baseline \emph{versus} HistUS}
    \label{fig:win_losts_bar_phi_HistUS}
    \end{subfigure}%
    
    \begin{subfigure}{\textwidth}
        \centering
    \includegraphics[width=\textwidth]
    {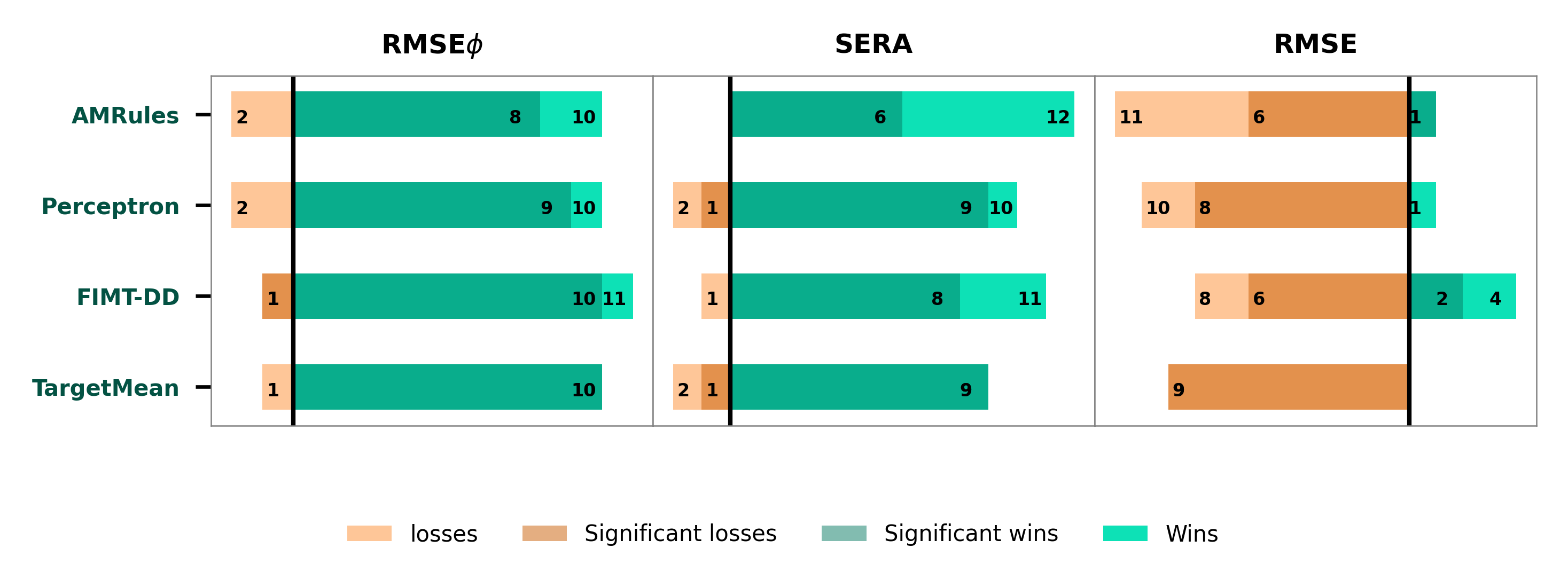}
    \caption{
    Baseline \emph{versus} HistOS}
    \label{fig:win_losts_bar_phi_HistUS}
    \end{subfigure}%

    \caption{
    Comparison of wins and losses, along with significant outcomes at a 95\% confidence level, derived from no sampling (Baseline) against histogram-based sampling (HistUS and HistOS) using $RMSE_{\phi}$, $SERA$, and $RMSE$.}
    \label{fig:winlost_HIST_Baseline_metrics}
\end{figure}

To enhance the clarity and interpretability of our results, we have transformed each table into a graphical representation. Figure~\ref{fig:winlost_HIST_Baseline_metrics} illustrate the wins and losses observed for each learner employing our sampling strategies compared to its corresponding baseline model. The significance of these outcomes, assessed at a 95\% confidence level using the Wilcoxon signed-rank test, is also depicted. The bar charts categorize the results into significant wins (substantially better performance), wins (better performance but not statistically significant), significant losses (substantially worse performance), and losses (worse performance but not statistically significant). Additionally, Critical Difference (CD) diagrams, employing the post-hoc Nemenyi test~\citep{demvsar2006statistical}, provide a comparative assessment of the learners' contributions across different result groups in the experiments in Figures~\ref{fig:fig_CD_B_Phi}-\ref{fig:cd_hist_rmse} in Appendix~\ref{apdx:Additional_Results}. 

\begin{table}[!bt]
    \centering
    \caption{
    Summary of results of histogram-based sampling strategy HistUS compared with no sampling (Baseline): average rank and number of significant wins (\#w) at 95\% confidence level, according to $RMSE_{\phi}$, $SERA$, and $RMSE$. 
    }
    \label{tab:Sum_BASELINE_HistUS_METRIC}

    \sisetup{table-format=1.2, detect-weight=true, detect-inline-weight=math}
    \begin{tabular*}{\textwidth}{@{\extracolsep{\fill}} ll S[table-format=1.2] @{\hspace{-1em}} r
        S[table-format=1.2] @{\hspace{-1em}} r
        S[table-format=1.2] @{\hspace{-1em}} r}
         \toprule
        \multirow{2}{*}{\textbf{Algorithm}} & \multirow{2}{*}{\textbf{Approach}} & \multicolumn{2}{c}{\textbf{$RMSE_{\phi}$}} & \multicolumn{2}{c}{\textbf{$SERA$}}  & \multicolumn{2}{c}{\textbf{$RMSE$}} \\
        \cmidrule{3-8}
        & &
        avg\_rank & \#w &
         avg\_rank & \#w &
         avg\_rank & \#w \\
        \midrule
        \multirow{2}{*}{\textbf{AMRules}}
        & Baseline & 3.83  & (0)    & 6.25 & (0) & \textbf{2.38} & (8)  \\
        & HistUS   & \textbf{2.42} & (8) & \textbf{3.58} & (10) & 3.83 & (1)  \\
        \midrule
        \multirow{2}{*}{\textbf{Perceptron}}
        & Baseline & 5.58 & (1) & 5.50 & (1) & \textbf{3.92} & (10)  \\
        & HistUS  & \textbf{3.83} & (9) & \textbf{2.75} & (10) & 5.50 & (0)  \\
        \midrule
        \multirow{2}{*}{\textbf{FIMT-DD}}
        & Baseline & 3.92 & (1) & 5.33 & (0) & \textbf{2.79} & (8)  \\
        & HistUS  & \textbf{2.96} & (7) & \textbf{3.0} & (12) & 4.67 & (1)  \\
        \midrule
        \multirow{2}{*}{\textbf{TargetMean}}
        & Baseline & 7.46 & (0) & 6.62 & (1) & \textbf{5.83} & (9)  \\
        & HistUS  & \textbf{6.00} & (10) & \textbf{2.96} & (9) & 7.08 & (0)  \\
        \bottomrule
    \end{tabular*}
\end{table}

\begin{table}[!bt]
    \centering
    \caption{
    Summary of results of histogram-based sampling strategy HistOS compared with no sampling (Baseline): average rank and number of significant wins (\#w) at 95\% confidence level, according to $RMSE_{\phi}$, $SERA$, and $RMSE$. 
    }
    \label{tab:Sum_BASELINE_HistOS_METRIC}
    \sisetup{table-format=1.2, detect-weight=true, detect-inline-weight=math}
    \begin{tabular*}{\textwidth}{@{\extracolsep{\fill}} ll
        S[table-format=1.2] @{\hspace{-1em}} r
        S[table-format=1.2] @{\hspace{-1em}} r
        S[table-format=1.2] @{\hspace{-1em}} r}
         \toprule
        \multirow{2}{*}{\textbf{Algorithm}} & \multirow{2}{*}{\textbf{Approach}} & \multicolumn{2}{c}{\textbf{$RMSE_{\phi}$}} & \multicolumn{2}{c}{\textbf{$SERA$}}  & \multicolumn{2}{c}{\textbf{$RMSE$}} \\
        \cmidrule{3-8}
        & &
        avg\_rank & \#w &
         avg\_rank & \#w &
         avg\_rank & \#w \\
        \midrule
        \multirow{2}{*}{\textbf{AMRules}}
        & Baseline & 4.25 & (0) & 5.92 & (0) & \textbf{2.29} & (6)  \\
        & HistOS  & \textbf{2.58} & (8) & \textbf{3.92} & (6) & 3.67 & (1)  \\
        \midrule
        \multirow{2}{*}{\textbf{Perceptron}}
        & Baseline & 5.50 & (0) & 5.42 & (1) & \textbf{4.21} & (8)  \\
        & HistOS  & \textbf{3.75} & (9) & \textbf{2.75} & (9) & 5.71 & (0)  \\
        \midrule
        \multirow{2}{*}{\textbf{FIMT-DD}}
        & Baseline & 4.33 & (1) & 5.17 & (0) & \textbf{2.88} & (6)  \\
        & HistOS  & \textbf{2.33} & (10) & \textbf{3.25} & (8) & 4.0 & (2)  \\
        \midrule
        \multirow{2}{*}{\textbf{TargetMean}}
        & Baseline & 7.38 & (0) & 6.46 & (1) & \textbf{6.04} & (9)  \\
        & HistOS  & \textbf{5.88} & (10) & \textbf{3.12} & (9) & 7.21 & (0)  \\
        \bottomrule
    \end{tabular*}
\end{table}

Tables~\ref{tab:Sum_BASELINE_HistUS_METRIC} and~\ref{tab:Sum_BASELINE_HistOS_METRIC} also summarize the experimental results, presenting the average ranks and the number of significant wins for various regression algorithms under three conditions: no sampling (baseline), histogram-based undersampling (HistUS), and histogram-based oversampling (HistOS). The results indicate that both HistUS and HistOS outperform the baseline in predicting rare cases, as reflected by their lower average ranks and higher numbers of significant wins in the $RMSE_{\phi}$ and $SERA$ metrics. This suggests that the proposed histogram-based sampling strategies are effective in addressing the challenges of imbalanced regression data.
Despite the improvements in rare case prediction, the baseline models tend to achieve better overall regression performance, as measured by the standard $RMSE$ metric. This suggests that while histogram-based sampling improves performance in prediction of rare cases, it may compromise overall regression accuracy.

Overall, these results highlight the effectiveness of the proposed histogram-based sampling strategies, HistUS and HistOS, particularly in addressing the online imbalanced regression problem and improving predictions for rare cases, though at the potential cost of overall regression performance.

\subsection{Comparative Analysis}
This section presents a comprehensive comparative analysis of the proposed histogram-based sampling strategies (HistUS and HistOS) against each other and their Chebyshev-based counterparts (ChebyUS and ChebyOS). The evaluation focuses on three key performance metrics —\textit{RMSE}$_{\phi}$, \textit{SERA}, and traditional \textit{RMSE}— to assess their effectiveness in rare case prediction and overall regression performance. We first compare HistUS and HistOS to understand their relative strengths in balancing rare case emphasis with model generalization. Subsequently, we contrast the histogram-based methods with the earlier Chebyshev-based approaches, analyzing their predictive outcomes across diverse regression algorithms, including AMRules, Perceptron, FIMT-DD, and TargetMean.
\subsubsection{Comparison between HistUS and HistOS methods}
\begin{figure}[tb]
    \centering
        \includegraphics[width=\textwidth]{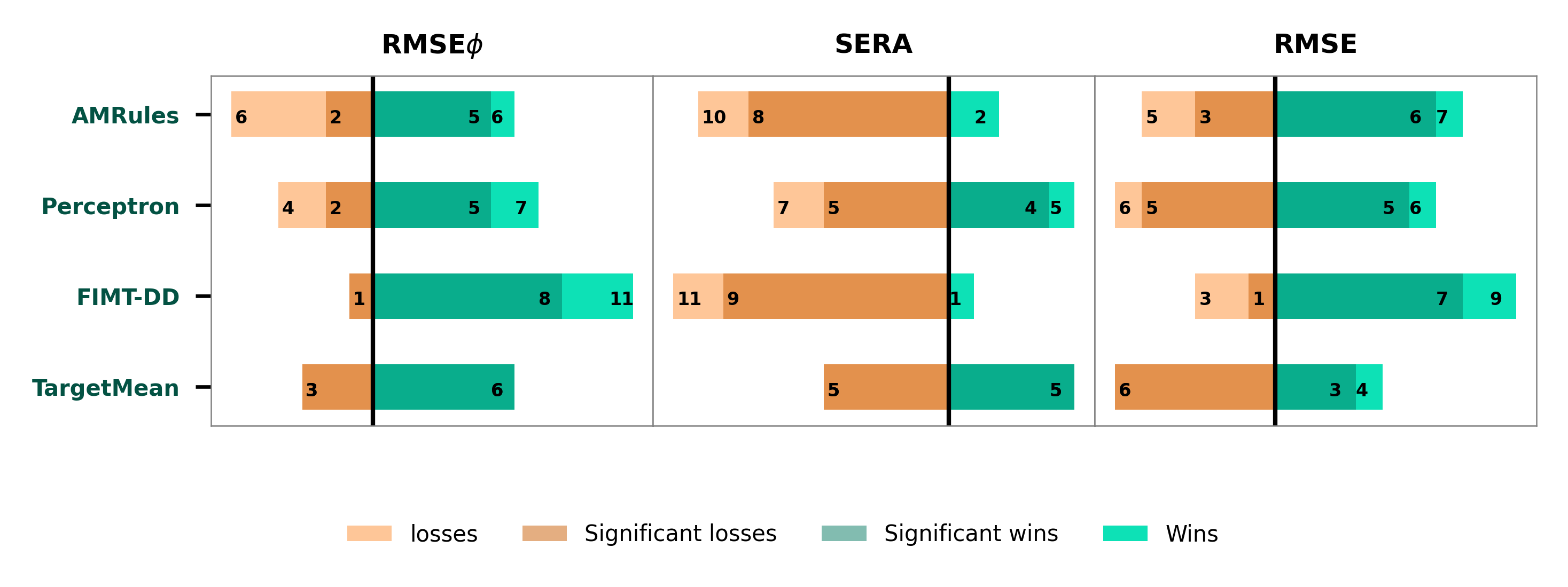}
    \caption{
    Comparison of wins and losses, along with significant outcomes at a 95\% confidence level, derived from histogram-based oversampling (HistOS) \emph{versus} histogram-based undersampling (HistUS) using $RMSE_{\phi}$, $SERA$, and $RMSE$.}
    \label{fig:win_losts_HistUS_OS}
\end{figure}

The results of the comparison, summarized in Figure~\ref{fig:win_losts_HistUS_OS} and Table~\ref{tab:Sum_HistUS_OS_METRIC}, provide insights into the effectiveness of these strategies in enhancing predictive performance for rare cases. The analysis of $RMSE_{\phi}$ demonstrates that HistOS consistently outperforms HistUS in rare case prediction, as indicated by its lower average ranks and a higher number of significant wins across all four regression algorithms. Notably, HistOS significantly improves FIMT-DD, securing the highest number of wins. On the other hand, HistUS demonstrates superior performance based on the $SERA$ metric, particularly for AMRules, Perceptron, and FIMT-DD. FIMT-DD achieves the highest number of significant wins (9) with HistUS, followed closely by AMRules with 8 wins, indicating that HistUS is more effective in optimizing $SERA$ for these models. However, for TargetMean, the choice between HistUS and HistOS has little impact, as their average ranks and significant wins remain similar.

Conversely, when evaluating overall regression performance using $RMSE$, the performance gap between HistUS and HistOS varies across models. HistOS again demonstrates strong results for FIMT-DD and AMRules while achieves a more balanced performance across Perceptron. However, the results for TargetMean indicate that HistOS may lead to performance degradation in $RMSE$, evidenced by its higher rank and fewer significant wins. The details of the results are presented in Tables~\ref{tab:detail_HistUS_HistOS_RMSE_phi} to~\ref{tab:detail_HistUS_HistOS_RMSE} and Figure~\ref{fig:cd_histUS_OS} in Appendix~\ref{apdx:Additional_Results}.

Overall, the results highlight a clear trade-off: while HistOS generally improves rare case prediction and shows greater significance in $RMSE_{\phi}$, HistUS performs better in terms of $SERA$. Meanwhile, the impact on $RMSE$ varies depending on the model.

\begin{table}[!h]
    \centering
    \caption{Comparison of histogram-based sampling strategies: average rank and number of significant wins (\#w) at a 95\% confidence level, according to $RMSE_{\phi}$, $SERA$, and $RMSE$. 
    }
    \label{tab:Sum_HistUS_OS_METRIC}
     \sisetup{table-format=1.2, detect-weight=true, detect-inline-weight=math}
    \begin{tabular*}{\textwidth}{@{\extracolsep{\fill}} ll 
        S[table-format=1.2] @{\hspace{-1em}} r 
        S[table-format=1.2] @{\hspace{-1em}} r 
        S[table-format=1.2] @{\hspace{-1em}} r}
        \toprule
        \multirow{2}{*}{\textbf{Algorithm}} & \multirow{2}{*}{\textbf{Approach}} & \multicolumn{2}{c}{\textbf{$RMSE_{\phi}$}} & \multicolumn{2}{c}{\textbf{$SERA$}}  & \multicolumn{2}{c}{\textbf{$RMSE$}} \\
        \cmidrule{3-8} 
        & &  
        avg\_rank & \#w &
         avg\_rank & \#w &
         avg\_rank & \#w \\
        \midrule
        \multirow{2}{*}{\textbf{AMRules}} 
        & HistUS  & 3.25 & (2) & \textbf{4.75} & (8) & 3.25 & (3)  \\
        & HistOS  & \textbf{3.17} & (5) & 6.42 & (0) & \textbf{2.75} & (6)  \\
        \midrule
        \multirow{2}{*}{\textbf{Perceptron}} 
        & HistUS  & 4.71 & (2) & \textbf{3.75} & (5) & 4.83 & (5)  \\
        & HistOS  & \textbf{4.38} & (5) & 4.33 & (4) & \textbf{4.50} & (5)  \\
        \midrule
        \multirow{2}{*}{\textbf{FIMT-DD}} 
        & HistUS  & 4.42 & (1) & \textbf{3.75} & (9) & 4.33 & (1)  \\
        & HistOS  & \textbf{2.83} & (8) & 5.67 & (1) & \textbf{3.25} & (7)  \\
        \midrule
        \multirow{2}{*}{\textbf{TargetMean}} 
        & HistUS  & 6.83 & (3) & 3.67 & (5) & \textbf{6.50} & (6)  \\
        & HistOS  & \textbf{6.42} & (6) & 3.67 & (5) & 6.58 & (3)  \\
        \bottomrule
    \end{tabular*}
\end{table}

    \subsubsection{Comparing Histogram and Chebyshev-based approaches}
    \label{cheby_hist_comparision}

\begin{figure}[tb]
    \centering
    \begin{subfigure}{0.9\textwidth}
        \centering    
        \includegraphics[width=\textwidth]{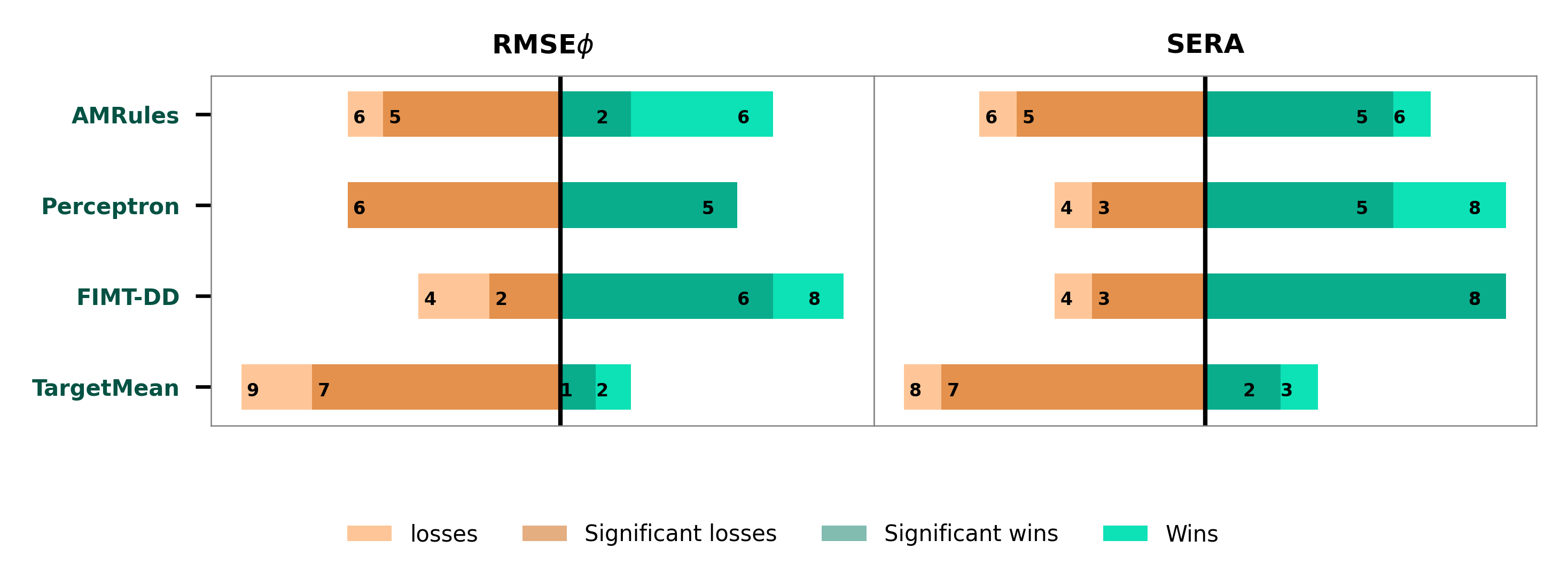}
        \caption{
        ChebyUS \emph{vs} HistUS
        }
        \label{fig:winLostbars_ChebyHist__RMSE_phi}
    \end{subfigure}


    \begin{subfigure}{0.9\textwidth}
        \centering    
        \includegraphics[width=\textwidth]{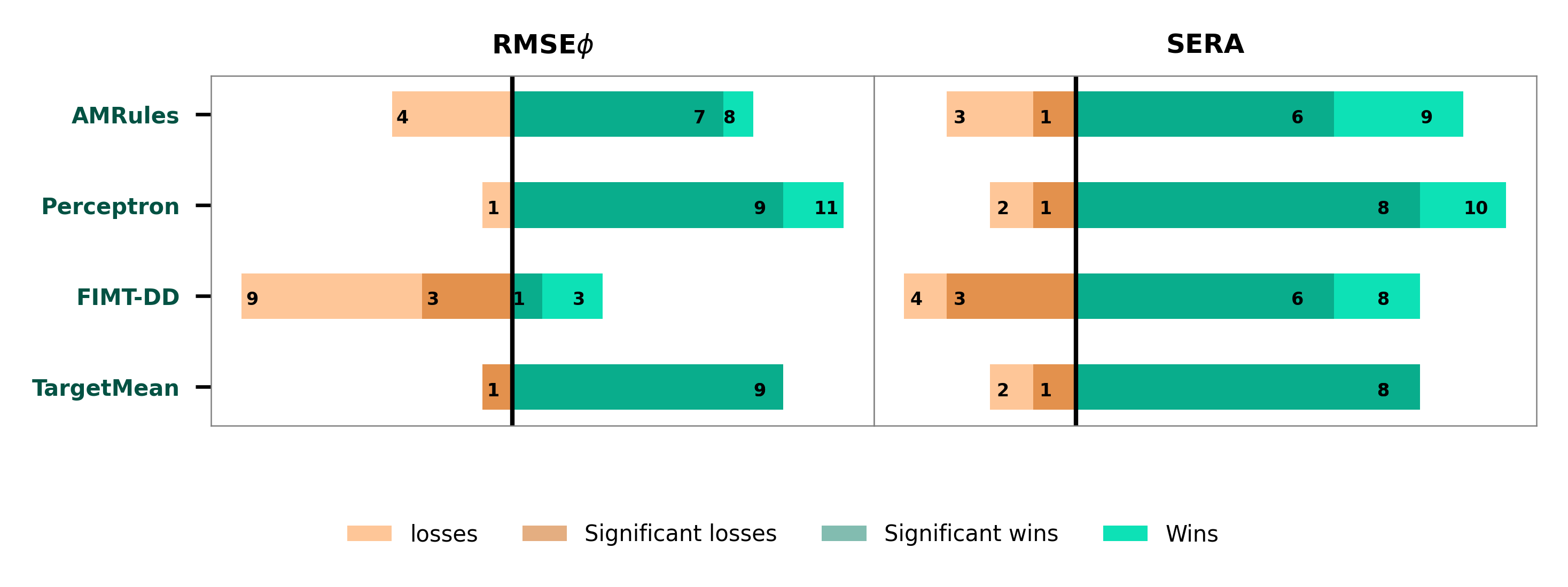}
        \caption{ 
        ChebyOS \emph{vs} HistOS
        }
        \label{fig:winLostbars_ChebyHist__SERA}
    \end{subfigure}%
    
    \caption{
     Comparison of wins and losses, along with significant outcomes at a 95\% confidence level, derived from Chebyshev-based sampling  \emph{versus} histogram-based undersampling using $RMSE_{\phi}$, $SERA$.
    }
    \label{fig:winlost_ChebyHist}
\end{figure}

This section presents a comparative analysis of histogram-based and Chebyshev-based sampling strategies in both undersampling (ChebyUS, HistUS) and oversampling (ChebyOS, HistOS) scenarios. The evaluation is conducted using $RMSE_{\phi}$ and $SERA$ metrics across four regression algorithms to assess the effectiveness of each approach in addressing rare case predictions. To ensure a fair comparison, all datasets are designed with rare cases positioned exclusively in the extreme tails of the target value distribution, as Chebyshev-based methods can only detect rare regions in these areas. The results, presented in Figure~\ref{fig:winlost_ChebyHist} and Tables~\ref{tab:Sum_ChebyUS_HistUS_METRIC} and~\ref{tab:Sum_ChebyOS_HistOS_METRIC}, highlight how effectively each method addresses rare case predictions.

In the undersampling scenario, ChebyUS generally outperforms HistUS in terms of $RMSE_{\phi}$, achieving lower average ranks and a higher number of significant wins across most regression algorithms. Specifically, ChebyUS demonstrates superior performance for AMRules (5 wins), Perceptron (6 wins), and TargetMean (7 wins) when compared to HistUS. An exception is observed in the case of FIMT-DD, where HistUS achieves a better results for $RMSE_{\phi}$ (6 wins vs. 2 for ChebyUS). These findings suggest that Chebyshev-based undersampling is more effective in reducing prediction errors for rare cases. Regarding $SERA$, the performance of ChebyUS and HistUS varies across models. While ChebyUS exhibits better results for AMRules and TargetMean, HistUS demonstrates superior performance for FIMT-DD and Perceptron.

In the oversampling scenario, HistOS exhibits superior performance across both $RMSE_{\phi}$ and $SERA$ in almost all cases. The only exception is observed in the comparison for FIMT-DD using $RMSE_{\phi}$, where ChebyOS achieves slightly better results than HistOS.

Overall, the findings indicate that Chebyshev-based undersampling (ChebyUS) is particularly effective for rare case prediction, consistently outperforming the histogram-based method in terms of $RMSE_{\phi}$. However, the effectiveness based on $SERA$ is model-dependent, with HistOS demonstrating competitive results in certain cases. In oversampling settings, the histogram-based method (HistOS) proves to be more efficient in rare case prediction across both evaluation metrics.

\begin{table}[!htb]
    \centering
    \caption{Comparison of
    Chebyshev-based and histogram-based Undersampling strategies: average rank and number of significant wins (\#w) at 95\% confidence level, according to $RMSE_{\phi}$ and $SERA$. 
    }
    \label{tab:Sum_ChebyUS_HistUS_METRIC}
    \sisetup{table-format=1.2, detect-weight=true, detect-inline-weight=math}
    \begin{tabular*}{\textwidth}{@{\extracolsep{\fill}} ll
        S[table-format=1.2] @{\hspace{-1em}} r
        S[table-format=1.2] @{\hspace{-1em}} r}
        \toprule
        \multirow{2}{*}{\textbf{Algorithm}} & \multirow{2}{*}{\textbf{Approach}} & \multicolumn{2}{c}{\textbf{$RMSE_{\phi}$}} & \multicolumn{2}{c}{\textbf{$SERA$}} \\
        \cmidrule{3-6}
        & &
        avg\_rank & \#w &
         avg\_rank & \#w \\
        \midrule
        \multirow{2}{*}{\textbf{AMRules}}
        & ChebyUS  & \textbf{2.83} & (5)  & \textbf{4.92} & (5) \\
        & HistUS   & 3.08 & (2)  & 5.42 & (5)  \\
        \midrule
        \multirow{2}{*}{\textbf{Perceptron}}
        & ChebyUS  & \textbf{4.21} & (6) & 4.42 & (3)  \\
        & HistUS   & 4.46 & (5)  & 4.42 & (5)  \\
        \midrule
        \multirow{2}{*}{\textbf{FIMT-DD}}
        & ChebyUS  & 4.71 & (2)  & 4.83 & (3)  \\
        & HistUS   & \textbf{3.79} & (6)  & \textbf{4.50} & (8)  \\
        \midrule
        \multirow{2}{*}{\textbf{TargetMean}}
        & ChebyUS  & \textbf{5.92} & (7)  & \textbf{3.29} & (7) \\
        & HistUS   & 7.00 & (1)  & 4.21 & (2)  \\
        \bottomrule
    \end{tabular*}
\end{table}

\begin{table}[!htb]
    \centering
    \caption{Comparison of
    Chebyshev-based and histogram-based oversampling strategies: average rank and number of significant wins (\#w) at 95\% confidence level, according to $RMSE_{\phi}$ and $SERA$. 
    }
    \label{tab:Sum_ChebyOS_HistOS_METRIC}
    \sisetup{table-format=1.2, detect-weight=true, detect-inline-weight=math}
    \begin{tabular*}{\textwidth}{@{\extracolsep{\fill}} ll
        S[table-format=1.2] @{\hspace{-1em}} r
        S[table-format=1.2] @{\hspace{-1em}} r}
        \toprule
        \multirow{2}{*}{\textbf{Algorithm}} & \multirow{2}{*}{\textbf{Approach}} & \multicolumn{2}{c}{\textbf{$RMSE_{\phi}$}} & \multicolumn{2}{c}{\textbf{$SERA$}} \\
        \cmidrule{3-6}
        & &
        avg\_rank & \#w &
         avg\_rank & \#w \\
        \midrule
        \multirow{2}{*}{\textbf{AMRules}}
        & ChebyOS  & 3.50 & (0)  & 6.00 & (1)  \\
        & HistOS   & \textbf{3.00} & (7)  & \textbf{4.50} & (6)  \\
        \midrule
        \multirow{2}{*}{\textbf{Perceptron}}
        & ChebyOS  & 5.42 & (0)  & 5.08 & (1)  \\
        & HistOS   & \textbf{4.33} & (9)  & \textbf{3.08} & (8)  \\
        \midrule
        \multirow{2}{*}{\textbf{FIMT-DD}}
        & ChebyOS  & \textbf{2.92} & (3)  & 5.08 & (3)  \\
        & HistOS   & 3.50 & (1)  & \textbf{3.58} & (6)  \\
        \midrule
        \multirow{2}{*}{\textbf{TargetMean}}
        & ChebyOS  & 7.25 & (1)  & 5.33 & (1)  \\
        & HistOS   & \textbf{6.08} & (9)  & \textbf{3.33} & (8)  \\
        \bottomrule
    \end{tabular*}
\end{table}

\section{Conclusion and Future work} 
\label{sec4:Conclusion_and_Future_work}
In this paper, we tackled the challenge of learning from imbalanced regression data streams, where rare cases can occur at any point in the distribution. We introduced two novel sampling strategies, Histogram-based Undersampling (HistUS) and Histogram-based Oversampling (HistOs), designed to dynamically detect and handle rare cases using an online histogram-based heuristic. These strategies were developed to overcome a critical limitation of prior approaches, ChebyUS and ChebyOS, which assume that rare cases are confined to the distribution tails. By leveraging histogram-based rarity detection, our methods adaptively adjust the data sampling process, ensuring that rare instances are effectively recognized and utilized for model training.

The experimental evaluation, conducted on a diverse set of one synthetic and 13 real-world datasets, demonstrated that the proposed approaches consistently outperform baseline models in predictive performance. The results indicate that HistUS and HistOS reduce prediction errors for rare cases and improve the overall model generalization. Furthermore, our study highlights the role of two key parameters, $\beta$ and $\alpha$, which influence the probability estimation and oversampling rate in our histogram-based methods. A deeper exploration of their impact revealed that tuning these parameters can significantly enhance the effectiveness of rare case detection and model training.

Despite these promising results, several areas remain open for future research. One critical direction is to extend our methods to explicitly handle concept drift, a common challenge in evolving data streams. While the online histogram inherently adapts to changes in data distribution, a more formalized drift detection mechanism could further improve performance under non-stationary conditions. Another important avenue is the investigation of adaptive parameter tuning strategies, where $\beta$ and $\alpha$ could be dynamically optimized during the learning process rather than being set empirically.

Additionally, while our current framework was tested on standard regression datasets, its applicability to real-world high-dimensional and multi-target regression problems remains an open question. Future work could explore modifications to our approach that allow efficient handling of multi-output regression tasks. Moreover, integrating our sampling strategies with advanced ensemble learning techniques could further enhance robustness against extreme cases and concept drift.

In conclusion, our proposed histogram-based sampling approaches effectively tackle imbalanced regression data streams, offering substantial improvements over traditional Chebyshev-based methods. Our method's ability to detect and learn from rare instances throughout the entire distribution — rather than just at the tails — makes it a valuable tool for real-time data stream applications. Future research will focus on extending these methods to more complex streaming environments, further enhancing their adaptability and integration into broader machine-learning frameworks.

\bibliography{sn-bibliography}

\clearpage
\appendix
\section{Synthetic Data Generation
}
\label{apdx:Synthetic_Data_Structure}
The dataset consists of one independent variable \( x \) and one dependent variable \( y \), with values of \( y \) constrained within the range \([1,1000]\). To introduce imbalanced data distributions, 95\% of the instances were designated as frequent, appearing within two specific intervals, \([150,350]\) and \([650,850]\). At the same time, the remaining 5\% were classified as rare and distributed across the lower tail \([0,150]\), the middle range \([350,650]\), and the upper tail \([850,1000]\). Different functional relationships were employed for each segment to ensure diverse patterns in the data. The frequent instances followed quadratic relationships, whereas the rare instances were modeled with cubic transformations to introduce non-linearity. The values of \( x \) were sampled from uniform distributions tailored to their respective regions, ensuring controlled variability across the dataset.

The functions governing the relationship between the independent variable \( x \) and the dependent variable \( y \) are defined as follows. For frequent instances, the relationship is modeled using a quadratic function:

\begin{equation}
    \label{eq:func_freq}
    y = 0.001 x^2 + c
\end{equation}

\noindent where \( c \) is a constant ensuring continuity within the specified frequent intervals. 

In contrast, rare instances are modeled using cubic transformations to introduce non-linearity through the function:

\begin{equation}
    \label{eq:func_rare_tails}
    y = 0.0001 x^3
\end{equation}

\noindent for rare instances in the lower tail, and through the function: 

\begin{equation}
    \label{eq:func_rare_middle}
    y = 0.00001 (x - 350)^3 + c
\end{equation}

\noindent for rare instances in the middle interval, ensuring a clear distinction from frequent cases. 

\section{Range Queries Aggregates Dataset}
\label{apdx:Range-Queries-Aggregates}
\begin{figure}[t]
\centering    
    \includegraphics[width=0.6\textwidth]{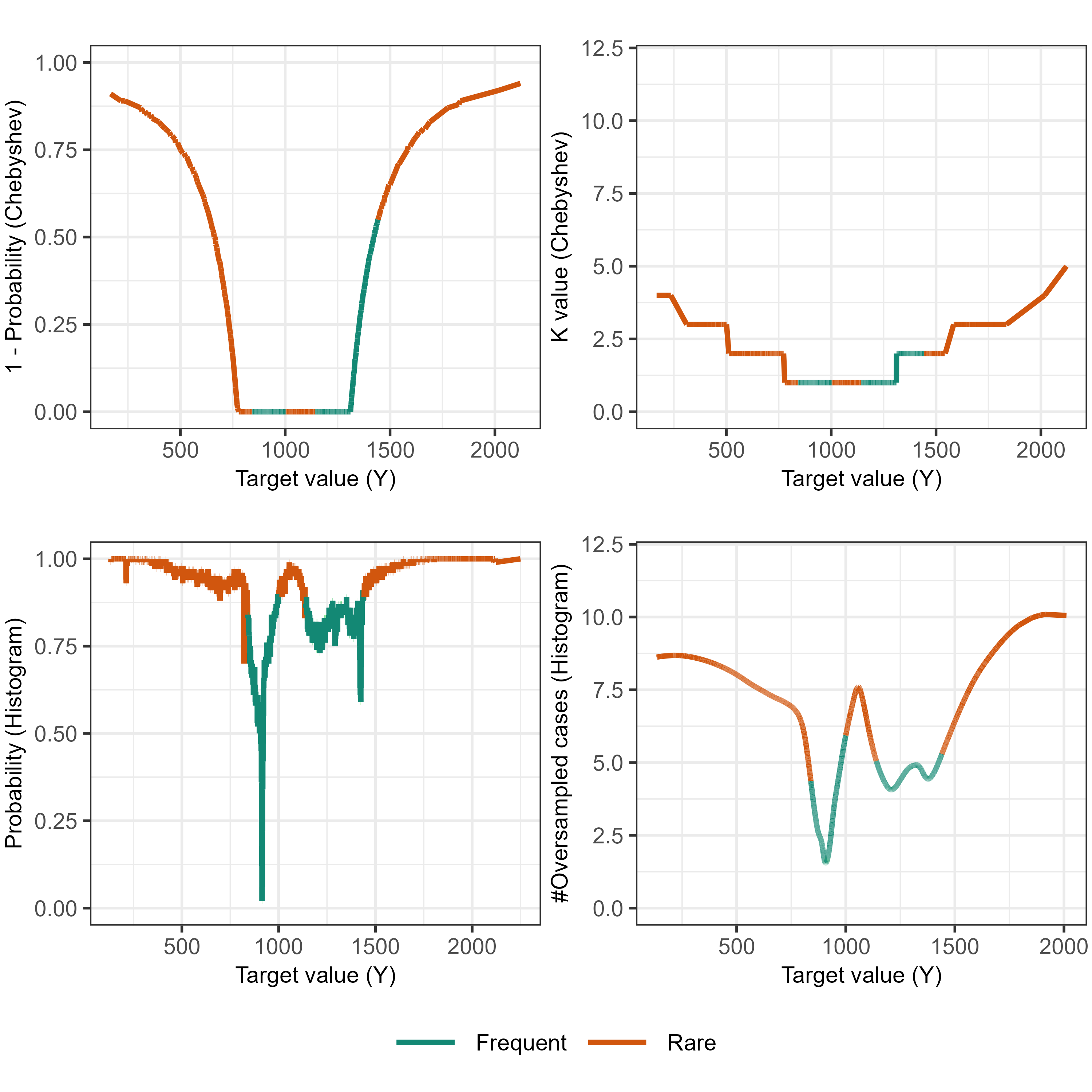}
    \caption{Comparison of Probability and \( K \) values between the Chebyshev-based and the histogram-based approaches for the synthetic dataset.}
    \label{fig:Range_dataset_prob_relevancy}
\end{figure}

\begin{figure}[!h]
    \centering
      \begin{subfigure}{.48\textwidth}
        \centering    \includegraphics[width=.9\textwidth]{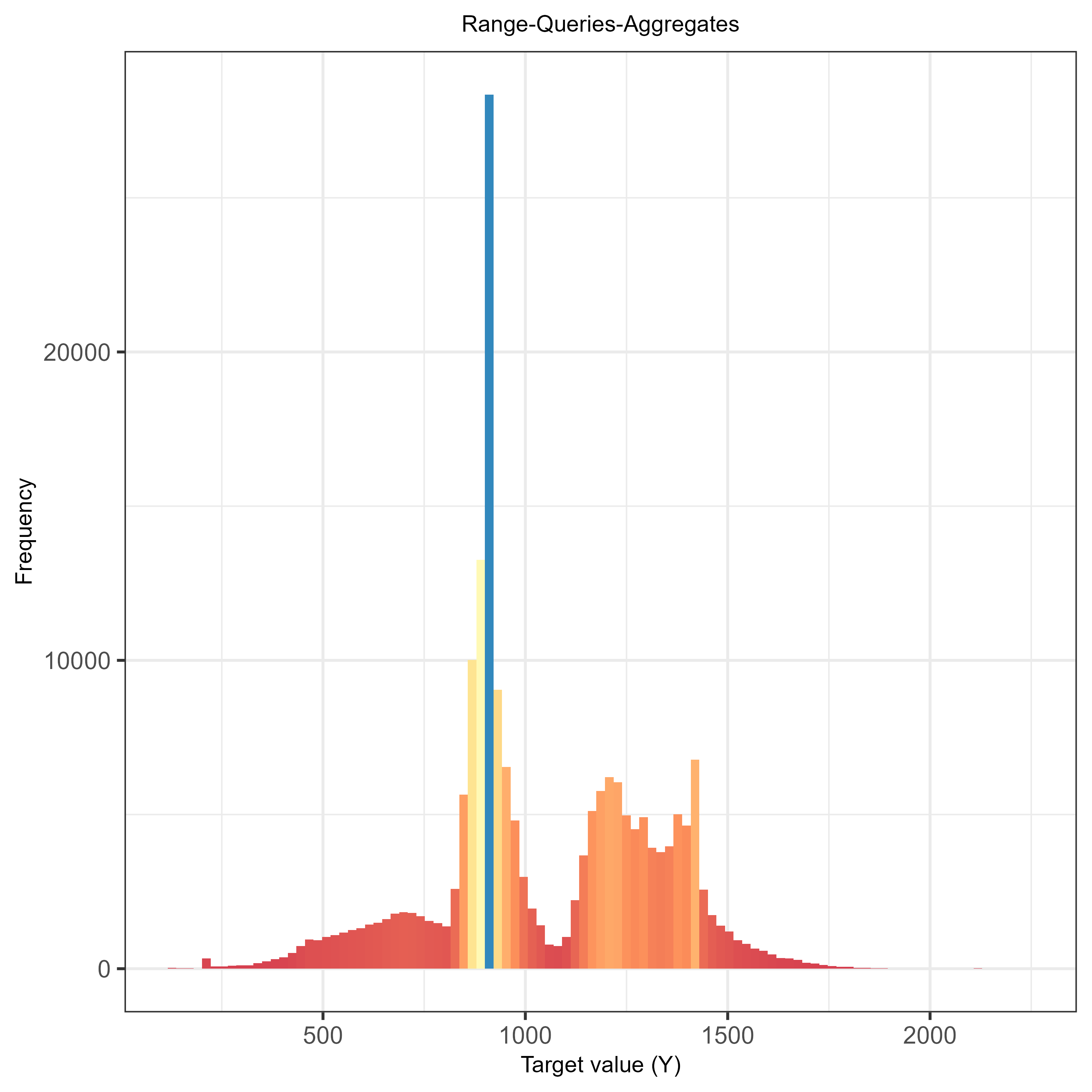}
        \caption{}
        \label{fig:6_Range_dataset_histogram}
    \end{subfigure}
    \begin{subfigure}{.48\textwidth}
        \centering    \includegraphics[width=.9\textwidth]{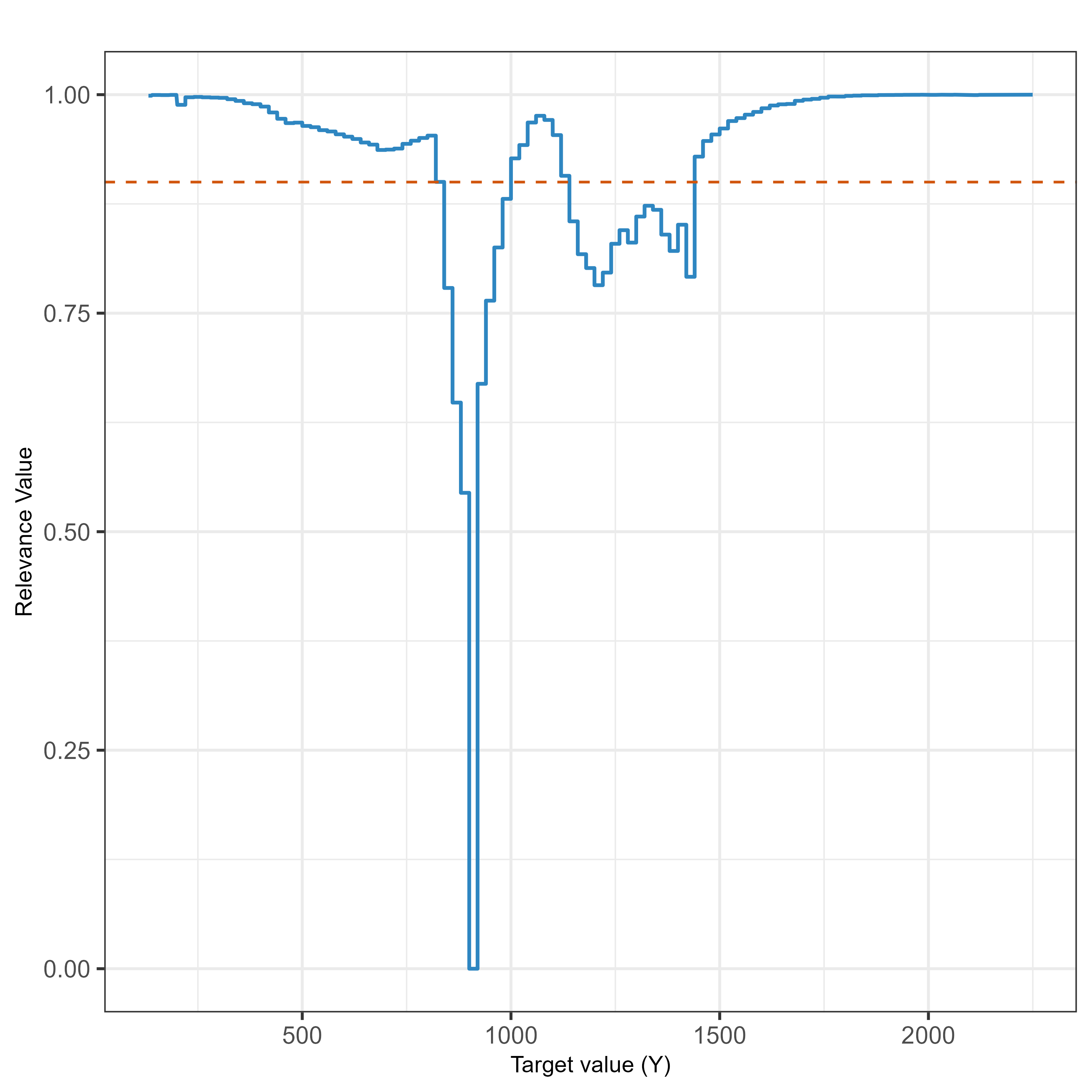}
        \caption{}
    \label{fig:6_Range_dataset_relevancy}
    \end{subfigure}%
    \hfill

    \caption{Histogram of target values (a) and respective relevance values (b) of common and rare cases considering $thr_\phi$ of 0.9 for "Range Queries Aggregates" dataset}\label{fig:sythetic_data_entire_dataset_histogram}
\end{figure}

The \emph{Range Queries Aggregates} dataset is a subset of the Query Analytics Workloads Dataset from the UCI Machine Learning Repository. It consists of synthetic range query workloads generated using Gaussian distributions applied to a real dataset. Each query represents a rectangular region in a 2D space, defined by coordinates (X, Y) and corresponding ranges (X-range, Y-range). For each query, the dataset provides aggregate scalar values, including count, sum, and average. To highlight the limitations of Chebyshev-based approaches for this dataset, Figure \ref{fig:Range_dataset_prob_relevancy} presents the derived values. The top two plots show the inverse Chebyshev probability (\(1 - \text{Probability}\)) and the \(K\) values. The inverse probability determines instance selection in ChebyUS, emphasizing rare cases with higher values. Meanwhile, \(K\) controls instance replication in ChebyOS, where larger values lead to increased oversampling of rare instances. The bottom plots in the figure depict the probability values and the number of oversampled instances generated by histogram-based approaches. Figures~\ref{fig:6_Range_dataset_histogram} and~\ref{fig:6_Range_dataset_relevancy} show the histogram and the relevance values assigned to target values, respectively. The dashed orange line indicates the threshold (\(thr_\phi\)) for distinguishing rare cases.

\clearpage
\section{Benchmark Datasets}
\label{apdx:Additional_Dataset_Insights}
\begin{figure}[!h]
\centering    
    \includegraphics[width=\textwidth]{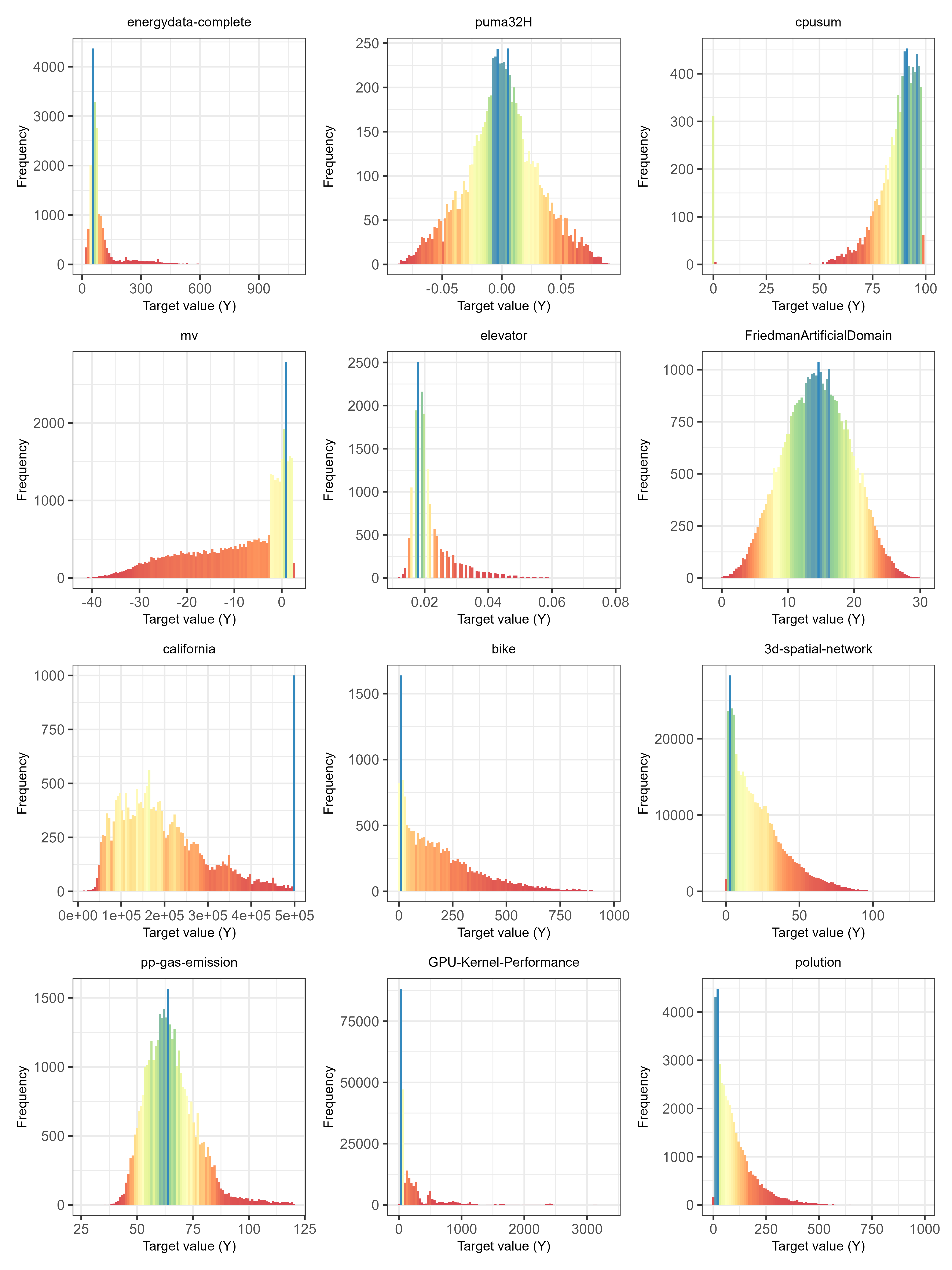}
    \caption{Histograms of the target values for all datasets}
    \label{fig:allhistogramdatsets}
\end{figure}

\begin{figure}[!h]
\centering    
    \includegraphics[width=\textwidth]{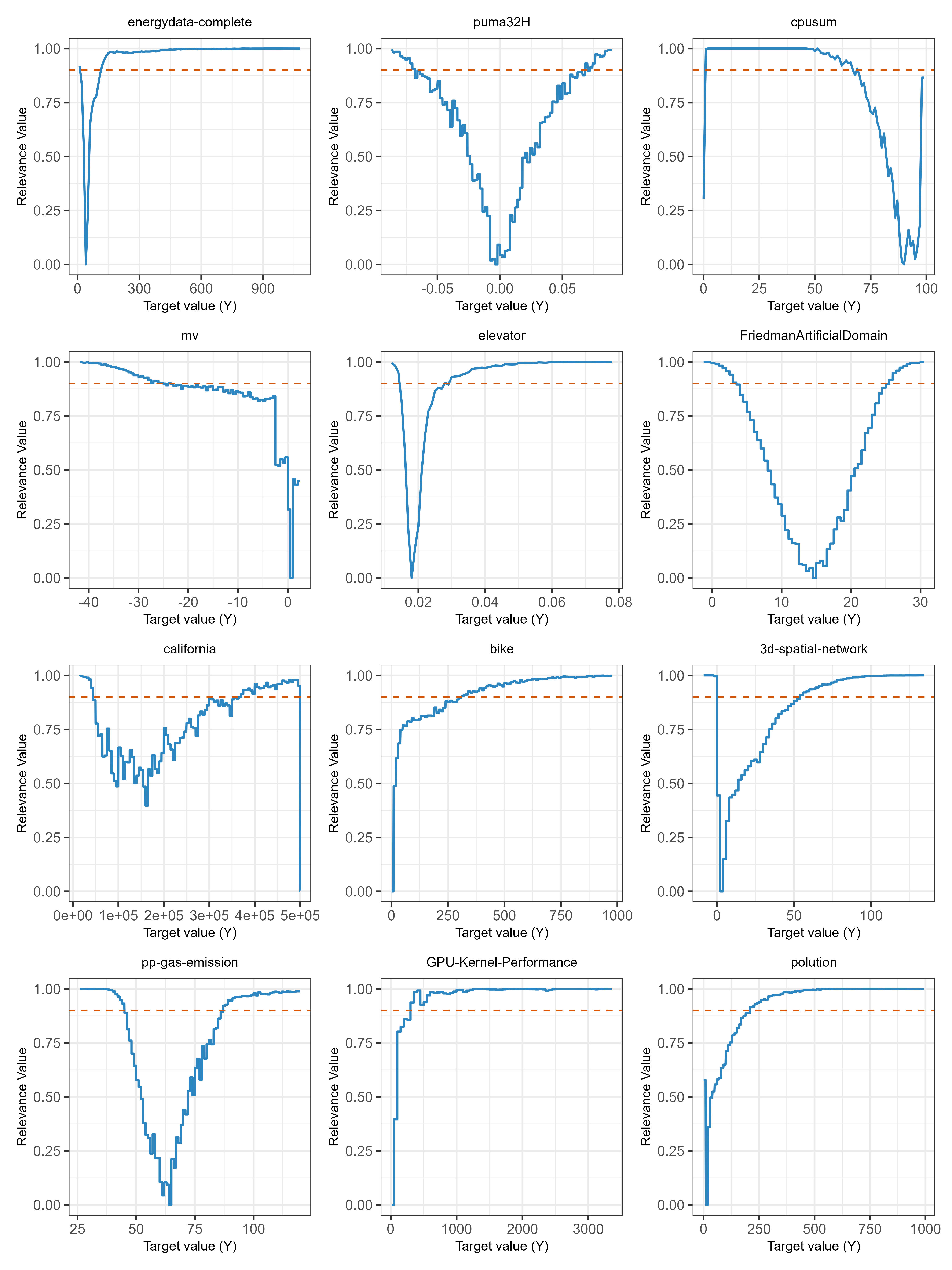}
    \caption{Relevance values of the target variable for all datasets. The dashed orange lines represent the threshold (\(thr_\phi= 0.9\)) used to distinguish rare cases.}
    \label{fig:allrelevancedatsets}
\end{figure}

Table~\ref{tab1} provides key dataset statistics, including the number of attributes, examples, and the count and percentage of rare cases for each dataset. Furthermore, 
To better understand the distribution of target variables and their corresponding relevance values in the datasets, Figures~\ref{fig:allhistogramdatsets} and~\ref{fig:allrelevancedatsets} display the histograms of the target value domains and the relevance values assigned to each target value, respectively.

\begin{table}[!htb]
\caption{Data sets characterization.}
\label{tab1}
\begin{tabular}
    {
    lrrr}
    \toprule
        \textbf{Data set} & \textbf{\#Attrs} & \textbf{\#Exs} & \textbf{\#Rare cases (\%)} \\
        \midrule
        3d-spatial-network & 5 & 434,873 & 31,255 (\%7.19) \\
        bike & 18 & 17,379 & 3,702 (\%21.3) \\
        california & 10 & 20,639 & 1,272 (\%6.16) \\
        cpusum & 14 & 8,192 & 366 (\%4.47) \\
        elevator & 20 & 16,599 & 2,172 (\%13.09) \\
        energydata-complete & 29 & 19,735 & 3,506 (\%17.77) \\
        FriedmanArtificialDomain & 12 & 40,768 & 807 (\%1.98) \\
        GPU-Kernel-Performance & 16 & 241,600 & 43,019 (\%17.81) \\
        mv & 12 & 40,768 & 4,582 (\%11.24) \\
        pollution & 13 & 41,757 & 4,693 (\%11.24) \\
        pp-gas-emission & 12 & 36,733 & 1,823 (\%4.96) \\
        puma32H & 34 & 8,192 & 211 (\%2.58) \\
    \bottomrule
\end{tabular}
\end{table}

\clearpage


\clearpage

\section{Additional Results}
\label{apdx:Additional_Results}
A comprehensive evaluation of the histogram-based sampling methods, HistUS and HistOS, was performed against their baseline models across 13 datasets over 10 repetitions. 
To ensure fair comparisons across datasets, the average values were \emph{normalized} by dividing each value by the maximum within its respective dataset.  

The results are summarized in Tables~\ref{tab:detail_HistUS_RMSE_phi} to~\ref{tab:detail_HistOS_RMSE}, covering multiple performance metrics. Tables~\ref{tab:detail_HistUS_RMSE_phi} and~\ref{tab:detail_HistOS_RMSE_phi} present the $RMSE_{\phi}$ evaluation for HistUS and HistOS, respectively, demonstrating their effectiveness in reducing predictive error. Similarly, Tables~\ref{tab:detail_HistUS_SERA} and~\ref{tab:detail_HistOS_SERA} report the $SERA$ values. The standard $RMSE$ results for both undersampling and oversampling strategies are detailed in Tables~\ref{tab:detail_HistUS_RMSE} and~\ref{tab:detail_HistOS_RMSE}, providing insights into overall predictive accuracy.  

\begin{figure}[!h]
    \centering
    \begin{subfigure}{.5\textwidth}
        \centering
        \includegraphics[width=.9\linewidth]{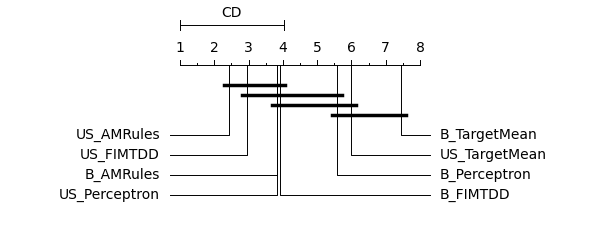}  
        \caption{Sampling Strategy: \texttt{HistUS}} 
        \label{fig:CD_HistUS_phi}
    \end{subfigure}%
    \begin{subfigure}{.5\textwidth}
        \centering
        \includegraphics[width=.9\linewidth]{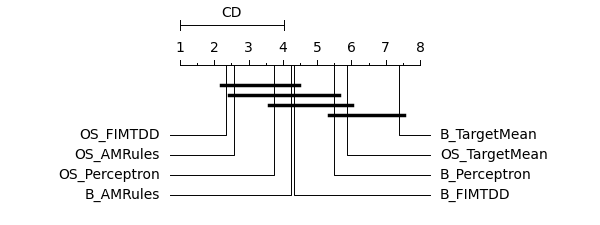}  
        \caption{Sampling Strategy: \texttt{HistOS}} 
        \label{fig:CD_HistOS_phi}
    \end{subfigure}
    \caption{
CD diagrams based on the obtained $RMSE_{\phi}$ results, illustrating the performance comparison between the  HistUS (US) and HistOS (OS) sampling strategies and their respective baselines (B).}
    \label{fig:fig_CD_B_Phi}
\end{figure}

\begin{figure}[!h]
    \centering
    \begin{subfigure}{.5\textwidth}
        \centering
        \includegraphics[width=.9\linewidth]{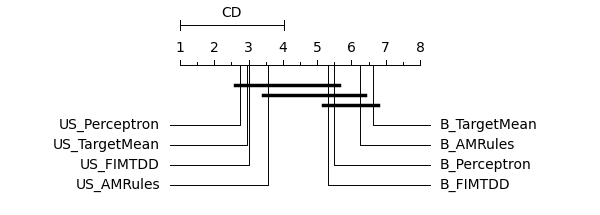}  
        \caption{Sampling Strategy: \texttt{HistUS}} 
        \label{fig:CD_HistUS_SERA}
    \end{subfigure}%
    \begin{subfigure}{.5\textwidth}
        \centering
        \includegraphics[width=.9\linewidth]{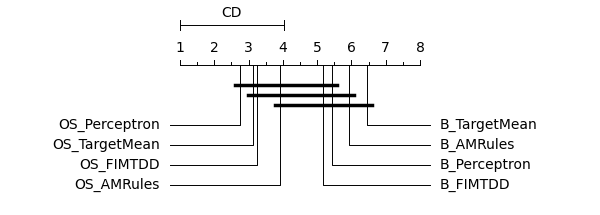}  
        \caption{Sampling Strategy: \texttt{HistOS}} 
        \label{fig:CD_HistOS_SERA}
    \end{subfigure}
    \caption{
    CD diagrams based on the obtained $SERA$ results, illustrating the performance comparison between the {\tt HistUS} and {\tt HistOS} sampling strategies and their respective baselines.} 
    \label{fig:cd_hist_sera}
\end{figure}

\begin{figure}[!h]
    \centering
    \begin{subfigure}{.5\textwidth}
        \centering
        \includegraphics[width=.9\linewidth]{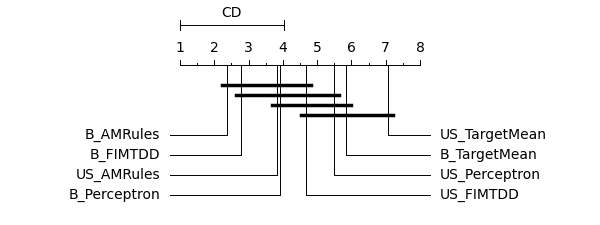}  
        \caption{Sampling Strategy: \texttt{HistUS}} 
        \label{fig:CD_HistUS_normal}
    \end{subfigure}%
    \begin{subfigure}{.5\textwidth}
        \centering
        \includegraphics[width=.9\linewidth]{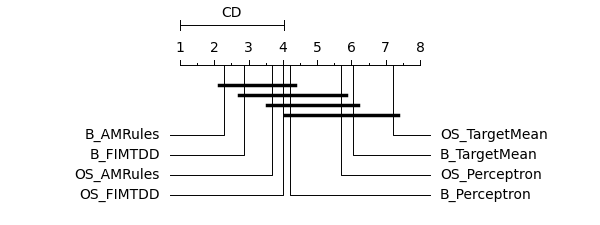}  
        \caption{Sampling Strategy: \texttt{HistOS}} 
        \label{fig:CD_HistOS_normal}
    \end{subfigure}
    \caption{
    CD diagrams based on the obtained $RMSE$ results, illustrating the performance comparison between the {\tt HistUS} and {\tt HistOS} sampling strategies and their respective baselines.} 
    \label{fig:cd_hist_rmse}
\end{figure}

Also, in Figures ~\ref{fig:fig_CD_B_Phi} to ~\ref{fig:cd_hist_rmse}, Critical Difference (CD) diagrams, based on the post-hoc Nemenyi, are used to compare HistUS and HistOS across various result metric functions.

\begin{figure}[H]
    \centering
    \begin{subfigure}{.48\textwidth}
        \centering
        \includegraphics[width=\linewidth]{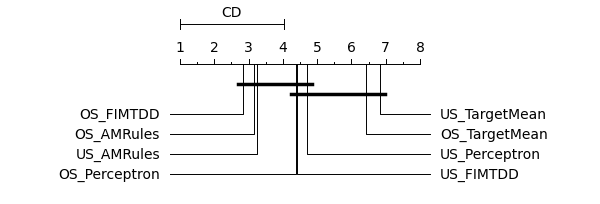}  
        \caption{Evaluation metric: $RMSE_{\phi}$} 
        \label{fig:CD_HistUS_OS_phi}
    \end{subfigure}%
    \hfill
    \begin{subfigure}{.48\textwidth}
        \centering
        \includegraphics[width=\linewidth]{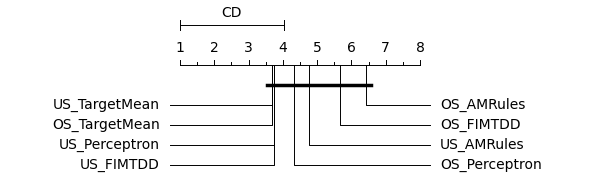}  
        \caption{Evaluation metric: $SERA$} 
        \label{fig:CD_HistUS_OS_sera}
    \end{subfigure}

    \vspace{1cm} 

    \begin{subfigure}{.6\textwidth} 
        \centering
        \includegraphics[width=\linewidth]{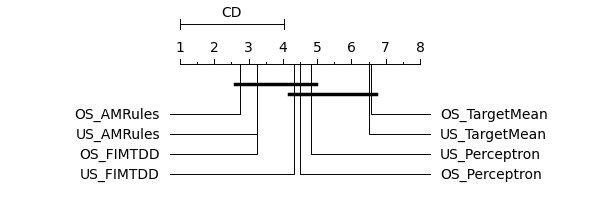}  
        \caption{Evaluation metric: $RMSE$} 
        \label{fig:CD_HistUS_OS_RMSE}
    \end{subfigure}
    
    \caption{
    CD diagrams comparing the \texttt{HistUS} and \texttt{HistOS} methods based on $RMSE_{\phi}$, $SERA$, and $RMSE$ evaluation metrics.}
    \label{fig:cd_histUS_OS}
\end{figure}

Additionally, direct comparisons between HistUS and HistOS are presented in Tables~\ref{tab:detail_HistUS_HistOS_RMSE_phi},~\ref{tab:detail_HistUS_HistOS_SERA}, and~\ref{tab:detail_HistUS_HistOS_RMSE}, and Figure~\ref{fig:cd_histUS_OS} offering a 
performance analysis of the two approaches.

\begin{figure}[H]
    \centering
    \begin{subfigure}{.48\textwidth}
        \centering
        \includegraphics[width=\linewidth]{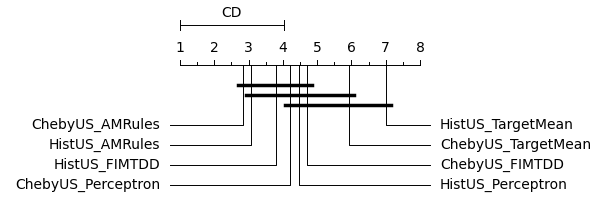}  
        \caption{Evaluation: $RMSE_{\phi}$, Methods: ChebyUS and HistUS} 
        \label{fig:CD_ChebyHist_US_phi}
    \end{subfigure}%
    \hfill
    \begin{subfigure}{.48\textwidth}
        \centering
        \includegraphics[width=\linewidth]{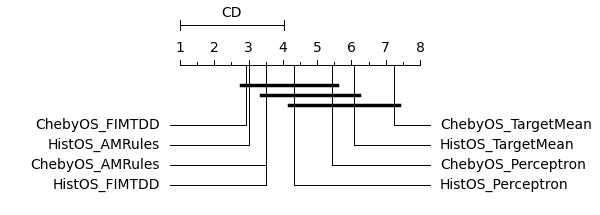}  
        \caption{Evaluation: $RMSE_{\phi}$, Methods: ChebyOS and HistOS} 
        \label{fig:CD_ChebyHist_OS_phi}
    \end{subfigure}

    \vspace{1cm} 

    \begin{subfigure}{.48\textwidth}
        \centering
        \includegraphics[width=\linewidth]{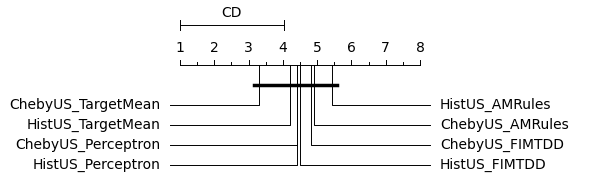}  
        \caption{Evaluation: $SERA$, Methods: ChebyUS and HistUS} 
        \label{fig:CD_ChebyHist_US_sera}
    \end{subfigure}%
    \hfill
    \begin{subfigure}{.48\textwidth}
        \centering
        \includegraphics[width=\linewidth]{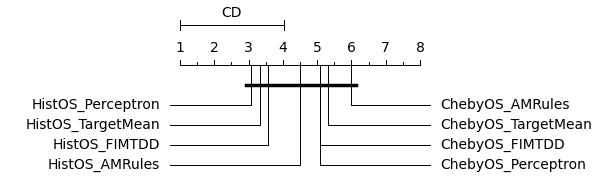}  
        \caption{Evaluation: $SERA$, Methods: ChebyOS and HistOS} 
        \label{fig:CD_ChebyHist_OS_sera}
    \end{subfigure}
    
    \caption{
    CD diagrams comparing ChebyUS, ChebyOS, HistUS, and HistOS methods based on $RMSE_{\phi}$ and $SERA$ evaluation metrics.}
    \label{fig:cd_cheby_hist}
\end{figure}

Furthermore, the complete results for the Chebyshev-based and histogram-based methods are provided in Tables~\ref{tab:detail_ChebyUS_HistUS_RMSE_phi} to~\ref{tab:detail_ChebyOS_HistOS_SERA} and Figure~\ref{fig:cd_cheby_hist}, allowing for a more comprehensive evaluation of these sampling strategies.

\begin{sidewaystable}
\centering
\caption{Evaluation: $RMSE_{\phi}$, Method: HistUS}
\label{tab:detail_HistUS_RMSE_phi}
\begin{tabular}
{lcccccccc}
\toprule
& \multicolumn{2}{@{}c@{}}{AMRules} & \multicolumn{2}{@{}c@{}}{Perceptron} & \multicolumn{2}{@{}c@{}}{FIMT-DD} & \multicolumn{2}{@{}c@{}}{TargetMean} \\
\cmidrule{2-9}
Dataset & Baseline & HistUS & Baseline & HistUS & Baseline & HistUS & Baseline & HistUS \\
\midrule
puma32H & 0.49 $\pm$ 0.30 & $\triangleright$\textbf{0.39 $\pm$ 0.51} & 0.71 $\pm$ 0.00 & $\triangleright$\textbf{0.57 $\pm$ 0.00} & \textbf{0.53 $\pm$ 1.00} & $\triangleleft$0.70 $\pm$ 0.36 & 1.00 $\pm$ 0.00 & 1.00 $\pm$ 0.00 \\ 
cpusum & \textbf{0.55 $\pm$ 0.41} & 0.56 $\pm$ 0.12 & \textbf{0.42 $\pm$ 0.02} & $\triangleleft$0.66 $\pm$ 1.00 & 0.62 $\pm$ 0.02 & \textbf{0.61 $\pm$ 0.05} & 1.00 $\pm$ 0.01 & $\triangleright$\textbf{0.73 $\pm$ 0.03} \\ 
mv & 0.05 $\pm$ 0.14 & $\triangleright$\textbf{0.03 $\pm$ 0.04} & 0.23 $\pm$ 0.01 & $\triangleright$\textbf{0.12 $\pm$ 0.03} & 0.10 $\pm$ 1.00 & $\triangleright$\textbf{0.06 $\pm$ 0.37} & 1.00 $\pm$ 0.02 & $\triangleright$\textbf{0.83 $\pm$ 0.02} \\ 
elevator & \textbf{0.00 $\pm$ 0.00} & 0.00 $\pm$ 0.00 & 0.00 $\pm$ 0.00 & \textbf{0.00 $\pm$ 0.00} & 0.00 $\pm$ 0.00 & \textbf{0.00 $\pm$ 0.00} & 0.00 $\pm$ 0.00 & $\triangleright$\textbf{0.00 $\pm$ 0.00} \\ 
energydata-complete & 0.96 $\pm$ 0.63 & $\triangleright$\textbf{0.79 $\pm$ 0.42} & 0.90 $\pm$ 0.23 & $\triangleright$\textbf{0.78 $\pm$ 0.19} & 0.97 $\pm$ 0.25 & $\triangleright$\textbf{0.85 $\pm$ 0.45} & 1.00 $\pm$ 0.35 & $\triangleright$\textbf{0.78 $\pm$ 0.71} \\ 
FriedmanArtificialDomain & 0.30 $\pm$ 0.33 & $\triangleright$\textbf{0.22 $\pm$ 0.88} & 0.34 $\pm$ 0.21 & $\triangleright$\textbf{0.22 $\pm$ 0.48} & 0.28 $\pm$ 1.00 & $\triangleright$\textbf{0.23 $\pm$ 0.87} & \textbf{1.00 $\pm$ 0.18} & 1.00 $\pm$ 0.19 \\ 
california & 0.56 $\pm$ 0.31 & \textbf{0.56 $\pm$ 0.48} & 0.59 $\pm$ 0.22 & $\triangleright$\textbf{0.57 $\pm$ 0.78} & \textbf{0.55 $\pm$ 0.37} & 0.55 $\pm$ 1.00 & 1.00 $\pm$ 0.15 & $\triangleright$\textbf{0.98 $\pm$ 0.23} \\ 
bike & 0.03 $\pm$ 0.08 & \textbf{0.03 $\pm$ 0.11} & \textbf{0.03 $\pm$ 0.01} & 0.03 $\pm$ 0.02 & \textbf{0.28 $\pm$ 0.25} & 0.29 $\pm$ 0.36 & 1.00 $\pm$ 0.11 & $\triangleright$\textbf{0.79 $\pm$ 0.13} \\ 
3d-spatial-network & 0.72 $\pm$ 0.81 & $\triangleright$\textbf{0.51 $\pm$ 0.25} & 0.96 $\pm$ 0.01 & $\triangleright$\textbf{0.66 $\pm$ 0.10} & 0.35 $\pm$ 0.51 & $\triangleright$\textbf{0.30 $\pm$ 1.00} & 1.00 $\pm$ 0.00 & $\triangleright$\textbf{0.74 $\pm$ 0.13} \\ 
pp-gas-emission & 0.46 $\pm$ 0.26 & $\triangleright$\textbf{0.40 $\pm$ 0.18} & 0.62 $\pm$ 0.11 & $\triangleright$\textbf{0.48 $\pm$ 0.08} & 0.53 $\pm$ 0.27 & $\triangleright$\textbf{0.51 $\pm$ 1.00} & 1.00 $\pm$ 0.14 & $\triangleright$\textbf{0.88 $\pm$ 0.24} \\ 
GPU-Kernel-Performance & 0.47 $\pm$ 0.98 & $\triangleright$\textbf{0.42 $\pm$ 0.38} & 0.68 $\pm$ 0.02 & $\triangleright$\textbf{0.58 $\pm$ 0.03} & 0.12 $\pm$ 0.25 & $\triangleright$\textbf{0.08 $\pm$ 0.03} & 1.00 $\pm$ 0.02 & $\triangleright$\textbf{0.86 $\pm$ 0.02} \\ 
polution & 0.77 $\pm$ 0.27 & $\triangleright$\textbf{0.59 $\pm$ 0.24} & 0.83 $\pm$ 0.04 & $\triangleright$\textbf{0.63 $\pm$ 0.03} & 0.67 $\pm$ 0.98 & $\triangleright$\textbf{0.55 $\pm$ 0.65} & 1.00 $\pm$ 0.10 & $\triangleright$\textbf{0.76 $\pm$ 0.16} \\ 
\midrule
Avg.Rank & 3.83 & 2.42 & 5.58 & 3.83 & 3.92 & 2.96 & 7.46 & 6.0 \\
\bottomrule
\end{tabular}
\end{sidewaystable}

\begin{sidewaystable}
\centering
\caption{Evaluation: $RMSE_{\phi}$, Method: HistOS}
\label{tab:detail_HistOS_RMSE_phi}
\begin{tabular}
{lcccccccc}
\toprule
& \multicolumn{2}{@{}c@{}}{AMRules} & \multicolumn{2}{@{}c@{}}{Perceptron} & \multicolumn{2}{@{}c@{}}{FIMT-DD} & \multicolumn{2}{@{}c@{}}{TargetMean} \\
\cmidrule{2-9}
Dataset & Baseline & HistUS & Baseline & HistUS & Baseline & HistUS & Baseline & HistUS \\
\midrule
puma32H & 0.49 $\pm$ 0.30 & \textbf{0.43 $\pm$ 0.99} & 0.71 $\pm$ 0.00 & $\triangleright$\textbf{0.57 $\pm$ 0.00} & 0.53 $\pm$ 1.00 & $\triangleright$\textbf{0.39 $\pm$ 0.51} & 1.00 $\pm$ 0.00 & 1.00 $\pm$ 0.00 \\ 
cpusum & 0.55 $\pm$ 0.41 & \textbf{0.43 $\pm$ 0.02} & \textbf{0.42 $\pm$ 0.02} & 0.47 $\pm$ 0.29 & 0.62 $\pm$ 0.02 & $\triangleright$\textbf{0.41 $\pm$ 0.01} & 1.00 $\pm$ 0.01 & $\triangleright$\textbf{0.85 $\pm$ 0.01} \\ 
mv & 0.05 $\pm$ 0.14 & $\triangleright$\textbf{0.03 $\pm$ 0.02} & 0.23 $\pm$ 0.01 & $\triangleright$\textbf{0.09 $\pm$ 0.01} & 0.10 $\pm$ 1.00 & $\triangleright$\textbf{0.04 $\pm$ 0.16} & 1.00 $\pm$ 0.02 & $\triangleright$\textbf{0.77 $\pm$ 0.02} \\ 
elevator & \textbf{0.00 $\pm$ 0.00} & 0.00 $\pm$ 0.00 & \textbf{0.00 $\pm$ 0.00} & 0.00 $\pm$ 0.00 & \textbf{0.00 $\pm$ 0.00} & $\triangleleft$1.00 $\pm$ 1.00 & 0.00 $\pm$ 0.00 & $\triangleright$\textbf{0.00 $\pm$ 0.00} \\ 
energydata-complete & 0.96 $\pm$ 0.63 & $\triangleright$\textbf{0.80 $\pm$ 0.65} & 0.90 $\pm$ 0.23 & $\triangleright$\textbf{0.78 $\pm$ 0.23} & 0.97 $\pm$ 0.25 & $\triangleright$\textbf{0.85 $\pm$ 1.00} & 1.00 $\pm$ 0.35 & $\triangleright$\textbf{0.76 $\pm$ 0.36} \\ 
FriedmanArtificialDomain & 0.30 $\pm$ 0.33 & $\triangleright$\textbf{0.23 $\pm$ 0.70} & 0.34 $\pm$ 0.21 & $\triangleright$\textbf{0.26 $\pm$ 0.14} & 0.28 $\pm$ 1.00 & $\triangleright$\textbf{0.20 $\pm$ 0.95} & \textbf{1.00 $\pm$ 0.18} & 1.00 $\pm$ 0.16 \\ 
california & 0.56 $\pm$ 0.31 & $\triangleright$\textbf{0.50 $\pm$ 0.47} & 0.59 $\pm$ 0.22 & $\triangleright$\textbf{0.52 $\pm$ 0.20} & 0.55 $\pm$ 0.37 & $\triangleright$\textbf{0.50 $\pm$ 0.41} & 1.00 $\pm$ 0.15 & $\triangleright$\textbf{0.90 $\pm$ 0.62} \\ 
bike & \textbf{0.03 $\pm$ 0.08} & 0.04 $\pm$ 0.21 & 0.03 $\pm$ 0.01 & \textbf{0.03 $\pm$ 0.12} & 0.28 $\pm$ 0.25 & \textbf{0.26 $\pm$ 1.00} & 1.00 $\pm$ 0.11 & $\triangleright$\textbf{0.71 $\pm$ 0.12} \\ 
3d-spatial-network & 0.72 $\pm$ 0.81 & $\triangleright$\textbf{0.46 $\pm$ 0.13} & 0.96 $\pm$ 0.01 & $\triangleright$\textbf{0.70 $\pm$ 0.01} & 0.35 $\pm$ 0.51 & $\triangleright$\textbf{0.25 $\pm$ 0.77} & 1.00 $\pm$ 0.00 & $\triangleright$\textbf{0.78 $\pm$ 0.01} \\ 
pp-gas-emission & 0.46 $\pm$ 0.26 & $\triangleright$\textbf{0.39 $\pm$ 0.22} & 0.62 $\pm$ 0.11 & $\triangleright$\textbf{0.48 $\pm$ 0.09} & 0.53 $\pm$ 0.27 & $\triangleright$\textbf{0.44 $\pm$ 0.75} & 1.00 $\pm$ 0.14 & $\triangleright$\textbf{0.91 $\pm$ 0.14} \\ 
GPU-Kernel-Performance & 0.47 $\pm$ 0.98 & $\triangleright$\textbf{0.39 $\pm$ 0.09} & 0.68 $\pm$ 0.02 & $\triangleright$\textbf{0.52 $\pm$ 0.03} & 0.12 $\pm$ 0.25 & $\triangleright$\textbf{0.07 $\pm$ 1.00} & 1.00 $\pm$ 0.02 & $\triangleright$\textbf{0.81 $\pm$ 0.02} \\ 
polution & 0.77 $\pm$ 0.27 & $\triangleright$\textbf{0.55 $\pm$ 0.11} & 0.83 $\pm$ 0.04 & $\triangleright$\textbf{0.58 $\pm$ 0.04} & 0.67 $\pm$ 0.98 & $\triangleright$\textbf{0.51 $\pm$ 1.00} & 1.00 $\pm$ 0.10 & $\triangleright$\textbf{0.75 $\pm$ 0.15} \\ 
\midrule
Avg.Rank & 4.25 & 2.58 & 5.5 & 3.75 & 4.33 & 2.33 & 7.38 & 5.88 \\
\bottomrule
\end{tabular}
\end{sidewaystable}

\begin{sidewaystable}
\centering
\caption{Evaluation: $SERA$, Method: HistUS}
\label{tab:detail_HistUS_SERA}
\begin{tabular}
{lcccccccc}
\toprule
& \multicolumn{2}{@{}c@{}}{AMRules} & \multicolumn{2}{@{}c@{}}{Perceptron} & \multicolumn{2}{@{}c@{}}{FIMT-DD} & \multicolumn{2}{@{}c@{}}{TargetMean} \\
\cmidrule{2-9}
Dataset & Baseline & HistUS & Baseline & HistUS & Baseline & HistUS & Baseline & HistUS \\
\midrule
puma32H & 0.75 $\pm$ 0.22 & $\triangleright$\textbf{0.57 $\pm$ 0.18} & 0.85 $\pm$ 0.00 & $\triangleright$\textbf{0.66 $\pm$ 0.02} & 0.65 $\pm$ 1.00 & $\triangleright$\textbf{0.50 $\pm$ 0.02} & 1.00 $\pm$ 0.00 & 1.00 $\pm$ 0.00 \\ 
cpusum & 1.00 $\pm$ 1.00 & $\triangleright$\textbf{0.69 $\pm$ 0.06} & \textbf{0.66 $\pm$ 0.00} & 0.66 $\pm$ 0.03 & 0.91 $\pm$ 0.07 & $\triangleright$\textbf{0.77 $\pm$ 0.12} & 0.73 $\pm$ 0.00 & $\triangleright$\textbf{0.60 $\pm$ 0.01} \\ 
mv & 0.64 $\pm$ 0.23 & $\triangleright$\textbf{0.61 $\pm$ 0.26} & 0.66 $\pm$ 0.00 & $\triangleright$\textbf{0.58 $\pm$ 0.00} & 1.00 $\pm$ 0.78 & $\triangleright$\textbf{0.88 $\pm$ 0.60} & 0.94 $\pm$ 0.00 & $\triangleright$\textbf{0.78 $\pm$ 0.00} \\ 
elevator & 0.91 $\pm$ 0.24 & \textbf{0.84 $\pm$ 1.00} & 1.00 $\pm$ 0.05 & $\triangleright$\textbf{0.81 $\pm$ 0.77} & 0.93 $\pm$ 0.19 & $\triangleright$\textbf{0.65 $\pm$ 0.05} & 0.98 $\pm$ 0.00 & $\triangleright$\textbf{0.41 $\pm$ 0.00} \\ 
energydata-complete & 1.00 $\pm$ 0.09 & $\triangleright$\textbf{0.75 $\pm$ 0.39} & 0.98 $\pm$ 0.02 & $\triangleright$\textbf{0.75 $\pm$ 0.32} & 0.94 $\pm$ 0.28 & $\triangleright$\textbf{0.74 $\pm$ 1.00} & 0.98 $\pm$ 0.01 & $\triangleright$\textbf{0.68 $\pm$ 0.73} \\ 
FriedmanArtificialDomain & 0.59 $\pm$ 0.05 & $\triangleright$\textbf{0.44 $\pm$ 0.11} & 0.61 $\pm$ 0.03 & $\triangleright$\textbf{0.46 $\pm$ 0.06} & 0.57 $\pm$ 0.11 & $\triangleright$\textbf{0.40 $\pm$ 1.00} & 1.00 $\pm$ 0.02 & \textbf{1.00 $\pm$ 0.01} \\ 
california & 0.98 $\pm$ 0.40 & $\triangleright$\textbf{0.95 $\pm$ 1.00} & \textbf{0.94 $\pm$ 0.58} & $\triangleleft$0.94 $\pm$ 0.37 & 1.00 $\pm$ 0.68 & $\triangleright$\textbf{0.98 $\pm$ 0.74} & \textbf{0.90 $\pm$ 0.34} & $\triangleleft$0.90 $\pm$ 0.34 \\ 
bike & 1.00 $\pm$ 1.00 & \textbf{0.96 $\pm$ 0.49} & 0.86 $\pm$ 0.07 & $\triangleright$\textbf{0.86 $\pm$ 0.10} & 0.86 $\pm$ 0.05 & $\triangleright$\textbf{0.82 $\pm$ 0.22} & 0.89 $\pm$ 0.01 & $\triangleright$\textbf{0.74 $\pm$ 0.02} \\ 
3d-spatial-network & 0.96 $\pm$ 0.07 & $\triangleright$\textbf{0.66 $\pm$ 1.00} & 0.98 $\pm$ 0.00 & $\triangleright$\textbf{0.57 $\pm$ 0.06} & 0.97 $\pm$ 0.14 & $\triangleright$\textbf{0.72 $\pm$ 0.54} & 1.00 $\pm$ 0.00 & $\triangleright$\textbf{0.61 $\pm$ 0.10} \\ 
pp-gas-emission & 0.82 $\pm$ 0.06 & $\triangleright$\textbf{0.61 $\pm$ 0.82} & 0.85 $\pm$ 0.04 & $\triangleright$\textbf{0.61 $\pm$ 0.69} & 0.76 $\pm$ 0.50 & $\triangleright$\textbf{0.55 $\pm$ 0.84} & 1.00 $\pm$ 0.02 & $\triangleright$\textbf{0.78 $\pm$ 1.00} \\ 
GPU-Kernel-Performance & 0.97 $\pm$ 0.03 & $\triangleright$\textbf{0.86 $\pm$ 0.08} & 0.92 $\pm$ 0.00 & $\triangleright$\textbf{0.79 $\pm$ 0.00} & 0.85 $\pm$ 0.59 & $\triangleright$\textbf{0.50 $\pm$ 0.84} & 1.00 $\pm$ 0.00 & $\triangleright$\textbf{0.70 $\pm$ 0.01} \\ 
polution & 1.00 $\pm$ 0.03 & $\triangleright$\textbf{0.80 $\pm$ 0.51} & 1.00 $\pm$ 0.03 & $\triangleright$\textbf{0.76 $\pm$ 0.12} & 0.98 $\pm$ 1.00 & $\triangleright$\textbf{0.80 $\pm$ 0.71} & 1.00 $\pm$ 0.07 & $\triangleright$\textbf{0.67 $\pm$ 0.39} \\ 
\midrule
Avg.Rank & 6.25 & 3.58 & 5.5 & 2.75 & 5.33 & 3.0 & 6.62 & 2.96 \\
\bottomrule
\end{tabular}
\end{sidewaystable}

\begin{sidewaystable}
\centering
\caption{Evaluation: $SERA$, Method: HistOS}
\label{tab:detail_HistOS_SERA}
\begin{tabular}
{lcccccccc}
\toprule
& \multicolumn{2}{@{}c@{}}{AMRules} & \multicolumn{2}{@{}c@{}}{Perceptron} & \multicolumn{2}{@{}c@{}}{FIMT-DD} & \multicolumn{2}{@{}c@{}}{TargetMean} \\
\cmidrule{2-9}
Dataset & Baseline & HistUS & Baseline & HistUS & Baseline & HistUS & Baseline & HistUS \\
\midrule
puma32H & 0.75 $\pm$ 0.22 & \textbf{0.71 $\pm$ 0.07} & 0.85 $\pm$ 0.00 & $\triangleright$\textbf{0.76 $\pm$ 0.01} & 0.65 $\pm$ 1.00 & \textbf{0.61 $\pm$ 0.23} & 1.00 $\pm$ 0.00 & 1.00 $\pm$ 0.00 \\ 
cpusum & 1.00 $\pm$ 1.00 & \textbf{0.85 $\pm$ 0.22} & \textbf{0.66 $\pm$ 0.00} & 0.68 $\pm$ 0.11 & 0.91 $\pm$ 0.07 & $\triangleright$\textbf{0.88 $\pm$ 0.06} & 0.73 $\pm$ 0.00 & $\triangleright$\textbf{0.65 $\pm$ 0.00} \\ 
mv & 0.64 $\pm$ 0.23 & \textbf{0.60 $\pm$ 0.35} & 0.66 $\pm$ 0.00 & $\triangleright$\textbf{0.57 $\pm$ 0.00} & 1.00 $\pm$ 0.78 & $\triangleright$\textbf{0.88 $\pm$ 1.00} & 0.94 $\pm$ 0.00 & $\triangleright$\textbf{0.74 $\pm$ 0.00} \\ 
elevator & 0.91 $\pm$ 0.24 & \textbf{0.88 $\pm$ 0.88} & 1.00 $\pm$ 0.05 & \textbf{0.92 $\pm$ 0.54} & 0.93 $\pm$ 0.19 & \textbf{0.86 $\pm$ 0.62} & 0.98 $\pm$ 0.00 & $\triangleright$\textbf{0.41 $\pm$ 0.00} \\ 
energydata-complete & 1.00 $\pm$ 0.09 & $\triangleright$\textbf{0.75 $\pm$ 0.37} & 0.98 $\pm$ 0.02 & $\triangleright$\textbf{0.74 $\pm$ 0.15} & 0.94 $\pm$ 0.28 & $\triangleright$\textbf{0.74 $\pm$ 0.87} & 0.98 $\pm$ 0.01 & $\triangleright$\textbf{0.65 $\pm$ 0.19} \\ 
FriedmanArtificialDomain & 0.59 $\pm$ 0.05 & $\triangleright$\textbf{0.52 $\pm$ 0.06} & 0.61 $\pm$ 0.03 & $\triangleright$\textbf{0.51 $\pm$ 0.05} & 0.57 $\pm$ 0.11 & $\triangleright$\textbf{0.51 $\pm$ 0.16} & \textbf{1.00 $\pm$ 0.02} & 1.00 $\pm$ 0.01 \\ 
california & 0.98 $\pm$ 0.40 & \textbf{0.98 $\pm$ 0.81} & \textbf{0.94 $\pm$ 0.58} & $\triangleleft$0.94 $\pm$ 0.54 & \textbf{1.00 $\pm$ 0.68} & 1.00 $\pm$ 1.00 & \textbf{0.90 $\pm$ 0.34} & $\triangleleft$0.91 $\pm$ 0.24 \\ 
bike & 1.00 $\pm$ 1.00 & \textbf{1.00 $\pm$ 0.56} & 0.86 $\pm$ 0.07 & $\triangleright$\textbf{0.85 $\pm$ 0.05} & 0.86 $\pm$ 0.05 & \textbf{0.86 $\pm$ 0.12} & 0.89 $\pm$ 0.01 & $\triangleright$\textbf{0.70 $\pm$ 0.00} \\ 
3d-spatial-network & 0.96 $\pm$ 0.07 & $\triangleright$\textbf{0.74 $\pm$ 0.30} & 0.98 $\pm$ 0.00 & $\triangleright$\textbf{0.61 $\pm$ 0.00} & 0.97 $\pm$ 0.14 & $\triangleright$\textbf{0.84 $\pm$ 0.43} & 1.00 $\pm$ 0.00 & $\triangleright$\textbf{0.65 $\pm$ 0.01} \\ 
pp-gas-emission & 0.82 $\pm$ 0.06 & $\triangleright$\textbf{0.71 $\pm$ 0.47} & 0.85 $\pm$ 0.04 & $\triangleright$\textbf{0.66 $\pm$ 0.14} & 0.76 $\pm$ 0.50 & $\triangleright$\textbf{0.64 $\pm$ 0.62} & 1.00 $\pm$ 0.02 & $\triangleright$\textbf{0.84 $\pm$ 0.29} \\ 
GPU-Kernel-Performance & 0.97 $\pm$ 0.03 & $\triangleright$\textbf{0.93 $\pm$ 0.15} & 0.92 $\pm$ 0.00 & $\triangleright$\textbf{0.71 $\pm$ 0.00} & 0.85 $\pm$ 0.59 & $\triangleright$\textbf{0.55 $\pm$ 1.00} & 1.00 $\pm$ 0.00 & $\triangleright$\textbf{0.61 $\pm$ 0.01} \\ 
polution & 1.00 $\pm$ 0.03 & $\triangleright$\textbf{0.81 $\pm$ 0.38} & 1.00 $\pm$ 0.03 & $\triangleright$\textbf{0.75 $\pm$ 0.04} & 0.98 $\pm$ 1.00 & $\triangleright$\textbf{0.82 $\pm$ 0.18} & 1.00 $\pm$ 0.07 & $\triangleright$\textbf{0.66 $\pm$ 0.13} \\ 
\midrule
Avg.Rank & 5.92 & 3.92 & 5.42 & 2.75 & 5.17 & 3.25 & 6.46 & 3.12 \\
\bottomrule
\end{tabular}
\end{sidewaystable}

\begin{sidewaystable}
\centering
\caption{Evaluation: $RMSE$, Method: HistUS}
\label{tab:detail_HistUS_RMSE}
\begin{tabular}
{lcccccccc}
\toprule
& \multicolumn{2}{@{}c@{}}{AMRules} & \multicolumn{2}{@{}c@{}}{Perceptron} & \multicolumn{2}{@{}c@{}}{FIMT-DD} & \multicolumn{2}{@{}c@{}}{TargetMean} \\
\cmidrule{2-9}
Dataset & Baseline & HistUS & Baseline & HistUS & Baseline & HistUS & Baseline & HistUS \\
\midrule
puma32H & \textbf{0.37 $\pm$ 0.00} & $\triangleleft$0.54 $\pm$ 0.20 & \textbf{0.56 $\pm$ 0.00} & $\triangleleft$0.74 $\pm$ 0.00 & \textbf{0.37 $\pm$ 0.00} & $\triangleleft$1.00 $\pm$ 1.00 & 0.56 $\pm$ 0.00 & 0.56 $\pm$ 0.00 \\ 
cpusum & \textbf{0.35 $\pm$ 0.02} & $\triangleleft$0.62 $\pm$ 0.05 & \textbf{0.48 $\pm$ 0.00} & $\triangleleft$1.00 $\pm$ 1.00 & \textbf{0.42 $\pm$ 0.00} & $\triangleleft$0.51 $\pm$ 0.01 & \textbf{0.79 $\pm$ 0.00} & $\triangleleft$0.84 $\pm$ 0.00 \\ 
mv & \textbf{0.15 $\pm$ 1.00} & 0.15 $\pm$ 0.17 & \textbf{0.39 $\pm$ 0.00} & $\triangleleft$0.41 $\pm$ 0.01 & 0.14 $\pm$ 0.31 & $\triangleright$\textbf{0.12 $\pm$ 0.07} & \textbf{0.90 $\pm$ 0.00} & $\triangleleft$0.96 $\pm$ 0.01 \\ 
elevator & 0.00 $\pm$ 0.00 & \textbf{0.00 $\pm$ 0.00} & 0.00 $\pm$ 0.00 & \textbf{0.00 $\pm$ 0.00} & 0.00 $\pm$ 0.00 & \textbf{0.00 $\pm$ 0.00} & 0.00 $\pm$ 0.00 & 0.00 $\pm$ 0.00 \\ 
energydata-complete & \textbf{0.70 $\pm$ 0.03} & $\triangleleft$0.86 $\pm$ 0.09 & \textbf{0.68 $\pm$ 0.03} & $\triangleleft$0.85 $\pm$ 0.08 & \textbf{0.76 $\pm$ 0.08} & $\triangleleft$0.98 $\pm$ 0.34 & \textbf{0.73 $\pm$ 0.02} & $\triangleleft$0.91 $\pm$ 0.07 \\ 
FriedmanArtificialDomain & \textbf{0.47 $\pm$ 0.01} & $\triangleleft$0.58 $\pm$ 0.04 & \textbf{0.53 $\pm$ 0.00} & $\triangleleft$0.58 $\pm$ 0.01 & \textbf{0.45 $\pm$ 0.02} & $\triangleleft$0.69 $\pm$ 1.00 & \textbf{1.00 $\pm$ 0.00} & 1.00 $\pm$ 0.00 \\ 
california & \textbf{0.58 $\pm$ 0.03} & $\triangleleft$0.59 $\pm$ 0.09 & \textbf{0.60 $\pm$ 0.03} & $\triangleleft$0.60 $\pm$ 0.02 & \textbf{0.64 $\pm$ 1.00} & 0.64 $\pm$ 0.50 & \textbf{0.97 $\pm$ 0.03} & $\triangleleft$0.98 $\pm$ 0.04 \\ 
bike & \textbf{0.03 $\pm$ 0.02} & 0.03 $\pm$ 0.04 & \textbf{0.04 $\pm$ 0.01} & 0.04 $\pm$ 0.01 & \textbf{0.35 $\pm$ 0.47} & $\triangleleft$0.39 $\pm$ 1.00 & \textbf{0.86 $\pm$ 0.03} & $\triangleleft$0.94 $\pm$ 0.07 \\ 
3d-spatial-network & \textbf{0.66 $\pm$ 0.08} & $\triangleleft$0.77 $\pm$ 1.00 & \textbf{0.81 $\pm$ 0.00} & $\triangleleft$1.00 $\pm$ 0.09 & \textbf{0.39 $\pm$ 0.23} & $\triangleleft$0.48 $\pm$ 0.83 & \textbf{0.82 $\pm$ 0.00} & $\triangleleft$1.00 $\pm$ 0.15 \\ 
pp-gas-emission & \textbf{0.51 $\pm$ 0.05} & $\triangleleft$0.61 $\pm$ 0.29 & \textbf{0.62 $\pm$ 0.02} & $\triangleleft$0.75 $\pm$ 0.47 & \textbf{0.63 $\pm$ 0.15} & $\triangleleft$0.92 $\pm$ 1.00 & \textbf{0.89 $\pm$ 0.02} & $\triangleleft$1.00 $\pm$ 0.22 \\ 
GPU-Kernel-Performance & 0.45 $\pm$ 0.24 & $\triangleright$\textbf{0.45 $\pm$ 0.21} & \textbf{0.64 $\pm$ 0.01} & $\triangleleft$0.67 $\pm$ 0.01 & \textbf{0.12 $\pm$ 0.16} & 0.13 $\pm$ 0.15 & \textbf{0.83 $\pm$ 0.01} & $\triangleleft$0.92 $\pm$ 0.01 \\ 
polution & \textbf{0.68 $\pm$ 0.05} & $\triangleleft$0.77 $\pm$ 0.15 & \textbf{0.71 $\pm$ 0.01} & $\triangleleft$0.82 $\pm$ 0.08 & \textbf{0.63 $\pm$ 0.52} & $\triangleleft$0.74 $\pm$ 0.44 & \textbf{0.83 $\pm$ 0.02} & $\triangleleft$0.98 $\pm$ 0.14 \\ 
\midrule
Avg.Rank & 2.38 & 3.83 & 3.92 & 5.5 & 2.79 & 4.67 & 5.83 & 7.08 \\
\bottomrule
\end{tabular}
\end{sidewaystable}

\begin{sidewaystable}
\centering
\caption{Evaluation: $RMSE$, Method: HistOS}
\label{tab:detail_HistOS_RMSE}
\begin{tabular}
{lcccccccc}
\toprule
& \multicolumn{2}{@{}c@{}}{AMRules} & \multicolumn{2}{@{}c@{}}{Perceptron} & \multicolumn{2}{@{}c@{}}{FIMT-DD} & \multicolumn{2}{@{}c@{}}{TargetMean} \\
\cmidrule{2-9}
Dataset & Baseline & HistUS & Baseline & HistUS & Baseline & HistUS & Baseline & HistUS \\
\midrule
puma32H & \textbf{0.37 $\pm$ 0.00} & 0.41 $\pm$ 0.36 & 0.56 $\pm$ 0.00 & 0.56 $\pm$ 0.00 & \textbf{0.37 $\pm$ 0.00} & $\triangleleft$0.46 $\pm$ 0.57 & 0.56 $\pm$ 0.00 & 0.56 $\pm$ 0.00 \\ 
cpusum & \textbf{0.35 $\pm$ 0.02} & 0.37 $\pm$ 0.01 & \textbf{0.48 $\pm$ 0.00} & 0.49 $\pm$ 0.01 & 0.42 $\pm$ 0.00 & $\triangleright$\textbf{0.30 $\pm$ 0.00} & \textbf{0.79 $\pm$ 0.00} & $\triangleleft$0.80 $\pm$ 0.00 \\ 
mv & \textbf{0.15 $\pm$ 1.00} & 0.15 $\pm$ 0.51 & \textbf{0.39 $\pm$ 0.00} & $\triangleleft$0.41 $\pm$ 0.00 & 0.14 $\pm$ 0.31 & $\triangleright$\textbf{0.07 $\pm$ 0.03} & \textbf{0.90 $\pm$ 0.00} & $\triangleleft$1.00 $\pm$ 0.01 \\ 
elevator & \textbf{0.00 $\pm$ 0.00} & 0.00 $\pm$ 0.00 & \textbf{0.00 $\pm$ 0.00} & 0.00 $\pm$ 0.00 & \textbf{0.00 $\pm$ 0.00} & $\triangleleft$1.00 $\pm$ 1.00 & 0.00 $\pm$ 0.00 & 0.00 $\pm$ 0.00 \\ 
energydata-complete & \textbf{0.70 $\pm$ 0.03} & $\triangleleft$0.90 $\pm$ 0.12 & \textbf{0.68 $\pm$ 0.03} & $\triangleleft$0.89 $\pm$ 0.14 & \textbf{0.76 $\pm$ 0.08} & $\triangleleft$1.00 $\pm$ 1.00 & \textbf{0.73 $\pm$ 0.02} & $\triangleleft$0.96 $\pm$ 0.06 \\ 
FriedmanArtificialDomain & \textbf{0.47 $\pm$ 0.01} & $\triangleleft$0.49 $\pm$ 0.03 & \textbf{0.53 $\pm$ 0.00} & $\triangleleft$0.54 $\pm$ 0.00 & \textbf{0.45 $\pm$ 0.02} & $\triangleleft$0.46 $\pm$ 0.03 & 1.00 $\pm$ 0.00 & 1.00 $\pm$ 0.00 \\ 
california & \textbf{0.58 $\pm$ 0.03} & $\triangleleft$0.60 $\pm$ 0.02 & \textbf{0.60 $\pm$ 0.03} & $\triangleleft$0.61 $\pm$ 0.03 & \textbf{0.64 $\pm$ 1.00} & 0.65 $\pm$ 0.65 & \textbf{0.97 $\pm$ 0.03} & $\triangleleft$1.00 $\pm$ 0.07 \\ 
bike & \textbf{0.03 $\pm$ 0.02} & 0.03 $\pm$ 0.07 & 0.04 $\pm$ 0.01 & \textbf{0.04 $\pm$ 0.14} & 0.35 $\pm$ 0.47 & \textbf{0.34 $\pm$ 0.68} & \textbf{0.86 $\pm$ 0.03} & $\triangleleft$1.00 $\pm$ 0.05 \\ 
3d-spatial-network & \textbf{0.66 $\pm$ 0.08} & $\triangleleft$0.71 $\pm$ 0.18 & \textbf{0.81 $\pm$ 0.00} & $\triangleleft$0.95 $\pm$ 0.01 & \textbf{0.39 $\pm$ 0.23} & 0.40 $\pm$ 1.00 & \textbf{0.82 $\pm$ 0.00} & $\triangleleft$0.95 $\pm$ 0.00 \\ 
pp-gas-emission & \textbf{0.51 $\pm$ 0.05} & $\triangleleft$0.53 $\pm$ 0.13 & \textbf{0.62 $\pm$ 0.02} & $\triangleleft$0.69 $\pm$ 0.07 & \textbf{0.63 $\pm$ 0.15} & $\triangleleft$0.69 $\pm$ 0.47 & \textbf{0.89 $\pm$ 0.02} & $\triangleleft$0.96 $\pm$ 0.06 \\ 
GPU-Kernel-Performance & 0.45 $\pm$ 0.24 & $\triangleright$\textbf{0.44 $\pm$ 0.09} & \textbf{0.64 $\pm$ 0.01} & $\triangleleft$0.73 $\pm$ 0.02 & 0.12 $\pm$ 0.16 & \textbf{0.12 $\pm$ 1.00} & \textbf{0.83 $\pm$ 0.01} & $\triangleleft$1.00 $\pm$ 0.01 \\ 
polution & \textbf{0.68 $\pm$ 0.05} & $\triangleleft$0.79 $\pm$ 0.20 & \textbf{0.71 $\pm$ 0.01} & $\triangleleft$0.86 $\pm$ 0.04 & \textbf{0.63 $\pm$ 0.52} & $\triangleleft$0.73 $\pm$ 1.00 & \textbf{0.83 $\pm$ 0.02} & $\triangleleft$1.00 $\pm$ 0.05 \\ 
\midrule
Avg.Rank & 2.29 & 3.67 & 4.21 & 5.71 & 2.88 & 4.0 & 6.04 & 7.21 \\
\bottomrule
\end{tabular}
\end{sidewaystable}

\clearpage

\begin{sidewaystable}
\centering
\caption{Evaluation: $RMSE_{\phi}$, Method: HistUS vs HistOS}
\label{tab:detail_HistUS_HistOS_RMSE_phi}
\begin{tabular}
{lcccccccc}
\toprule
& \multicolumn{2}{@{}c@{}}{AMRules} & \multicolumn{2}{@{}c@{}}{Perceptron} & \multicolumn{2}{@{}c@{}}{FIMT-DD} & \multicolumn{2}{@{}c@{}}{TargetMean} \\
\cmidrule{2-9}
Dataset & HistUS & HistOS & HistUS & HistOS & HistUS & HistOS & HistUS & HistOS \\
\midrule
puma32H & \textbf{0.39 $\pm$ 0.51} & 0.43 $\pm$ 0.99 & 0.57 $\pm$ 0.00 & 0.57 $\pm$ 0.00 & 0.70 $\pm$ 0.36 & $\triangleright$\textbf{0.39 $\pm$ 0.51} & 1.00 $\pm$ 0.00 & 1.00 $\pm$ 0.00 \\ 
cpusum & 0.56 $\pm$ 0.12 & $\triangleright$\textbf{0.43 $\pm$ 0.02} & 0.66 $\pm$ 1.00 & \textbf{0.47 $\pm$ 0.29} & 0.61 $\pm$ 0.05 & $\triangleright$\textbf{0.41 $\pm$ 0.01} & \textbf{0.73 $\pm$ 0.03} & $\triangleleft$0.85 $\pm$ 0.01 \\ 
mv & \textbf{0.03 $\pm$ 0.04} & 0.03 $\pm$ 0.02 & 0.12 $\pm$ 0.03 & $\triangleright$\textbf{0.09 $\pm$ 0.01} & 0.06 $\pm$ 0.37 & $\triangleright$\textbf{0.04 $\pm$ 0.16} & 0.83 $\pm$ 0.02 & $\triangleright$\textbf{0.77 $\pm$ 0.02} \\ 
elevator & \textbf{0.00 $\pm$ 0.00} & 0.00 $\pm$ 0.00 & \textbf{0.00 $\pm$ 0.00} & 0.00 $\pm$ 0.00 & \textbf{0.00 $\pm$ 0.00} & $\triangleleft$1.00 $\pm$ 1.00 & 0.00 $\pm$ 0.00 & 0.00 $\pm$ 0.00 \\ 
energydata-complete & \textbf{0.79 $\pm$ 0.42} & $\triangleleft$0.80 $\pm$ 0.65 & 0.78 $\pm$ 0.19 & $\triangleright$\textbf{0.78 $\pm$ 0.23} & 0.85 $\pm$ 0.45 & \textbf{0.85 $\pm$ 1.00} & 0.78 $\pm$ 0.71 & $\triangleright$\textbf{0.76 $\pm$ 0.36} \\ 
FriedmanArtificialDomain & \textbf{0.22 $\pm$ 0.88} & $\triangleleft$0.23 $\pm$ 0.70 & \textbf{0.22 $\pm$ 0.48} & $\triangleleft$0.26 $\pm$ 0.14 & 0.23 $\pm$ 0.87 & $\triangleright$\textbf{0.20 $\pm$ 0.95} & 1.00 $\pm$ 0.19 & 1.00 $\pm$ 0.16 \\ 
california & 0.56 $\pm$ 0.48 & $\triangleright$\textbf{0.50 $\pm$ 0.47} & 0.57 $\pm$ 0.78 & $\triangleright$\textbf{0.52 $\pm$ 0.20} & 0.55 $\pm$ 1.00 & $\triangleright$\textbf{0.50 $\pm$ 0.41} & 0.98 $\pm$ 0.23 & $\triangleright$\textbf{0.90 $\pm$ 0.62} \\ 
bike & \textbf{0.03 $\pm$ 0.11} & 0.04 $\pm$ 0.21 & 0.03 $\pm$ 0.02 & \textbf{0.03 $\pm$ 0.12} & 0.29 $\pm$ 0.36 & \textbf{0.26 $\pm$ 1.00} & 0.79 $\pm$ 0.13 & $\triangleright$\textbf{0.71 $\pm$ 0.12} \\ 
3d-spatial-network & 0.51 $\pm$ 0.25 & $\triangleright$\textbf{0.46 $\pm$ 0.13} & \textbf{0.66 $\pm$ 0.10} & $\triangleleft$0.70 $\pm$ 0.01 & 0.30 $\pm$ 1.00 & $\triangleright$\textbf{0.25 $\pm$ 0.77} & \textbf{0.74 $\pm$ 0.13} & $\triangleleft$0.78 $\pm$ 0.01 \\ 
pp-gas-emission & 0.40 $\pm$ 0.18 & \textbf{0.39 $\pm$ 0.22} & \textbf{0.48 $\pm$ 0.08} & 0.48 $\pm$ 0.09 & 0.51 $\pm$ 1.00 & $\triangleright$\textbf{0.44 $\pm$ 0.75} & \textbf{0.88 $\pm$ 0.24} & $\triangleleft$0.91 $\pm$ 0.14 \\ 
GPU-Kernel-Performance & 0.42 $\pm$ 0.38 & $\triangleright$\textbf{0.39 $\pm$ 0.09} & 0.58 $\pm$ 0.03 & $\triangleright$\textbf{0.52 $\pm$ 0.03} & 0.08 $\pm$ 0.03 & \textbf{0.07 $\pm$ 1.00} & 0.86 $\pm$ 0.02 & $\triangleright$\textbf{0.81 $\pm$ 0.02} \\ 
polution & 0.59 $\pm$ 0.24 & $\triangleright$\textbf{0.55 $\pm$ 0.11} & 0.63 $\pm$ 0.03 & $\triangleright$\textbf{0.58 $\pm$ 0.04} & 0.55 $\pm$ 0.65 & $\triangleright$\textbf{0.51 $\pm$ 1.00} & 0.76 $\pm$ 0.16 & $\triangleright$\textbf{0.75 $\pm$ 0.15} \\ 
\midrule
Avg.Rank & 3.25 & 3.17 & 4.71 & 4.38 & 4.42 & 2.83 & 6.83 & 6.42 \\
\bottomrule
\end{tabular}
\end{sidewaystable}

\begin{sidewaystable}
\centering
\caption{Evaluation: $SERA$, Method: HistUS vs HistOS}
\label{tab:detail_HistUS_HistOS_SERA}
\begin{tabular}
{lcccccccc}
\toprule
& \multicolumn{2}{@{}c@{}}{AMRules} & \multicolumn{2}{@{}c@{}}{Perceptron} & \multicolumn{2}{@{}c@{}}{FIMT-DD} & \multicolumn{2}{@{}c@{}}{TargetMean} \\
\cmidrule{2-9}
Dataset & HistUS & HistOS & HistUS & HistOS & HistUS & HistOS & HistUS & HistOS \\
\midrule
puma32H & \textbf{0.57 $\pm$ 0.18} & $\triangleleft$0.71 $\pm$ 0.07 & \textbf{0.66 $\pm$ 0.02} & $\triangleleft$0.76 $\pm$ 0.01 & \textbf{0.50 $\pm$ 0.02} & $\triangleleft$0.61 $\pm$ 0.23 & 1.00 $\pm$ 0.00 & 1.00 $\pm$ 0.00 \\ 
cpusum & \textbf{0.69 $\pm$ 0.06} & $\triangleleft$0.85 $\pm$ 0.22 & \textbf{0.66 $\pm$ 0.03} & 0.68 $\pm$ 0.11 & \textbf{0.77 $\pm$ 0.12} & $\triangleleft$0.88 $\pm$ 0.06 & \textbf{0.60 $\pm$ 0.01} & $\triangleleft$0.65 $\pm$ 0.00 \\ 
mv & 0.61 $\pm$ 0.26 & \textbf{0.60 $\pm$ 0.35} & 0.58 $\pm$ 0.00 & $\triangleright$\textbf{0.57 $\pm$ 0.00} & \textbf{0.88 $\pm$ 0.60} & 0.88 $\pm$ 1.00 & 0.78 $\pm$ 0.00 & $\triangleright$\textbf{0.74 $\pm$ 0.00} \\ 
elevator & \textbf{0.84 $\pm$ 1.00} & 0.88 $\pm$ 0.88 & \textbf{0.81 $\pm$ 0.77} & $\triangleleft$0.92 $\pm$ 0.54 & \textbf{0.65 $\pm$ 0.05} & $\triangleleft$0.86 $\pm$ 0.62 & 0.41 $\pm$ 0.00 & 0.41 $\pm$ 0.00 \\ 
energydata-complete & 0.75 $\pm$ 0.39 & \textbf{0.75 $\pm$ 0.37} & 0.75 $\pm$ 0.32 & $\triangleright$\textbf{0.74 $\pm$ 0.15} & 0.74 $\pm$ 1.00 & \textbf{0.74 $\pm$ 0.87} & 0.68 $\pm$ 0.73 & $\triangleright$\textbf{0.65 $\pm$ 0.19} \\ 
FriedmanArtificialDomain & \textbf{0.44 $\pm$ 0.11} & $\triangleleft$0.52 $\pm$ 0.06 & \textbf{0.46 $\pm$ 0.06} & $\triangleleft$0.51 $\pm$ 0.05 & \textbf{0.40 $\pm$ 1.00} & $\triangleleft$0.51 $\pm$ 0.16 & \textbf{1.00 $\pm$ 0.01} & $\triangleleft$1.00 $\pm$ 0.01 \\ 
california & \textbf{0.95 $\pm$ 1.00} & $\triangleleft$0.98 $\pm$ 0.81 & \textbf{0.94 $\pm$ 0.37} & 0.94 $\pm$ 0.54 & \textbf{0.98 $\pm$ 0.74} & $\triangleleft$1.00 $\pm$ 1.00 & \textbf{0.90 $\pm$ 0.34} & $\triangleleft$0.91 $\pm$ 0.24 \\ 
bike & \textbf{0.96 $\pm$ 0.49} & 1.00 $\pm$ 0.56 & 0.86 $\pm$ 0.10 & \textbf{0.85 $\pm$ 0.05} & \textbf{0.82 $\pm$ 0.22} & $\triangleleft$0.86 $\pm$ 0.12 & 0.74 $\pm$ 0.02 & $\triangleright$\textbf{0.70 $\pm$ 0.00} \\ 
3d-spatial-network & \textbf{0.66 $\pm$ 1.00} & $\triangleleft$0.74 $\pm$ 0.30 & \textbf{0.57 $\pm$ 0.06} & $\triangleleft$0.61 $\pm$ 0.00 & \textbf{0.72 $\pm$ 0.54} & $\triangleleft$0.84 $\pm$ 0.43 & \textbf{0.61 $\pm$ 0.10} & $\triangleleft$0.65 $\pm$ 0.01 \\ 
pp-gas-emission & \textbf{0.61 $\pm$ 0.82} & $\triangleleft$0.71 $\pm$ 0.47 & \textbf{0.61 $\pm$ 0.69} & $\triangleleft$0.66 $\pm$ 0.14 & \textbf{0.55 $\pm$ 0.84} & $\triangleleft$0.64 $\pm$ 0.62 & \textbf{0.78 $\pm$ 1.00} & $\triangleleft$0.84 $\pm$ 0.29 \\ 
GPU-Kernel-Performance & \textbf{0.86 $\pm$ 0.08} & $\triangleleft$0.93 $\pm$ 0.15 & 0.79 $\pm$ 0.00 & $\triangleright$\textbf{0.71 $\pm$ 0.00} & \textbf{0.50 $\pm$ 0.84} & 0.55 $\pm$ 1.00 & 0.70 $\pm$ 0.01 & $\triangleright$\textbf{0.61 $\pm$ 0.01} \\ 
polution & \textbf{0.80 $\pm$ 0.51} & $\triangleleft$0.81 $\pm$ 0.38 & 0.76 $\pm$ 0.12 & $\triangleright$\textbf{0.75 $\pm$ 0.04} & \textbf{0.80 $\pm$ 0.71} & $\triangleleft$0.82 $\pm$ 0.18 & 0.67 $\pm$ 0.39 & $\triangleright$\textbf{0.66 $\pm$ 0.13} \\ 
\midrule
Avg.Rank & 4.75 & 6.42 & 3.75 & 4.33 & 3.75 & 5.67 & 3.67 & 3.67 \\
\bottomrule
\end{tabular}
\end{sidewaystable}

\begin{sidewaystable}
\centering
\caption{Evaluation: $RMSE$, Method: HistUS vs HistOS}
\label{tab:detail_HistUS_HistOS_RMSE}
\begin{tabular}
{lcccccccc}
\toprule
& \multicolumn{2}{@{}c@{}}{AMRules} & \multicolumn{2}{@{}c@{}}{Perceptron} & \multicolumn{2}{@{}c@{}}{FIMT-DD} & \multicolumn{2}{@{}c@{}}{TargetMean} \\
\cmidrule{2-9}
Dataset & HistUS & HistOS & HistUS & HistOS & HistUS & HistOS & HistUS & HistOS \\
\midrule
puma32H & 0.54 $\pm$ 0.20 & $\triangleright$\textbf{0.41 $\pm$ 0.36} & 0.74 $\pm$ 0.00 & $\triangleright$\textbf{0.56 $\pm$ 0.00} & 1.00 $\pm$ 1.00 & $\triangleright$\textbf{0.46 $\pm$ 0.57} & 0.56 $\pm$ 0.00 & 0.56 $\pm$ 0.00 \\ 
cpusum & 0.62 $\pm$ 0.05 & $\triangleright$\textbf{0.37 $\pm$ 0.01} & 1.00 $\pm$ 1.00 & $\triangleright$\textbf{0.49 $\pm$ 0.01} & 0.51 $\pm$ 0.01 & $\triangleright$\textbf{0.30 $\pm$ 0.00} & 0.84 $\pm$ 0.00 & $\triangleright$\textbf{0.80 $\pm$ 0.00} \\ 
mv & 0.15 $\pm$ 0.17 & \textbf{0.15 $\pm$ 0.51} & \textbf{0.41 $\pm$ 0.01} & $\triangleleft$0.41 $\pm$ 0.00 & 0.12 $\pm$ 0.07 & $\triangleright$\textbf{0.07 $\pm$ 0.03} & \textbf{0.96 $\pm$ 0.01} & $\triangleleft$1.00 $\pm$ 0.01 \\ 
elevator & \textbf{0.00 $\pm$ 0.00} & 0.00 $\pm$ 0.00 & \textbf{0.00 $\pm$ 0.00} & 0.00 $\pm$ 0.00 & \textbf{0.00 $\pm$ 0.00} & $\triangleleft$1.00 $\pm$ 1.00 & 0.00 $\pm$ 0.00 & 0.00 $\pm$ 0.00 \\ 
energydata-complete & \textbf{0.86 $\pm$ 0.09} & $\triangleleft$0.90 $\pm$ 0.12 & \textbf{0.85 $\pm$ 0.08} & $\triangleleft$0.89 $\pm$ 0.14 & \textbf{0.98 $\pm$ 0.34} & 1.00 $\pm$ 1.00 & \textbf{0.91 $\pm$ 0.07} & $\triangleleft$0.96 $\pm$ 0.06 \\ 
FriedmanArtificialDomain & 0.58 $\pm$ 0.04 & $\triangleright$\textbf{0.49 $\pm$ 0.03} & 0.58 $\pm$ 0.01 & $\triangleright$\textbf{0.54 $\pm$ 0.00} & 0.69 $\pm$ 1.00 & $\triangleright$\textbf{0.46 $\pm$ 0.03} & 1.00 $\pm$ 0.00 & \textbf{1.00 $\pm$ 0.00} \\ 
california & \textbf{0.59 $\pm$ 0.09} & $\triangleleft$0.60 $\pm$ 0.02 & \textbf{0.60 $\pm$ 0.02} & $\triangleleft$0.61 $\pm$ 0.03 & \textbf{0.64 $\pm$ 0.50} & 0.65 $\pm$ 0.65 & \textbf{0.98 $\pm$ 0.04} & $\triangleleft$1.00 $\pm$ 0.07 \\ 
bike & \textbf{0.03 $\pm$ 0.04} & 0.03 $\pm$ 0.07 & 0.04 $\pm$ 0.01 & \textbf{0.04 $\pm$ 0.14} & 0.39 $\pm$ 1.00 & $\triangleright$\textbf{0.34 $\pm$ 0.68} & \textbf{0.94 $\pm$ 0.07} & $\triangleleft$1.00 $\pm$ 0.05 \\ 
3d-spatial-network & 0.77 $\pm$ 1.00 & $\triangleright$\textbf{0.71 $\pm$ 0.18} & 1.00 $\pm$ 0.09 & $\triangleright$\textbf{0.95 $\pm$ 0.01} & 0.48 $\pm$ 0.83 & $\triangleright$\textbf{0.40 $\pm$ 1.00} & 1.00 $\pm$ 0.15 & $\triangleright$\textbf{0.95 $\pm$ 0.00} \\ 
pp-gas-emission & 0.61 $\pm$ 0.29 & $\triangleright$\textbf{0.53 $\pm$ 0.13} & 0.75 $\pm$ 0.47 & $\triangleright$\textbf{0.69 $\pm$ 0.07} & 0.92 $\pm$ 1.00 & $\triangleright$\textbf{0.69 $\pm$ 0.47} & 1.00 $\pm$ 0.22 & $\triangleright$\textbf{0.96 $\pm$ 0.06} \\ 
GPU-Kernel-Performance & 0.45 $\pm$ 0.21 & $\triangleright$\textbf{0.44 $\pm$ 0.09} & \textbf{0.67 $\pm$ 0.01} & $\triangleleft$0.73 $\pm$ 0.02 & 0.13 $\pm$ 0.15 & \textbf{0.12 $\pm$ 1.00} & \textbf{0.92 $\pm$ 0.01} & $\triangleleft$1.00 $\pm$ 0.01 \\ 
polution & \textbf{0.77 $\pm$ 0.15} & $\triangleleft$0.79 $\pm$ 0.20 & \textbf{0.82 $\pm$ 0.08} & $\triangleleft$0.86 $\pm$ 0.04 & 0.74 $\pm$ 0.44 & \textbf{0.73 $\pm$ 1.00} & \textbf{0.98 $\pm$ 0.14} & $\triangleleft$1.00 $\pm$ 0.05 \\ 
\midrule
Avg.Rank & 3.25 & 2.75 & 4.83 & 4.5 & 4.33 & 3.25 & 6.5 & 6.58 \\
\bottomrule
\end{tabular}
\end{sidewaystable}

\clearpage
\begin{sidewaystable}
\centering
\caption{Evaluation: $RMSE_{\phi}$, Methods: ChebyUS VS HistUS}
\label{tab:detail_ChebyUS_HistUS_RMSE_phi}
\begin{tabular}
{lcccccccc}
\toprule
& \multicolumn{2}{@{}c@{}}{AMRules} & \multicolumn{2}{@{}c@{}}{Perceptron} & \multicolumn{2}{@{}c@{}}{FIMT-DD} & \multicolumn{2}{@{}c@{}}{TargetMean} \\
\cmidrule{2-9}
Dataset & ChebyUS & HistUS & ChebyUS & HistUS & ChebyUS & HistUS & ChebyUS & HistUS \\
\midrule
puma32H & 0.40 $\pm$ 0.44 & \textbf{0.39 $\pm$ 0.51} & 0.57 $\pm$ 0.00 & 0.57 $\pm$ 0.00 & \textbf{0.66 $\pm$ 0.67} & 0.70 $\pm$ 0.36 & 1.00 $\pm$ 0.00 & 1.00 $\pm$ 0.00 \\ 
cpusum & \textbf{0.29 $\pm$ 0.08} & $\triangleleft$0.56 $\pm$ 0.12 & \textbf{0.45 $\pm$ 1.00} & $\triangleleft$0.66 $\pm$ 1.00 & 1.00 $\pm$ 0.57 & $\triangleright$\textbf{0.61 $\pm$ 0.05} & \textbf{0.26 $\pm$ 0.01} & $\triangleleft$0.73 $\pm$ 0.03 \\ 
mv & 0.03 $\pm$ 0.17 & \textbf{0.03 $\pm$ 0.04} & \textbf{0.07 $\pm$ 0.01} & $\triangleleft$0.12 $\pm$ 0.03 & 0.15 $\pm$ 0.16 & $\triangleright$\textbf{0.06 $\pm$ 0.37} & \textbf{0.65 $\pm$ 0.01} & $\triangleleft$0.83 $\pm$ 0.02 \\ 
elevator & 0.42 $\pm$ 0.16 & \textbf{0.00 $\pm$ 0.00} & 0.32 $\pm$ 0.06 & $\triangleright$\textbf{0.00 $\pm$ 0.00} & 0.07 $\pm$ 0.00 & \textbf{0.00 $\pm$ 0.00} & 0.07 $\pm$ 0.00 & \textbf{0.00 $\pm$ 0.00} \\ 
energydata-complete & \textbf{0.78 $\pm$ 0.01} & 0.79 $\pm$ 0.42 & \textbf{0.77 $\pm$ 0.02} & $\triangleleft$0.78 $\pm$ 0.19 & 0.95 $\pm$ 0.40 & $\triangleright$\textbf{0.85 $\pm$ 0.45} & \textbf{0.72 $\pm$ 1.00} & 0.78 $\pm$ 0.71 \\ 
FriedmanArtificialDomain & 0.24 $\pm$ 0.19 & $\triangleright$\textbf{0.22 $\pm$ 0.88} & 0.24 $\pm$ 0.04 & $\triangleright$\textbf{0.22 $\pm$ 0.48} & 0.24 $\pm$ 1.00 & \textbf{0.23 $\pm$ 0.87} & \textbf{1.00 $\pm$ 0.04} & 1.00 $\pm$ 0.19 \\ 
california & \textbf{0.45 $\pm$ 0.49} & $\triangleleft$0.56 $\pm$ 0.48 & \textbf{0.48 $\pm$ 0.76} & $\triangleleft$0.57 $\pm$ 0.78 & \textbf{0.50 $\pm$ 1.00} & $\triangleleft$0.55 $\pm$ 1.00 & \textbf{0.78 $\pm$ 0.19} & $\triangleleft$0.98 $\pm$ 0.23 \\ 
bike & 0.03 $\pm$ 0.05 & \textbf{0.03 $\pm$ 0.11} & 0.06 $\pm$ 0.05 & $\triangleright$\textbf{0.03 $\pm$ 0.02} & 0.49 $\pm$ 1.00 & $\triangleright$\textbf{0.29 $\pm$ 0.36} & \textbf{0.59 $\pm$ 0.02} & $\triangleleft$0.79 $\pm$ 0.13 \\ 
3d-spatial-network & \textbf{0.48 $\pm$ 0.52} & $\triangleleft$0.51 $\pm$ 0.25 & 0.67 $\pm$ 0.00 & $\triangleright$\textbf{0.66 $\pm$ 0.10} & \textbf{0.29 $\pm$ 0.15} & 0.30 $\pm$ 1.00 & \textbf{0.73 $\pm$ 0.01} & $\triangleleft$0.74 $\pm$ 0.13 \\ 
pp-gas-emission & 0.42 $\pm$ 0.28 & $\triangleright$\textbf{0.40 $\pm$ 0.18} & 0.51 $\pm$ 0.40 & $\triangleright$\textbf{0.48 $\pm$ 0.08} & 0.52 $\pm$ 1.00 & $\triangleright$\textbf{0.51 $\pm$ 1.00} & 0.93 $\pm$ 0.28 & $\triangleright$\textbf{0.88 $\pm$ 0.24} \\ 
GPU-Kernel-Performance & \textbf{0.39 $\pm$ 0.40} & $\triangleleft$0.42 $\pm$ 0.38 & \textbf{0.47 $\pm$ 0.01} & $\triangleleft$0.58 $\pm$ 0.03 & 0.27 $\pm$ 0.35 & $\triangleright$\textbf{0.08 $\pm$ 0.03} & \textbf{0.69 $\pm$ 0.01} & $\triangleleft$0.86 $\pm$ 0.02 \\ 
polution & \textbf{0.50 $\pm$ 0.31} & $\triangleleft$0.59 $\pm$ 0.24 & \textbf{0.50 $\pm$ 0.08} & $\triangleleft$0.63 $\pm$ 0.03 & \textbf{0.49 $\pm$ 0.30} & $\triangleleft$0.55 $\pm$ 0.65 & \textbf{0.63 $\pm$ 0.03} & $\triangleleft$0.76 $\pm$ 0.16 \\ 
\midrule
Avg.Rank & 2.83 & 3.08 & 4.21 & 4.46 & 4.71 & 3.79 & 5.92 & 7.0 \\
\bottomrule
\end{tabular}
\end{sidewaystable}

\begin{sidewaystable}
\centering
\caption{Evaluation: $RMSE_{\phi}$, Methods: ChebyOS VS HistOS}
\label{tab:detail_ChebyOS_HistOS_RMSE_phi}
\begin{tabular}
{lcccccccc}
\toprule
& \multicolumn{2}{@{}c@{}}{AMRules} & \multicolumn{2}{@{}c@{}}{Perceptron} & \multicolumn{2}{@{}c@{}}{FIMT-DD} & \multicolumn{2}{@{}c@{}}{TargetMean} \\
\cmidrule{2-9}
Dataset & ChebyOS & HistOS & ChebyOS & HistOS & ChebyOS & HistOS & ChebyOS & HistOS \\
\midrule
puma32H & \textbf{0.41 $\pm$ 0.81} & 0.43 $\pm$ 0.99 & 0.64 $\pm$ 0.69 & $\triangleright$\textbf{0.57 $\pm$ 0.00} & \textbf{0.17 $\pm$ 0.44} & $\triangleleft$0.39 $\pm$ 0.51 & 1.00 $\pm$ 0.00 & 1.00 $\pm$ 0.00 \\ 
cpusum & \textbf{0.16 $\pm$ 0.10} & 0.43 $\pm$ 0.02 & \textbf{0.18 $\pm$ 0.00} & 0.47 $\pm$ 0.29 & \textbf{0.20 $\pm$ 0.01} & $\triangleleft$0.41 $\pm$ 0.01 & \textbf{0.31 $\pm$ 0.03} & $\triangleleft$0.85 $\pm$ 0.01 \\ 
mv & 0.04 $\pm$ 0.19 & $\triangleright$\textbf{0.03 $\pm$ 0.02} & 0.17 $\pm$ 0.01 & $\triangleright$\textbf{0.09 $\pm$ 0.01} & \textbf{0.04 $\pm$ 0.18} & 0.04 $\pm$ 0.16 & 0.93 $\pm$ 0.01 & $\triangleright$\textbf{0.77 $\pm$ 0.02} \\ 
elevator & 0.43 $\pm$ 0.16 & \textbf{0.00 $\pm$ 0.00} & 0.88 $\pm$ 0.61 & \textbf{0.00 $\pm$ 0.00} & \textbf{0.44 $\pm$ 1.00} & 1.00 $\pm$ 1.00 & 0.14 $\pm$ 0.00 & $\triangleright$\textbf{0.00 $\pm$ 0.00} \\ 
energydata-complete & 0.86 $\pm$ 0.01 & $\triangleright$\textbf{0.80 $\pm$ 0.65} & 0.83 $\pm$ 0.02 & $\triangleright$\textbf{0.78 $\pm$ 0.23} & 0.87 $\pm$ 0.02 & \textbf{0.85 $\pm$ 1.00} & 0.90 $\pm$ 0.03 & $\triangleright$\textbf{0.76 $\pm$ 0.36} \\ 
FriedmanArtificialDomain & 0.27 $\pm$ 0.12 & $\triangleright$\textbf{0.23 $\pm$ 0.70} & 0.30 $\pm$ 0.02 & $\triangleright$\textbf{0.26 $\pm$ 0.14} & \textbf{0.19 $\pm$ 0.21} & 0.20 $\pm$ 0.95 & 1.00 $\pm$ 0.04 & 1.00 $\pm$ 0.16 \\ 
california & \textbf{0.50 $\pm$ 0.38} & 0.50 $\pm$ 0.47 & 0.54 $\pm$ 0.06 & $\triangleright$\textbf{0.52 $\pm$ 0.20} & \textbf{0.49 $\pm$ 0.83} & 0.50 $\pm$ 0.41 & 0.94 $\pm$ 0.08 & $\triangleright$\textbf{0.90 $\pm$ 0.62} \\ 
bike & \textbf{0.02 $\pm$ 0.04} & 0.04 $\pm$ 0.21 & 0.03 $\pm$ 0.03 & \textbf{0.03 $\pm$ 0.12} & \textbf{0.23 $\pm$ 0.31} & $\triangleleft$0.26 $\pm$ 1.00 & 0.92 $\pm$ 0.01 & $\triangleright$\textbf{0.71 $\pm$ 0.12} \\ 
3d-spatial-network & 0.57 $\pm$ 0.09 & $\triangleright$\textbf{0.46 $\pm$ 0.13} & 0.89 $\pm$ 0.00 & $\triangleright$\textbf{0.70 $\pm$ 0.01} & \textbf{0.25 $\pm$ 0.17} & 0.25 $\pm$ 0.77 & 0.95 $\pm$ 0.00 & $\triangleright$\textbf{0.78 $\pm$ 0.01} \\ 
pp-gas-emission & 0.41 $\pm$ 0.07 & $\triangleright$\textbf{0.39 $\pm$ 0.22} & 0.54 $\pm$ 0.26 & $\triangleright$\textbf{0.48 $\pm$ 0.09} & \textbf{0.44 $\pm$ 0.23} & 0.44 $\pm$ 0.75 & 0.97 $\pm$ 0.22 & $\triangleright$\textbf{0.91 $\pm$ 0.14} \\ 
GPU-Kernel-Performance & 0.42 $\pm$ 0.15 & $\triangleright$\textbf{0.39 $\pm$ 0.09} & 0.52 $\pm$ 0.01 & $\triangleright$\textbf{0.52 $\pm$ 0.03} & 0.08 $\pm$ 1.00 & \textbf{0.07 $\pm$ 1.00} & 0.90 $\pm$ 0.02 & $\triangleright$\textbf{0.81 $\pm$ 0.02} \\ 
polution & 0.63 $\pm$ 0.17 & $\triangleright$\textbf{0.55 $\pm$ 0.11} & 0.71 $\pm$ 0.05 & $\triangleright$\textbf{0.58 $\pm$ 0.04} & 0.55 $\pm$ 1.00 & $\triangleright$\textbf{0.51 $\pm$ 1.00} & 0.91 $\pm$ 0.02 & $\triangleright$\textbf{0.75 $\pm$ 0.15} \\ 
\midrule
Avg.Rank & 3.5 & 3.0 & 5.42 & 4.33 & 2.92 & 3.5 & 7.25 & 6.08 \\
\bottomrule
\end{tabular}
\end{sidewaystable}

\begin{sidewaystable}
\centering
\caption{Evaluation: $SERA$, Methods: ChebyUS VS HistUS}
\label{tab:detail_ChebyUS_HistUS_SERA}
\begin{tabular}
{lcccccccc}
\toprule
& \multicolumn{2}{@{}c@{}}{AMRules} & \multicolumn{2}{@{}c@{}}{Perceptron} & \multicolumn{2}{@{}c@{}}{FIMT-DD} & \multicolumn{2}{@{}c@{}}{TargetMean} \\
\cmidrule{2-9}
Dataset & ChebyUS & HistUS & ChebyUS & HistUS & ChebyUS & HistUS & ChebyUS & HistUS \\
\midrule
puma32H & 0.62 $\pm$ 0.32 & $\triangleright$\textbf{0.57 $\pm$ 0.18} & 0.70 $\pm$ 0.03 & $\triangleright$\textbf{0.66 $\pm$ 0.02} & 0.54 $\pm$ 0.61 & $\triangleright$\textbf{0.50 $\pm$ 0.02} & 1.00 $\pm$ 0.01 & 1.00 $\pm$ 0.00 \\ 
cpusum & \textbf{0.56 $\pm$ 0.12} & $\triangleleft$0.69 $\pm$ 0.06 & 0.74 $\pm$ 0.82 & \textbf{0.66 $\pm$ 0.03} & 0.89 $\pm$ 0.36 & $\triangleright$\textbf{0.77 $\pm$ 0.12} & \textbf{0.52 $\pm$ 0.00} & $\triangleleft$0.60 $\pm$ 0.01 \\ 
mv & 0.68 $\pm$ 0.78 & \textbf{0.61 $\pm$ 0.26} & 0.60 $\pm$ 0.00 & \textbf{0.58 $\pm$ 0.00} & 0.99 $\pm$ 0.09 & $\triangleright$\textbf{0.88 $\pm$ 0.60} & \textbf{0.71 $\pm$ 0.00} & $\triangleleft$0.78 $\pm$ 0.00 \\ 
elevator & \textbf{0.82 $\pm$ 1.00} & 0.84 $\pm$ 1.00 & 0.89 $\pm$ 0.64 & \textbf{0.81 $\pm$ 0.77} & \textbf{0.58 $\pm$ 0.01} & $\triangleleft$0.65 $\pm$ 0.05 & \textbf{0.41 $\pm$ 0.00} & 0.41 $\pm$ 0.00 \\ 
energydata-complete & \textbf{0.66 $\pm$ 0.05} & $\triangleleft$0.75 $\pm$ 0.39 & \textbf{0.67 $\pm$ 0.02} & $\triangleleft$0.75 $\pm$ 0.32 & \textbf{0.68 $\pm$ 0.03} & $\triangleleft$0.74 $\pm$ 1.00 & \textbf{0.59 $\pm$ 1.00} & $\triangleleft$0.68 $\pm$ 0.73 \\ 
FriedmanArtificialDomain & 0.47 $\pm$ 0.38 & $\triangleright$\textbf{0.44 $\pm$ 0.11} & 0.49 $\pm$ 0.02 & $\triangleright$\textbf{0.46 $\pm$ 0.06} & 0.43 $\pm$ 1.00 & $\triangleright$\textbf{0.40 $\pm$ 1.00} & 1.00 $\pm$ 0.02 & \textbf{1.00 $\pm$ 0.01} \\ 
california & 0.99 $\pm$ 0.40 & $\triangleright$\textbf{0.95 $\pm$ 1.00} & 0.94 $\pm$ 0.87 & $\triangleright$\textbf{0.94 $\pm$ 0.37} & 1.00 $\pm$ 0.79 & $\triangleright$\textbf{0.98 $\pm$ 0.74} & 0.90 $\pm$ 0.10 & $\triangleright$\textbf{0.90 $\pm$ 0.34} \\ 
bike & \textbf{0.88 $\pm$ 1.00} & $\triangleleft$0.96 $\pm$ 0.49 & \textbf{0.86 $\pm$ 0.06} & 0.86 $\pm$ 0.10 & \textbf{0.81 $\pm$ 0.05} & 0.82 $\pm$ 0.22 & \textbf{0.66 $\pm$ 0.00} & $\triangleleft$0.74 $\pm$ 0.02 \\ 
3d-spatial-network & 0.70 $\pm$ 1.00 & $\triangleright$\textbf{0.66 $\pm$ 1.00} & 0.59 $\pm$ 0.01 & $\triangleright$\textbf{0.57 $\pm$ 0.06} & 0.80 $\pm$ 0.22 & $\triangleright$\textbf{0.72 $\pm$ 0.54} & \textbf{0.60 $\pm$ 0.00} & $\triangleleft$0.61 $\pm$ 0.10 \\ 
pp-gas-emission & 0.68 $\pm$ 1.00 & $\triangleright$\textbf{0.61 $\pm$ 0.82} & 0.65 $\pm$ 0.26 & $\triangleright$\textbf{0.61 $\pm$ 0.69} & 0.58 $\pm$ 0.46 & $\triangleright$\textbf{0.55 $\pm$ 0.84} & 0.89 $\pm$ 0.45 & $\triangleright$\textbf{0.78 $\pm$ 1.00} \\ 
GPU-Kernel-Performance & \textbf{0.72 $\pm$ 0.13} & $\triangleleft$0.86 $\pm$ 0.08 & \textbf{0.62 $\pm$ 0.00} & $\triangleleft$0.79 $\pm$ 0.00 & 0.66 $\pm$ 0.27 & $\triangleright$\textbf{0.50 $\pm$ 0.84} & \textbf{0.43 $\pm$ 0.00} & $\triangleleft$0.70 $\pm$ 0.01 \\ 
polution & \textbf{0.72 $\pm$ 0.38} & $\triangleleft$0.80 $\pm$ 0.51 & \textbf{0.71 $\pm$ 0.49} & $\triangleleft$0.76 $\pm$ 0.12 & \textbf{0.74 $\pm$ 1.00} & $\triangleleft$0.80 $\pm$ 0.71 & \textbf{0.59 $\pm$ 0.02} & $\triangleleft$0.67 $\pm$ 0.39 \\ 
\midrule
Avg.Rank & 4.92 & 5.42 & 4.42 & 4.42 & 4.83 & 4.5 & 3.29 & 4.21 \\
\bottomrule
\end{tabular}
\end{sidewaystable}

\begin{sidewaystable}
\centering
\caption{Evaluation: $SERA$, Methods: ChebyOS VS HistOS}
\label{tab:detail_ChebyOS_HistOS_SERA}
\begin{tabular}
{lcccccccc}
\toprule
& \multicolumn{2}{@{}c@{}}{AMRules} & \multicolumn{2}{@{}c@{}}{Perceptron} & \multicolumn{2}{@{}c@{}}{FIMT-DD} & \multicolumn{2}{@{}c@{}}{TargetMean} \\
\cmidrule{2-9}
Dataset & ChebyOS & HistOS & ChebyOS & HistOS & ChebyOS & HistOS & ChebyOS & HistOS \\
\midrule
puma32H & \textbf{0.70 $\pm$ 0.32} & 0.71 $\pm$ 0.07 & 0.81 $\pm$ 0.01 & $\triangleright$\textbf{0.76 $\pm$ 0.01} & \textbf{0.55 $\pm$ 0.25} & $\triangleleft$0.61 $\pm$ 0.23 & 1.00 $\pm$ 0.01 & 1.00 $\pm$ 0.00 \\ 
cpusum & 0.93 $\pm$ 1.00 & \textbf{0.85 $\pm$ 0.22} & \textbf{0.60 $\pm$ 0.01} & $\triangleleft$0.68 $\pm$ 0.11 & 1.00 $\pm$ 0.13 & $\triangleright$\textbf{0.88 $\pm$ 0.06} & \textbf{0.60 $\pm$ 0.01} & $\triangleleft$0.65 $\pm$ 0.00 \\ 
mv & 0.65 $\pm$ 0.55 & \textbf{0.60 $\pm$ 0.35} & 0.64 $\pm$ 0.00 & $\triangleright$\textbf{0.57 $\pm$ 0.00} & 0.95 $\pm$ 0.72 & \textbf{0.88 $\pm$ 1.00} & 0.91 $\pm$ 0.01 & $\triangleright$\textbf{0.74 $\pm$ 0.00} \\ 
elevator & 0.94 $\pm$ 0.32 & \textbf{0.88 $\pm$ 0.88} & 0.99 $\pm$ 0.12 & \textbf{0.92 $\pm$ 0.54} & \textbf{0.85 $\pm$ 0.16} & 0.86 $\pm$ 0.62 & 0.97 $\pm$ 0.00 & $\triangleright$\textbf{0.41 $\pm$ 0.00} \\ 
energydata-complete & 0.87 $\pm$ 0.03 & $\triangleright$\textbf{0.75 $\pm$ 0.37} & 0.86 $\pm$ 0.00 & $\triangleright$\textbf{0.74 $\pm$ 0.15} & 0.85 $\pm$ 0.09 & $\triangleright$\textbf{0.74 $\pm$ 0.87} & 0.84 $\pm$ 0.11 & $\triangleright$\textbf{0.65 $\pm$ 0.19} \\ 
FriedmanArtificialDomain & 0.54 $\pm$ 0.12 & $\triangleright$\textbf{0.52 $\pm$ 0.06} & 0.55 $\pm$ 0.03 & $\triangleright$\textbf{0.51 $\pm$ 0.05} & \textbf{0.50 $\pm$ 0.16} & $\triangleleft$0.51 $\pm$ 0.16 & 1.00 $\pm$ 0.02 & 1.00 $\pm$ 0.01 \\ 
california & \textbf{0.97 $\pm$ 0.35} & $\triangleleft$0.98 $\pm$ 0.81 & \textbf{0.91 $\pm$ 0.56} & 0.94 $\pm$ 0.54 & \textbf{0.99 $\pm$ 1.00} & $\triangleleft$1.00 $\pm$ 1.00 & \textbf{0.87 $\pm$ 0.39} & 0.91 $\pm$ 0.24 \\ 
bike & \textbf{0.97 $\pm$ 0.52} & 1.00 $\pm$ 0.56 & 0.86 $\pm$ 0.05 & \textbf{0.85 $\pm$ 0.05} & 0.86 $\pm$ 0.20 & \textbf{0.86 $\pm$ 0.12} & 0.83 $\pm$ 0.01 & $\triangleright$\textbf{0.70 $\pm$ 0.00} \\ 
3d-spatial-network & 0.87 $\pm$ 0.05 & $\triangleright$\textbf{0.74 $\pm$ 0.30} & 0.87 $\pm$ 0.00 & $\triangleright$\textbf{0.61 $\pm$ 0.00} & 0.93 $\pm$ 0.13 & $\triangleright$\textbf{0.84 $\pm$ 0.43} & 0.90 $\pm$ 0.00 & $\triangleright$\textbf{0.65 $\pm$ 0.01} \\ 
pp-gas-emission & 0.76 $\pm$ 0.41 & $\triangleright$\textbf{0.71 $\pm$ 0.47} & 0.76 $\pm$ 0.23 & $\triangleright$\textbf{0.66 $\pm$ 0.14} & 0.71 $\pm$ 0.49 & $\triangleright$\textbf{0.64 $\pm$ 0.62} & 0.96 $\pm$ 0.13 & $\triangleright$\textbf{0.84 $\pm$ 0.29} \\ 
GPU-Kernel-Performance & 0.96 $\pm$ 0.03 & $\triangleright$\textbf{0.93 $\pm$ 0.15} & 0.72 $\pm$ 0.00 & $\triangleright$\textbf{0.71 $\pm$ 0.00} & 0.67 $\pm$ 1.00 & $\triangleright$\textbf{0.55 $\pm$ 1.00} & 0.80 $\pm$ 0.00 & $\triangleright$\textbf{0.61 $\pm$ 0.01} \\ 
polution & 0.90 $\pm$ 0.45 & $\triangleright$\textbf{0.81 $\pm$ 0.38} & 0.86 $\pm$ 0.04 & $\triangleright$\textbf{0.75 $\pm$ 0.04} & 0.90 $\pm$ 0.47 & $\triangleright$\textbf{0.82 $\pm$ 0.18} & 0.86 $\pm$ 0.23 & $\triangleright$\textbf{0.66 $\pm$ 0.13} \\ 
\midrule
Avg.Rank & 6.0 & 4.5 & 5.08 & 3.08 & 5.08 & 3.58 & 5.33 & 3.33 \\
\bottomrule
\end{tabular}
\end{sidewaystable}

\clearpage


\end{document}